%% file: 0_paper.tex
\documentclass[preprint,3p]{elsarticle}

\usepackage{times}
\usepackage{latexsym}
\usepackage{pgfplots}
    \pgfplotsset{compat=1.18}
\usepackage{graphicx}
    \graphicspath{{./}} 
    \usepackage[export]{adjustbox} 
\usepackage{caption}
\usepackage{algorithm}
\usepackage{algpseudocode} 
\usepackage{amsmath}
    \makeatletter
    \g@addto@macro\normalsize{%
    \setlength\abovedisplayskip{6pt plus 2pt minus 2pt}
    \setlength\belowdisplayskip{6pt plus 2pt minus 2pt}
    \setlength\abovedisplayshortskip{3pt plus 1pt minus 1pt}
    \setlength\belowdisplayshortskip{3pt plus 1pt minus 1pt}
    }
    \makeatother
\usepackage{booktabs}
\usepackage{url}
\usepackage{multirow}
\usepackage{array}
\usepackage{comment}
\usepackage{relsize}
\usepackage{subcaption}
\usepackage{enumerate}
\usepackage{paralist}
\usepackage[table]{xcolor}
\usepackage{placeins} 
\usepackage{hyperref}
\usepackage{listings}
\usepackage[draft]{minted} 
\usepackage{markdown}
\usepackage[normalem]{ulem}
\usepackage{xfp}

\hypersetup{
    colorlinks,
    linkcolor={red!50!black},
    citecolor={blue!50!black},
    urlcolor={blue!80!black}
}

\usepackage[T1]{fontenc}

\usepackage[utf8]{inputenc}

\usepackage{microtype}
\usepackage{float}

\usepackage{inconsolata}

\newcolumntype{P}[1]{>{\centering\arraybackslash}p{#1}} 

\input macros
\input smallist

\begin{document}

\begin{frontmatter}
\title{Comparing verbal, visual and combined explanations for\\ 
Bayesian Network inferences}
\author[inst1]{Erik P. Nyberg}
\ead{erik.nyberg@monash.edu}
\author[inst1]{Steven Mascaro\corref{cor}}
\ead{steven.mascaro@monash.edu}
\author[inst1]{Ingrid Zukerman}
\ead{ingrid.zukerman@monash.edu}
\author[inst2]{Michael Wybrow}
\ead{michael.wybrow@monash.edu}
\author[inst1]{Duc-Minh Vo}
\ead{minh.vo1@monash.edu}
\author[inst3]{Ann Nicholson}
\ead{ann.nicholson@monash.edu}
\address[inst1]{Dept of Data Science and AI,
Faculty of Information Technology
Monash University, Australia}
\address[inst2]{Dept of Human-Centred Computing,
Faculty of Information Technology
Monash University, Australia}
\address[inst3]{Faculty of Information Technology
Monash University, Australia}

\cortext[cor]{Corresponding author}

\begin{abstract} 
Bayesian Networks (BNs) are an important tool for assisting probabilistic
reasoning, but despite being considered transparent models, people have
trouble understanding them, and current User Interfaces (UIs)
do not clarify the reasoning of BNs. To address this problem, we have 
designed template-based verbal and visual extensions to the standard BN 
UI, which can guide users through common inference patterns.


We conducted a user study to compare four UI variations: only verbal explanations, 
only visual explanations, a combination of both modalities, and a baseline UI with
no extensions. Our main findings are: (1)~users' task
performance was better with all three types of extensions than with the 
baseline UI on \define{Finding questions} about the immediate impact 
of an observation, \define{Path questions} about the paths that enable 
this impact, and \define{Common effect questions} about
how an observation's impact on one of its possible causes 
is altered by observing another possible cause;
and (2)~the verbal and visual modalities together yield a better user 
performance than just the verbal modality for Path questions, and just
the visual modality for Finding questions.

In addition, we briefly explored the feasibility of using Large Language Models
(LLMs) to implement the verbal explanations. This yielded mixed 
results, as even with considerable guidance, the output of the LLM was 
sometimes inaccurate or inconsistent.
\end{abstract}


\end{frontmatter}

\input{_highlights}

\mysection{Introduction}
\label{section:intro}
\input 1_intro.tex

\mysection{Related Research}
\label{section:related}
\input 2_related.tex
\mysection{Explaining BNs}
\label{section:system}
\input 3_system

\mysection{Experiment}
\label{section:experiment}
\input 4_experiment

\mysection{Results}
\label{section:results}
\input 5_results

\mysection{Implementing the explanatory designs using a Large Language Model (LLM)}
\label{section:LLM}
\input 6_llm
\mysection{Discussion}
\label{section:discussion}
\input 7_discussion

\section*{Acknowledgments}
\vspace*{-1mm}
The authors are grateful to David Lagnado and Ulrike Hahn for advice on our experiment design, Matthias Klapperstueck for assistance in coding our experiment animations, Tim Powers for advice on our statistical analysis, and Shravani Rajguru, Yang Shen and Hao Peng for assistance in coding our web-based tool.

This research was supported in part by grant DP200100040 from the
Australian Research Council. Ethics approval for the user study
was obtained from Monash University Human Research Ethics Committee
(ID-35999).

\bibliography{references}
\bibliographystyle{elsarticle-num-names}

\clearpage

\appendix
\section{Bayesian networks}
\label{appendix:BNs}
\input AppendixA-BNs
\clearpage

\section{Skill tests}
\label{appendix:skills}
\input AppendixB-Skills
\clearpage


\section{Verbal template for multiple paths}
\label{appendix:templates}
\input AppendixC-Templates

\clearpage

\section{Visual semantics and algorithms}
\label{appendix:visual}
\input Appendix-Visual
\clearpage

\section{Screenshots from the experiment}
\label{appendix:screenshots}
\input AppendixD-Screenshots
\clearpage

\section{Additional participant performance statistics and experimental results}
\label{appendix:results}
\input AppendixE-Results

\newpage
\section{Additional notes on LLM implementation testing}
\label{appendix:llm}

\input AppendixF-LLMs

\end{document}

%% file: macros.tex
\definecolor{cerulean}{rgb}{0.0,0.48,0.65}
\definecolor{caribbean}{rgb}{0.0,0.8,0.6}
\definecolor{paleblue}{rgb}{0.82,0.93,1}
\definecolor{auburn}{rgb}{0.43,0.21,0.1}
\definecolor{byzantium}{rgb}{0.44,0.16,0.39}
\definecolor{byzantine}{rgb}{0.74,0.2,0.64}
\definecolor{violet}{rgb}{0.54,0.17,0.89}
\definecolor{cornell}{rgb}{0.7,0.11,0.11}
\definecolor{lgray}{gray}{0.85}
\definecolor{mgray}{gray}{0.7}
\definecolor{dgreen}{rgb}{0,0.55,0}
\definecolor{mgreen}{rgb}{0,0.7,0}
\definecolor{palered}{rgb}{1,0.70,0.75}
\definecolor{pastelyellow}{rgb}{0.99,0.99,0.65}
\definecolor{purple}{rgb}{0.63,0.13,0.94}
\definecolor{myblue}{RGB}{31, 119, 180}
\definecolor{myred}{RGB}{214, 39, 40}
\definecolor{mygreen}{RGB}{44, 160, 44}

\newcommand\trackEN[1]{\textcolor{black}{#1}}

\newcommand{\directquery}{{direct prompt}}
\newcommand{\guidedquery}{{guided prompt}}

\newcommand{\pvalue}{\mbox{\em p-value}}
\newcommand{\eg}{e.g., }
\newcommand{\ie}{i.e., }

\newcommand{\PR}{\hbox{Pr}}

\newcommand{\myitem}{\vspace*{0mm}\item}
\newcommand{\mysection}{\vspace*{0mm}\section}
\newcommand{\mysubsection}{\vspace*{0mm}\subsection}

\newcommand{\myparagraph}{\vspace*{0mm}\paragraph}

\newcommand{\var}[1]{\text{\texttt{\small{\textmd{#1}}}}}
\newcommand{\varft}[1]{\text{\texttt{\footnotesize{\textmd{#1}}}}}
\newcommand{\varsc}[1]{\text{\texttt{\scriptsize{\textmd{#1}}}}}
\newcommand{\vartab}[1]{\text{\texttt{\fontsize{8}{9}\selectfont{\textmd{#1}}}}}
\newcommand{\guivar}[1]{\text{\textmd{\textbf{#1}}}}
\newcommand{\guivartab}[1]{\text{\textbf{#1}}}

\newcommand{\state}[1]{\text{\texttt{\small{\textmd{#1}}}}}
\newcommand{\stateft}[1]{\text{\texttt{\footnotesize{\textmd{#1}}}}}
\newcommand{\statetab}[1]{\text{\texttt{\footnotesize{\textmd{#1}}}}}
\newcommand{\guistate}[1]{\text{\textit{#1}}}
\newcommand{\guistatetab}[1]{\text{\textit{#1}}}

\newcommand{\xcond}[1]{#1}
\newcommand{\qtype}[1]{\textit{#1}}
\newcommand{\correct}[1]{\underline{#1}}

\newcommand\ttest{$t$\nobreakdash-test}

\newcommand\say{\text{\textmd{\textsc{Say}}}}
\newcommand{\inst}[1]{\textcolor{dgreen}{#1}}
\newcommand{\func}[1]{\textsc{\footnotesize{#1}}}
\newcommand\sho{\func{Highlight}}
\newcommand\diffbox{\func{Diffbox}}
\newcommand\glow{\func{Glow}}
\newcommand\fade{\func{Fade}}
\newcommand\find{\var{finding}}

\newcommand\findsc{\varsc{finding}}
\newcommand\target{\var{target}}
\newcommand\BNpath{\var{path}}
\newcommand\BNpathft{\varft{path}}
\newcommand{\bnpath}{\var{path}}
\newcommand{\bnpaths}{\var{Paths}}
\newcommand{\ipath}{\var{ipath}}
\newcommand{\ipaths}{\var{iPaths}}
\newcommand{\apath}{\var{apath}}
\newcommand{\apaths}{\var{aPaths}}
\newcommand{\otherFinds}{\Finds \setminus \{\find\}}
\newcommand{\Finds}{\mathbf{F}}
\newcommand{\interFinds}{\mathbf{I}}
\newcommand{\Evid}{\mathbf{E}}
\newcommand{\arc}{\var{arc}}
\newcommand{\arcsc}{\varsc{arc}}
\newcommand{\define}[1]{\textit{#1}}

\lstdefinestyle{prompt}{
breaklines=true,breakatwhitespace=true,basicstyle=\ttfamily\scriptsize,keepspaces=true,breakindent=0pt,xleftmargin=0pt,xrightmargin=0pt,frame=single,escapeinside={(:}{:)}
}

\newminted[textprompt]{text}{
  breaklines,
  breakindent=0pt,
  breaksymbolleft={},
  fontsize=\scriptsize,
  frame=single,
  xleftmargin=0pt,
  linenos=false,
  labelposition=topline,
  escapeinside=||,
  framesep=2mm,
  vspace=1.5\baselineskip
}

\newcommand{\plusguided}{\normalfont\textit{\textbf{Per the direct prompt plus the following further instructions:}}}
\newcommand{\myhrule}{\raisebox{0.5ex}{\rule{\linewidth}{0.4pt}}}

\DeclareUnicodeCharacter{2248}{$\approx$}
\DeclareUnicodeCharacter{2212}{\ensuremath{\mathchar`-}}

\definecolor{LightBlue}{HTML}{80F0FF}
\definecolor{MidOrange}{HTML}{FF9933}
\newcommand{\cellgradient}[3]{
    \cellcolor{LightBlue!\fpeval{trunc((#3 - #1) / (#2 - #1) * 100, 0)}!MidOrange}#3
}
\newcommand{\colbleu}[1]{\cellgradient{0}{0.8}{#1}}
\newcommand{\colmeteor}[1]{\cellgradient{0}{0.9}{#1}}
\newcommand{\colbert}[1]{\cellgradient{0}{0.82}{#1}}

\newcommand{\qtypelabel}[1]{{\footnotesize\color{red}{$\boldsymbol{\longleftarrow}$ \qtype{#1}}}}

\definecolor{commentblue}{RGB}{80,110,150}

\algrenewcommand\algorithmiccomment[1]{%
    \hfill\textcolor{commentblue}{
        $\triangleright$ #1
    }%
}

\algnewcommand{\CommentCont}[1]{%
    \hfill\textcolor{commentblue}{
        #1
    }%
}

\algnewcommand{\LineComment}[1]{%
    \Statex \textcolor{commentblue}{
        $\triangleright$ #1
    }%
}

\algnewcommand{\ForAllOneLine}[2]{%
    \State \algorithmicforall\ #1 \algorithmicdo\ #2%
}

%% file: smallist.tex
\labelwidth\leftmargini\advance\labelwidth-\labelsep \labelsep 5pt
\def\@listi{\leftmargin\leftmargini
   \labelwidth\leftmarginii\advance\labelwidth-\labelsep
   \topsep 0pt plus 1pt minus 0.5pt
   \parsep 0pt plus 0.5pt minus 0.5pt
   \itemsep 0}
\def\@listii{\leftmargin\leftmarginii
   \labelwidth\leftmarginii\advance\labelwidth-\labelsep
   \topsep 2pt plus 1pt minus 0.5pt
   \parsep 1pt plus 0.5pt minus 0.5pt
   \itemsep 0}
\def\@listiii{\leftmargin\leftmarginiii
    \labelwidth\leftmarginiii\advance\labelwidth-\labelsep
    \topsep 1pt plus 0.5pt minus 0.5pt
    \parsep \z@ \partopsep 0.5pt plus 0pt minus 0.5pt
    \itemsep 0}
\def\@listiv{\leftmargin\leftmarginiv
     \labelwidth\leftmarginiv\advance\labelwidth-\labelsep}
\def\@listv{\leftmargin\leftmarginv
     \labelwidth\leftmarginv\advance\labelwidth-\labelsep}
\def\@listvi{\leftmargin\leftmarginvi
     \labelwidth\leftmarginvi\advance\labelwidth-\labelsep}

%% file: _highlights.tex
\section*{Highlights}
\begin{itemize}
    \item Template-based verbal and visual tools that clarify Bayesian network inferences.
    \item Novelties include multimodality, animations and counterfactual scenarios.
    \item Contributions of both evidence and arcs measured by ablation.
    \item Designs validated by testing novice user performance on basic inference tasks.
    \item Trial LLM implementation produced inconsistencies even with considerable guidance.
\end{itemize}

%% file: 1_intro.tex
Understanding how a conclusion is derived from evidence is crucial
to support human reasoning and decision making. Bayesian Networks
(BNs) assist in this task by modelling uncertainty and performing
probabilistic reasoning accurately. Unfortunately, even though BNs
are considered transparent models, this accuracy comes at the expense
of interpretability~\cite{Hennessy2020ExplainingBN}. The field of
\define{Explainable AI} (\define{XAI}\/) aims to make the outcomes of
AI models understandable, yet relatively few XAI methods have been devised for BNs.

User interfaces (UIs) for BNs have stabilised around a standard paradigm
that provides a common set of graphical features, \eg those in Netica and
GeNIe.\footnote{Netica$\,$\href{https://www.norsys.com}{www.norsys.com},$\,$GeNIe$\,$\href{https://www.bayesfusion.com}{www.bayesfusion.com}}
While this paradigm has provided many benefits (including several
BN software tools), methods for understanding and reasoning with BNs
have stagnated.  To address this shortfall, we have designed verbal
and visual extensions to the standard UI platforms, which increase the
transparency of BNs by guiding users through common inference patterns.

There are several user studies on the standard UI paradigm or
extensions thereof (Section~\ref{section:related}), but none of them
measure users' comprehension of different types of explanations ---
\citet{ButzEtAl2022}, \citet{Sevilla2024ExplainingBN} and
\citet{Babakov2025ExplainingBN} canvass users' views and obtain 
self-reports of understanding, but do not validate users' understanding 
through tasks. 
In contrast, we conducted an online experiment which comprises tasks that
require uses to demonstrate their understanding of a BN's reasoning. These
tasks enable us to compare users' understanding obtained from the unadorned 
elements of the standard BN UI, which is our baseline, with the understanding
obtained from the new verbal and visual elements we designed.

Our main contributions are as follows.
\begin{description}
\leftmargin=0pt \itemindent=0pt
\myitem[Design of Explanations:] 
We use a multimodal approach to clarify Bayesian network inferences, 
in which our verbal and visual tools can convey similar information in parallel, 
or be used alone. Our design is structured to support user queries, by presenting 
a brief summary first that includes multiple options to request more detail.
Novelly, we use ablation to compute the contributions of arcs as well as evidence,
and we explain evidence interactions by showing counterfactual scenarios.
Our visual explanations use colours and animations to show how altered contributions 
are passed along multiple active paths.
For our task-based validation study with novice users,
the visual component was semi-automated and the verbal component was
manually crafted with a template-based approach in mind,
in order to test how our best conceptions affect users' understanding of BNs.

\myitem[Experimental Results:]  
(1)~Users perform better with all three types of UI extensions
(visual, verbal and both combined)
than with the baseline UI (with no extensions) on three types of questions:
\qtype{Finding questions} about the 
immediate impact of an observation, \qtype{Path questions} about the 
paths that enable this impact, and \qtype{Common effect questions} about
how an observation's impact on one of its possible causes 
is altered by observing another possible cause.
(These \define{common effect} relationships~\cite{Pearl1988}
are essential for understanding the reasoning of BNs~\cite{KeshmirianEtAl2024}, 
but they are generally difficult to grasp~\cite{RehderWaldmann2017}.)
(2)~The verbal and visual modalities together yield a better user 
performance than just the verbal modality for Path questions, and just the 
visual modality for Finding questions. These results are statistically 
significant ($\pvalue < 0.05$).
\end{description}

This paper is organised as follows. Related research is discussed in
Section~\ref{section:related}, and our verbal and visual explanations
in Section~\ref{section:system}. Section~\ref{section:experiment}
describes the experimental design, followed by our results in
Section~\ref{section:results}. A preliminary implementation of the 
verbal explanation templates using a Large Language Model (LLM) appears 
in Section~\ref{section:LLM}, and concluding remarks in
Section~\ref{section:discussion}.

%% file: 2_related.tex
\begin{figure}[t]
\centering
\begin{subfigure}[b]{0.49\textwidth}
    \includegraphics[width=\textwidth]{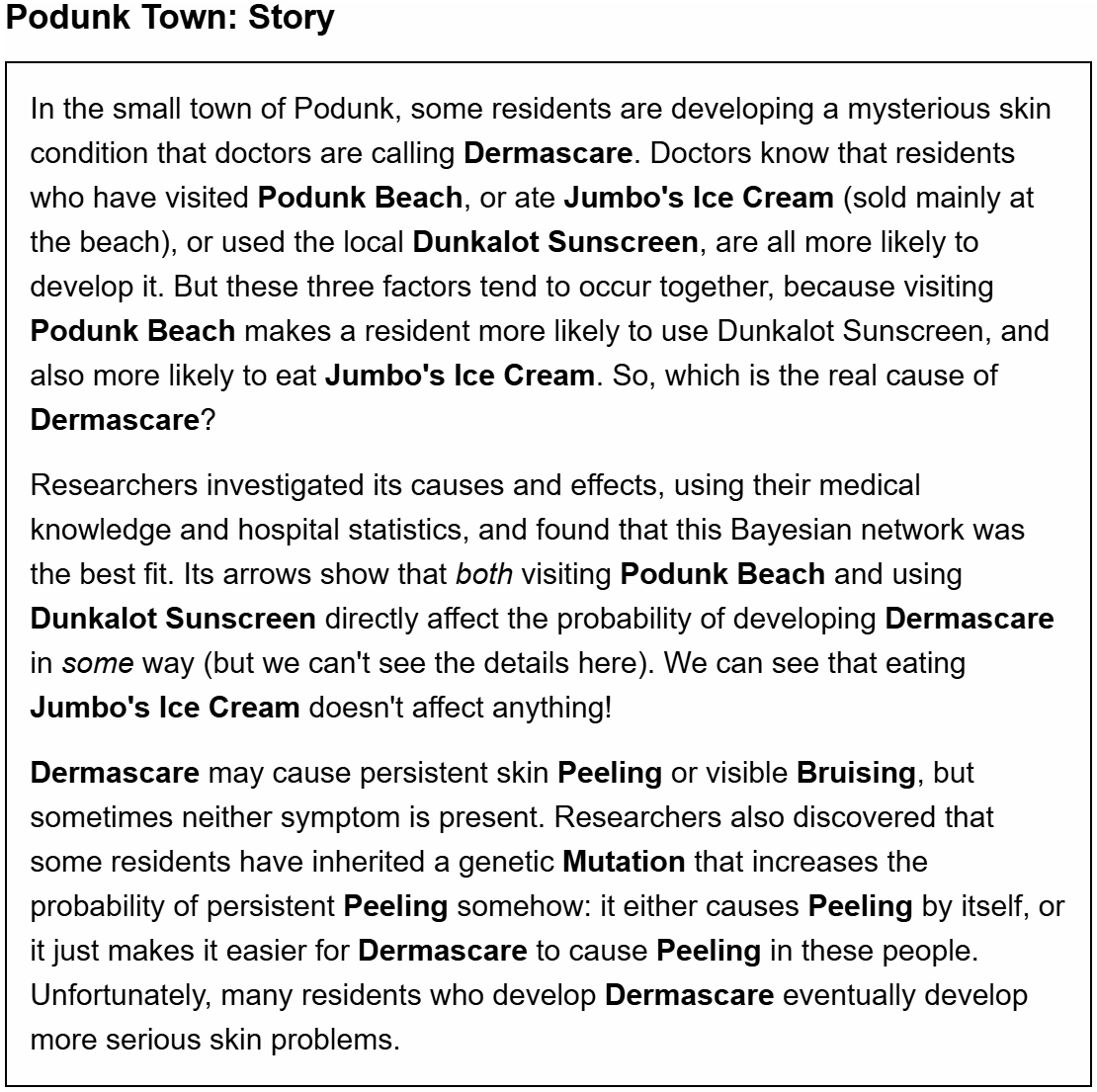}
    \caption{Podunk story\label{fig:podunk_story}}
\end{subfigure}
\hfill
\begin{subfigure}[b]{0.49\textwidth}
    \begin{minipage}{\textwidth}
        \centering
        \includegraphics[width=\textwidth]{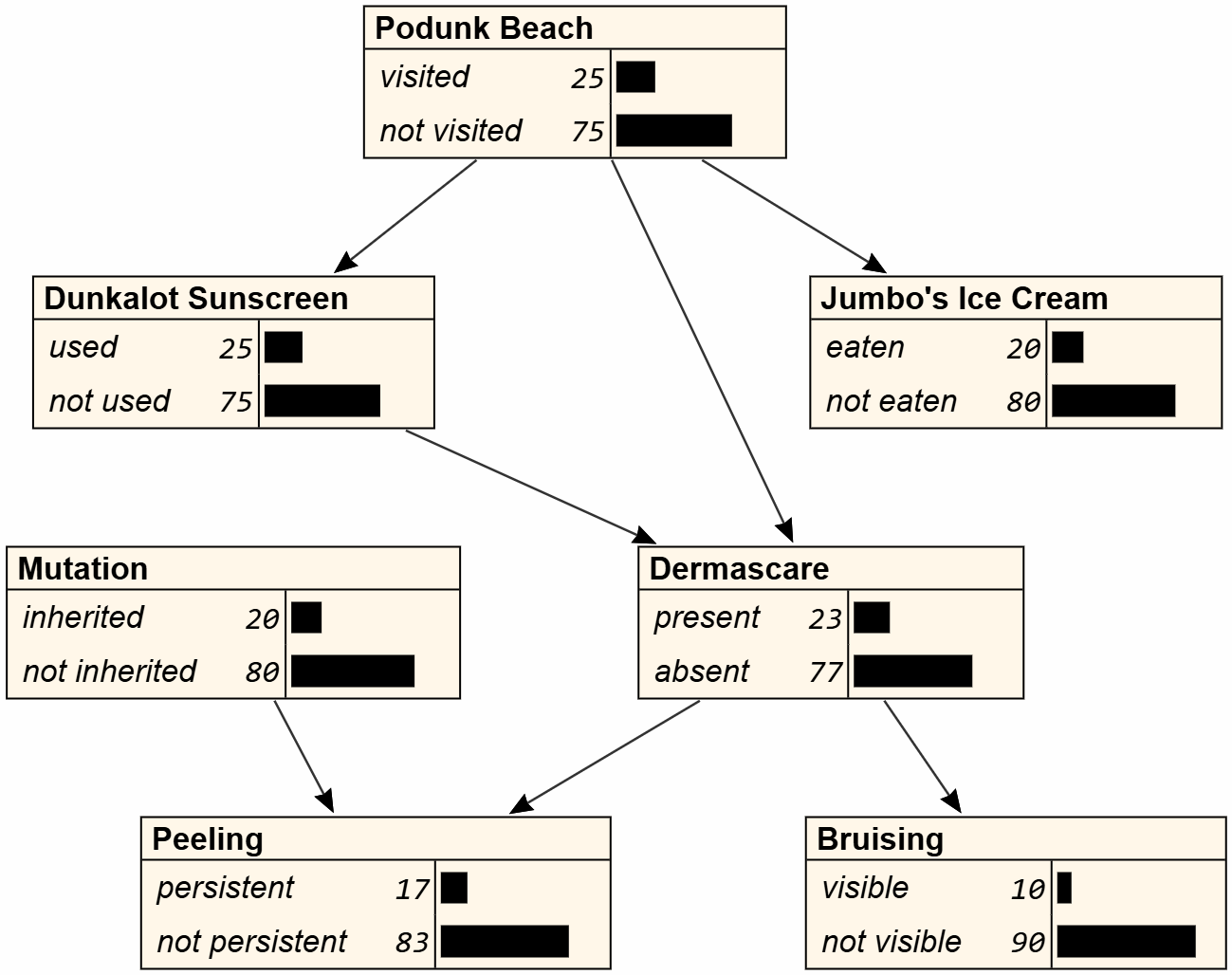}
        \vspace*{0.1\textwidth}
    \end{minipage}
    \caption{Podunk BN\label{fig:podunk_base}}
\end{subfigure}
    \caption{Narrative about Podunk Town, and Podunk BN with no instantiated variables.\label{fig:podunk_intro}}
\end{figure}

A Bayesian Network (BN) is a probabilistic graphical model that
represents a set of variables and their conditional dependencies
using a directed acyclic graph. The variables are encoded as
nodes, and 
each arc from a \define{parent node} to a \define{child node} encodes 
a dependency: the probability distribution of the child node 
given the states of its parents is specified by 
a \define{Conditional Probability Table} (\define{CPT}\/). 
In our examples, each arc reflects a causal influence.
Figure~\ref{fig:podunk_base} shows the Podunk BN from our 
experiment, 
with the accompanying narrative in Figure~\ref{fig:podunk_story}.
The target variable is \var{Dermascare}, a fictitious skin condition 
that may be caused by visiting \var{Podunk$\,$Beach} and/or applying 
\var{Dunkalot$\,$Sunscreen}. \var{Dermascare} in turn may cause \var{Bruising} and \var{Peeling}, and another cause of \var{Peeling} 
may be a genetic \var{Mutation}. This ``V-shape'', representing
two possible causes for \var{Peeling}, is a key pattern called a 
\define{common effect}~\cite{Pearl1988}, which can lead to inferences
that are difficult to understand.\footnote{Visiting \varft{Podunk$\,$Beach} 
and applying \varft{Dunkalot$\,$Sunscreen} form another V-shape with \varft{
Dermascare}, but this V-shape has a ``covering arc'' (between \varft{Podunk$\,$Beach}
and \varft{Dunkalot$\,$Sunscreen}), which makes it behave differently 
from an uncovered V-shape.}
Finally, visiting \var{Podunk Beach} may also cause people to consume 
\var{Jumbo's$\,$Ice$\,$Cream}.


BNs are useful for reasoning under uncertainty based on available 
evidence, and modelling complex decision-making 
processes~\cite{JanssensEtAl2004}. Their structure can reflect 
complex relationships, partly encoded in the rules of
d-separation~\cite{Pearl1988}, which dictate 
when an observed variable cannot change the probability of 
another variable. 
However, users often have trouble
understanding how these rules affect inferences in BNs~\cite{Rehder2025},
and may struggle to interpret CPTs~\cite{Eddy1982}. Even experts 
cannot foresee how evidence propagates over a long sequence of CPTs 
in a path, or which of several paths is stronger~\cite{PilditchEtAl2025}.
Thus, while BNs are considered transparent models, in practice their 
interpretability is impaired by these issues~\cite{Hennessy2020ExplainingBN}.


\mysubsection{Natural Language Explanations for BNs}
\label{subsection:template-nlg-bn}
\citet{Hennessy2020ExplainingBN} argue that natural language explanations
are essential for understanding BNs, and that ``the extraction of
information from a BN should be viewed as a content determination
stage''. Early research addressed various explanatory tasks, such as
generating ``micro explanations'' of the belief in individual
nodes~\cite{SemberZukerman1989}; generating qualitative explanations of
the structure, reasoning and interactions in a BN~\cite{Druzdzel1996};
listing observed and unobserved evidence and its paths of
influence~\cite{HaddawyEtAl1997}; and using an abductive process to
generate arguments that explain BN inferences~\cite{ZukermanEtAlAAAI98}.



In later work, \citet{LacaveEtAl2007} offered verbal
explanations of nodes, and graphical explanations of links
and chains of reasoning; and \citet{Yap2008ExplainingII} used the
\define{Markov blanket}~\cite{Pearl1988} to explain 
prediction reliability despite missing
data (the Markov blanket of a node comprises its parent nodes, its
children and the other parents of its children). 
\citet{TimmerEtAl2015,TimmerEtAl2017} extracted a \define{support
graph} from a BN, and given a set of observations, used it to build
arguments --- this research was extended in the legal domain with a
question-answering capability~\cite{Keppens2019ExplainableBN}, and 
with \define{idioms} (BN fragments) that reflect specific
scenarios~\cite{VlekEtAl2016}. \citet{NybergEtAl2022}'s BARD system
also employed common idioms, and generated statements about the effect
of new evidence on probabilities in a BN; and \citet{KyrimiEtAl2020}'s 
approach incrementally explained inferences of hybrid BNs (with discrete
and continuous nodes), selecting important evidence that supports or
contradicts a prediction, and determining intermediate variables. Finally,
\citet{Sevilla2024ExplainingBN} introduced a technique that explains BNs
using \define{factor arguments}, which are bipartite subgraphs of BNs
where nodes and factors (of joint probability distributions) are interleaved.


The only works that conduct quantitative user studies are that of 
\citet{ButzEtAl2022}, \citet{Sevilla2024ExplainingBN} and \citet{Babakov2025ExplainingBN}.
\citet{ButzEtAl2022} compared users' views and self-reported understanding of
the explanations in~\cite{VlekEtAl2016,TimmerEtAl2017,KyrimiEtAl2020}
and their own natural language realisation of the 
\define{Most Probable Explanation}~\cite{Pearl1988} (the assignment of
values to variables that maximises the conditional probability of these values 
given the evidence).
\citet{Sevilla2024ExplainingBN} and \citet{Babakov2025ExplainingBN} 
compared users' views and self-reported understanding of baseline explanations
and explanations generated by the above-mentioned factor argumentation 
procedure. However, these understandings were never verified with task performance.

All these methods employ template-based approaches to generate
explanations (and so do we, at this stage). As noted by
\citet{Hennessy2020ExplainingBN}, such approaches are valuable for
illustrating local patterns, but do not adapt or scale well.





\mysubsection{Visual Explanations for BNs}
\label{subsection:visual-cbns}
Several attempts to explain BNs have employed visual tools, \eg
\cite{ZapataRivera1999VisualizationOB,Chiang2006VisualizingGP,Cossalter2011VisualizingAU,Champion2017VisualizingTC}.
Early systems like {Netica} and {GeNIe} used pie charts and bar charts for
marginal and, in some cases, conditional probabilities~\cite{druzdzel1999smile}, 
offering scalable overviews,
and isolating node-level information. \citet{Chiang2006VisualizingGP} used
heat maps to visualise CPTs, such that each cell encodes a probability value
as a colour, and employed edge colours (green/red) to encode the sign of the 
correlation coefficient between nodes. These tools have emphasised node-level 
probability inspection, rather than aiding users' understanding of inferences.

Belief propagation was addressed by several systems.
\citet{ZapataRivera1999VisualizationOB} compared the usability
of different cues (temporal order, colour, size, proximity and
animation) for visualising changes in probabilities, and later
enabled interactive belief tracing through direct inspection of
evidence impact~\cite{ZapataRivera2000InspectingAV}. \citet{Cossalter2011VisualizingAU}
used \define{bubble lines} that show how beliefs change after evidence 
is added. This design helped users follow belief updates without 
losing sight of the overall BN structure, but it did not explain
why these updates occurred. Lastly, \citet{Champion2017VisualizingTC}'s
\define{inference diffs} highlighted belief changes across network states,
ranking nodes by the magnitude of their impact.

%% file: 3_system.tex

For the main experiment, we employed (1)~verbal explanations
derived from templates (\ref{appendix:templates}), and 
(2)~visual explanations comprising both static and interactive animated markup 
generated using a semi-automated tool (\ref{appendix:visual}). Both have been incorporated into a 
prototype UI that has been shown to produce understandable
verbal and visual explanations, which incorporate important BN 
elements, such as causal relations, multiple interacting reasoning paths,
\define{explaining-away}~\cite{Pearl1988} and \define{enabling}~\cite{FentonNeil2018} relations.
Explaining-away is a type of inference where we know a common effect
may have occurred due to either of two causes.
Learning that one cause is present helps to explain the effect, 
which \emph{lowers} the probability of the other cause being present.
However, there are other types of inference via a known common effect.
An enabler cannot cause a common effect by itself, 
but does increase the power of another cause to produce it. 
Here, learning that the enabler is present \emph{increases} the probability
that the other cause has produced the effect.

\mysubsection{Measuring the contribution of features}
We measure the contribution of three key features: findings, arcs and interactions 
between findings. These contributions are conveyed verbally and visually.

\begin{figure}[t]
\centering
 \begin{subfigure}[b]{0.17\textwidth}
    \begin{minipage}{\textwidth}
        \centering
        \includegraphics[width=\textwidth]{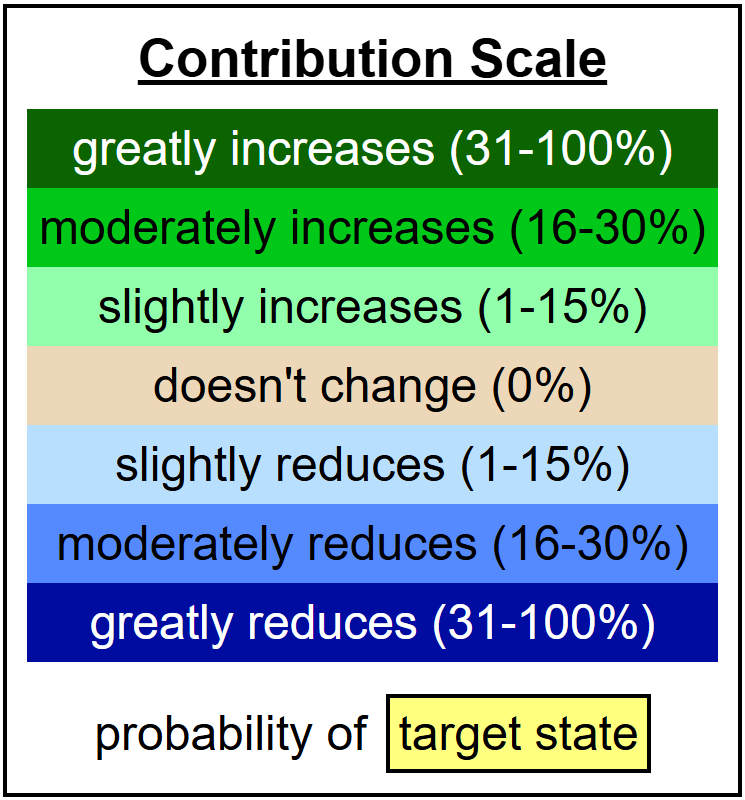}
    \end{minipage}
    \caption{Contribution scale\label{fig:podunk_summary_scale}}
\end{subfigure}
    \hfill
\begin{subfigure}[b]{0.45\textwidth}
    \includegraphics[width=\textwidth]{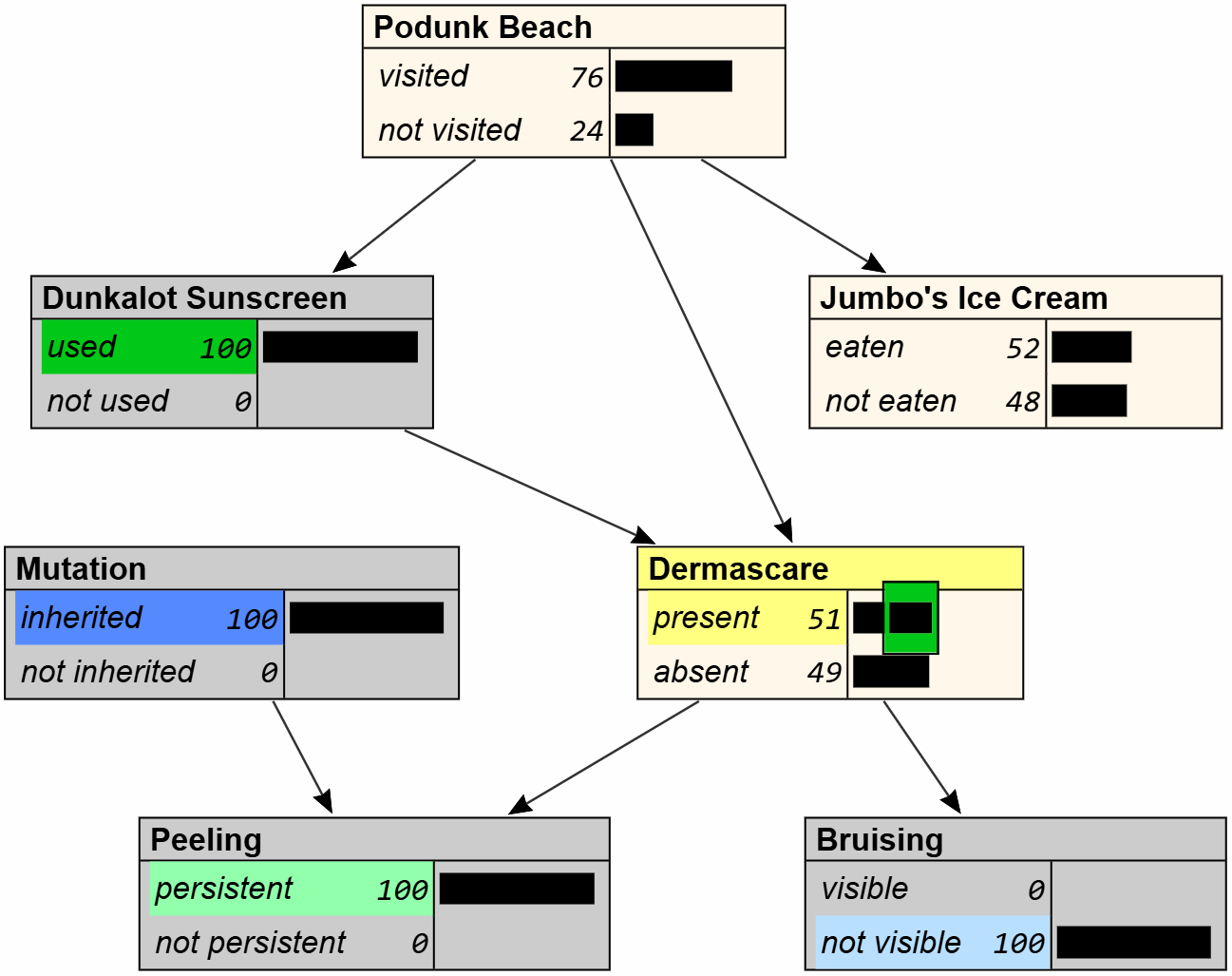}
    \caption{Visual explanation\label{fig:podunk_summary_visual}}
\end{subfigure}
    \hfill
\begin{subfigure}[b]{0.35\textwidth}
    \begin{minipage}{\textwidth}
        \includegraphics[width=\textwidth]{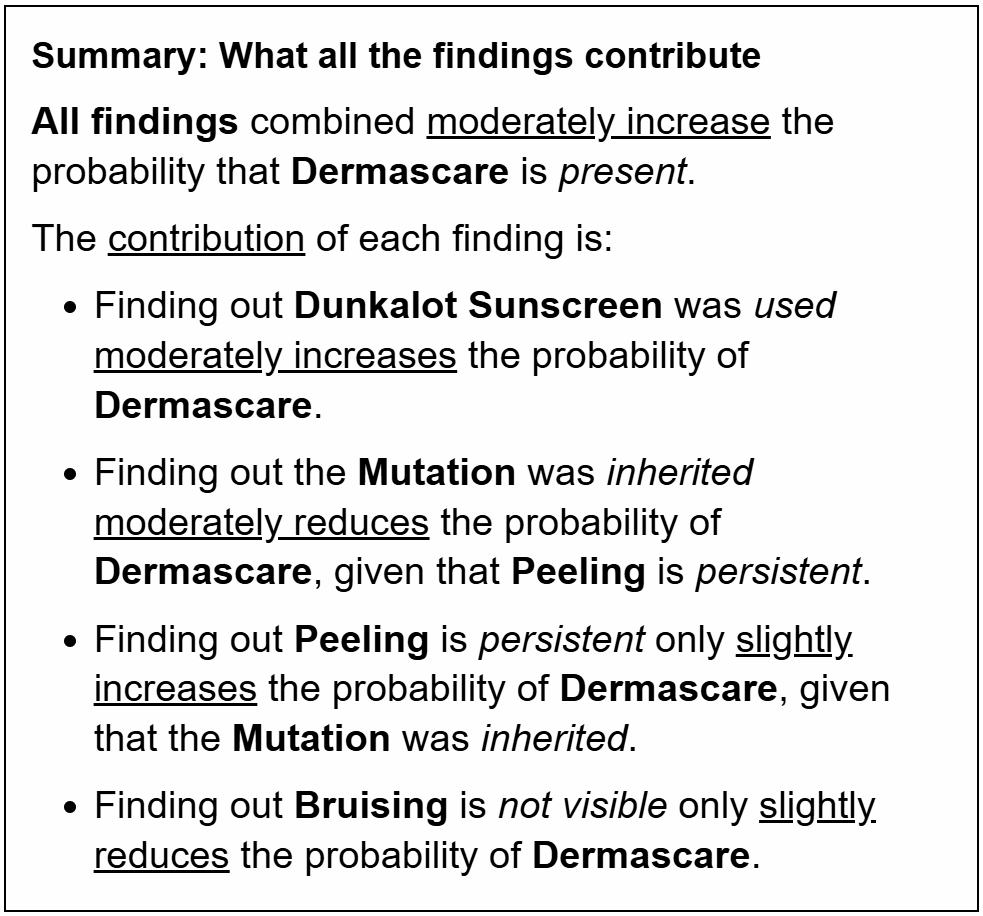}
        \vspace*{0.015\textwidth}
    \end{minipage}
    \caption{Verbal explanation\label{fig:podunk_summary_verbal}}
\end{subfigure}
\caption{Summary explanation: (a)~scale, (b)~visual, and (c)~verbal explanation 
for the contribution of four observations in the Podunk BN to the probability 
that \varft{Dermascare} is present.\label{fig:podunk_summary}}
\end{figure}

\begin{description}
\leftmargin=0pt \itemindent=0pt
\item[Findings:] One of the main difficulties faced by BN users is that the order in
which observations are added to a BN can radically alter the difference
each observation makes, even though the combined impact of all observations is the same.
We circumvent this problem by 
defining an ablation measure: the \define{contribution} of each finding in the set of 
all findings, $\find \in \Finds$, is how much it alters the probability of a target state 
when added to everything else we know (\ie the other variable instantiations
$\otherFinds$ and the BN $G$). These contributions calculated by $C_G(\cdot)$ are 
expressed as differences in percentage points (\eg $+25\%, -17\%$):
$$C_G(\find,\Finds) = \PR_G(\target|\Finds) - \PR_G(\target|\otherFinds)$$
They are conveyed verbally and visually using the 
terms and colour coding respectively of the scale in Figure~\ref{fig:podunk_summary_scale}. 

\item[Arcs:] Novelly, we also determine and convey the contributions 
of arcs by calculating the difference between the contribution of 
any set of findings $\Evid$ with or without a particular \arc: 
$$C_G(\arc, \Evid) = C_G(\Evid) - C_{G\setminus\arcsc}(\Evid)$$
Let $X$ be the parent of $Y$ through \arc, 
$\Pi_G(Y)$ be the parents of $Y$ in BN $G$, and $U = \Pi_G(Y)\setminus X$ be the parents of $Y$ in $G$ not including $X$. We implement the removal of \arc\ from $G$ by marginalising $X$ from the CPT of $Y$, using the probabilities for $X$ conditioned on $U$ in $G$:
$$\PR_{G\setminus\arcsc}(Y|\Pi_{G\setminus\arcsc}(Y)) = \sum_{X}\PR_G(Y|U,X)\PR_G(X|U)$$
This preserves $\PR_G(Y|U)$, such that $\PR_{G\setminus\arcsc}(Y|U)=\PR_G(Y|U)$, but the full joint distribution now contains the additional independence assumption $Y \perp X | U $. This is the same approach used by \citet{vanengelen1997} for simplification of a BN via arc deletion,
who also proved that the KL divergence it produces with the original graph is equal to 
the conditional mutual information of the removed arc. 
Hence, this procedure is a reasonable candidate for measuring the effect of arc removal, 
and thereby arc contribution.

A finding can contribute only via its active paths (a.k.a.~d-connections) to the target, 
the set \bnpaths. To see how much each arc on a path allows the 
finding to contribute, we start by calculating the contribution of the arc given all the evidence,
$C_G(\arc, \Finds)$. 
We then contrast this with its contribution in the alternative scenario where the finding is removed, 
$C_G(\arc, \Finds\setminus\find)$. 
This arc contribution difference 
reveals how much the arc enables the finding's contribution.

We explain these contribution differences in sequence along each $\bnpath \in \bnpaths$, 
including the corresponding changes in probability distributions on the intermediate variables. 
Contrasts in arc contributions are shown by adding or changing arc colours in our animations (Section~\ref{section:visual}).
In our experiment BNs,
each intermediate element 
(arc or variable) passes on the same contribution difference to the next 
until the path reaches the target.
We call a path \define{stronger} than another if it passes on a larger contribution difference,
and explain 
stronger paths before weaker ones.

\item[Interactions between findings:] We say other findings $\interFinds$
\define{interact} with \find\ if knowing $\interFinds$ substantially alters the 
contribution of \find, viz. 
$|C_G(\find,\Finds) - C_G(\find,\Finds\setminus\interFinds)| \geq 5\%$.\footnote{This formula 
is equivalent to $|C_G(\interFinds,\Finds) - C_G(\interFinds,\Finds\setminus\{\findsc\})| \geq 5\%$, 
showing that interaction is a symmetrical relation.
$\pm 5\%$ was chosen because this is the tolerance for ``about an even chance'' 
in the probability scale defined in~\cite{ICD-203-2015}.}
These other findings 
can only alter the \target\ probability in this way 
via $\ipaths$: paths from $\interFinds$ to \target\ that are active given $\Finds$ 
or active given $\Finds\setminus\{\find\}$. So, to explain interactions we 
only need to mention or highlight elements (variables and arcs) in a \bnpath\ or \ipath, 
$\var{el} \in \bigcup\bnpaths\cup\bigcup\ipaths$. 
We compute three contribution differences for these elements, to explain:
\begin{enumerate}
    \item[1.]
    how adding the finding without the interacting findings 
    makes a difference to element contributions, 
    which change from 
    $C_G(\var{el},\Finds\setminus(\{\find\}\cup\interFinds))$ to $C_G(\var{el},\Finds\setminus\interFinds)$.
    \item[2a.]
    how instead adding the interacting findings 
    without the finding changes the situation, \ie which 
    probability or path differences will alter the contribution of \find, 
    \eg increasing the probability of a common effect, or blocking a path. 
    These differences will be reflected in element contributions, which change here from 
    $C_G(\var{el},\Finds\setminus(\{\find\}\cup\interFinds))$ to $C_G(\var{el},\Finds\setminus\{\find\})$, 
    but some contribution differences due to $\interFinds$ may not be evident until the next step,
    since adding \find\ can activate some \ipaths.
    \item[2b.]
    how adding the finding with  the interacting findings 
    now makes a difference to 
    the element contributions, 
    which change from $C_G(\var{el},\Finds\setminus\{\find\})$ to $C_G(\var{el},\Finds)$.
\end{enumerate}
\end{description}

All these contribution calculations for findings, arcs and interactions were automated in our experiment and prototype.

\mysubsection{Verbal explanations}
\label{section:verbal}
BN explanations can become unwieldy when they involve multiple 
observations and paths. As a first step to mitigate this problem, we 
divide our verbal (and visual) output into one 
summary explanation and several detailed explanations that
provide additional information. 
\begin{itemize}
\item
A summary explanation 
contains an overview of the combined impact
of all observations (``All findings'' segment of the text box in 
Figure~\ref{fig:podunk_summary_verbal}), and a list of the contributions of
the individual findings ranked in descending order of magnitude, noting
any interacting findings (``Contributions'' segment in the textbox in
Figure~\ref{fig:podunk_summary_verbal}).
\item
A detailed explanation for each finding describes
the paths that make a difference to its
contribution (Figure~\ref{fig:podunk_sunscreen_verbal}), and offers a
counterfactual explanation about the interaction between this 
finding and others (Figure~\ref{fig:podunk_peeling_verbal}).
\end{itemize}

The contribution scale in Figure~\ref{fig:podunk_summary_scale}
(replicated in Figures \ref{fig:podunk_sunscreen_scale} and~\ref{fig:podunk_peeling_scale}) 
shows the terms employed for different probability changes of the target state.
These changes were sourced from~\cite{ElsaesserHenrion1990}, and adapted as 
done in~\cite{MarufEtAl2023} to reduce the number of categories. In addition,
we made the categories for ``greatly increases'' and ``greatly reduces''
symmetric, in line with the other categories.

Verbal explanations for our user study were manually produced according
to templates like those in \ref{appendix:templates}, yielding texts
such as those in Figures~\ref{fig:podunk_summary_verbal}, \ref{fig:podunk_sunscreen_verbal} 
and~\ref{fig:podunk_peeling_verbal}.
Basic versions of these templates have been implemented in a prototype, where
the two detailed explanations 
(Figures \ref{fig:podunk_sunscreen_verbal} and~\ref{fig:podunk_peeling_verbal})
can be obtained by clicking on the corresponding observation in the summary 
explanation (Figure~\ref{fig:podunk_summary_verbal}).
The feasibility of using Large Language Models to generate verbal explanations similar 
to ours is explored in Section~\ref{section:LLM}.

\mysubsection{Visual explanations}
\label{section:visual}
Our visual explanations provide almost identical information
to the verbal explanations. The contribution scale in Figure~\ref{fig:podunk_summary_scale}
(replicated in Figures \ref{fig:podunk_sunscreen_scale} and~\ref{fig:podunk_peeling_scale})
shows the colours used to convey the sign and magnitude of contributions
to the target variable.

\begin{itemize}
\item
The summary explanation in Figure~\ref{fig:podunk_summary_visual} shows the 
contribution of each finding by a coloured highlight applied to 
its state name, 
and the combined impact of all findings is shown by a coloured 
box drawn on the target probability bar around the difference in probability (which may be an increase or reduction).

\item
The static detailed explanation about the \var{Sunscreen} finding in
Figure~\ref{fig:podunk_sunscreen_visual} and the \var{Peeling} finding
in Figure~\ref{fig:podunk_peeling_visual}  
also shows the contributions of 
arcs with coloured highlights, similar to~\cite{LacaveEtAl2007}. 
BN elements are faded if they do not make much difference to the finding's 
contribution, \ie are variables or arcs not in $\bnpaths$ or $\ipaths$,
\eg \var{Mutation}, \var{Peeling}, \var{Bruising} and the arcs that connect them in Figure~\ref{fig:podunk_sunscreen_visual}.

\begin{figure}[!ht]
    \centering
\begin{subfigure}[b]{0.17\textwidth}
    \begin{minipage}{\textwidth}
        \centering
        \includegraphics[width=\textwidth]{fig2a_scale_coloured-400}
        \vspace*{0.115\textwidth}
    \end{minipage}
    \caption{Contribution scale\label{fig:podunk_sunscreen_scale}}
\end{subfigure}
    \hfill
\begin{subfigure}[b]{0.45\textwidth}
    \begin{minipage}{\textwidth}
        \centering
        \includegraphics[width=\textwidth]{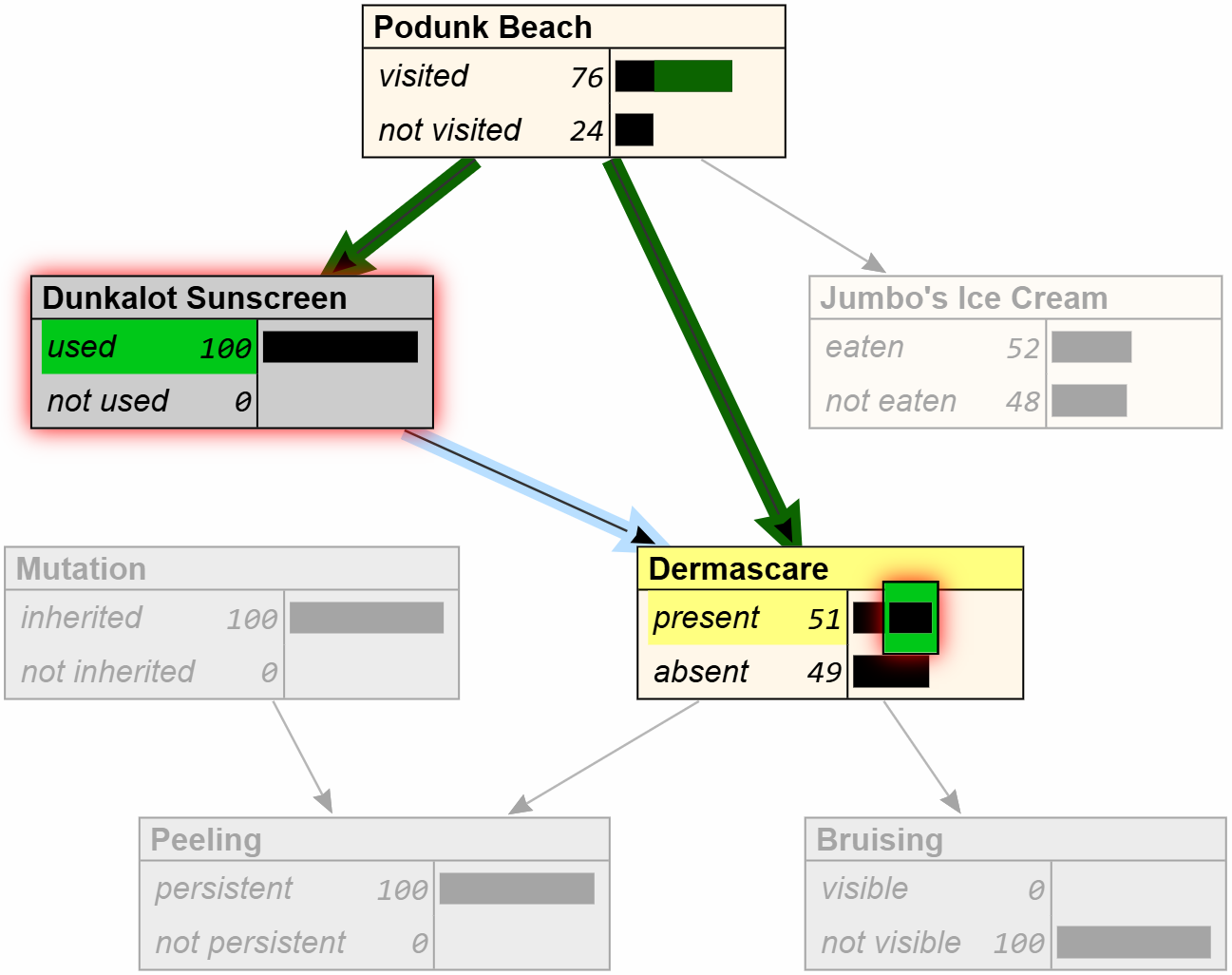}\vspace*{1ex}\\
        \hspace*{-0.075\textwidth}\includegraphics[width=1.15\textwidth]{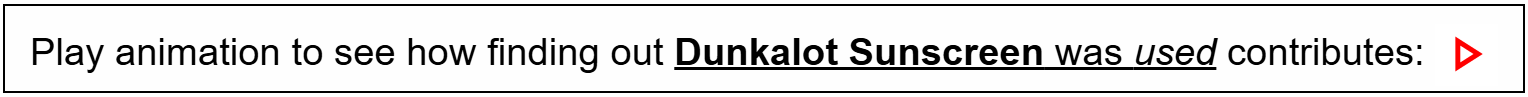}
    \end{minipage}
    \caption{Visual explanation\label{fig:podunk_sunscreen_visual}}
\end{subfigure}
    \hfill
\begin{subfigure}[b]{0.35\textwidth}
    \begin{minipage}{\textwidth}
        \centering
        \includegraphics[width=\textwidth]{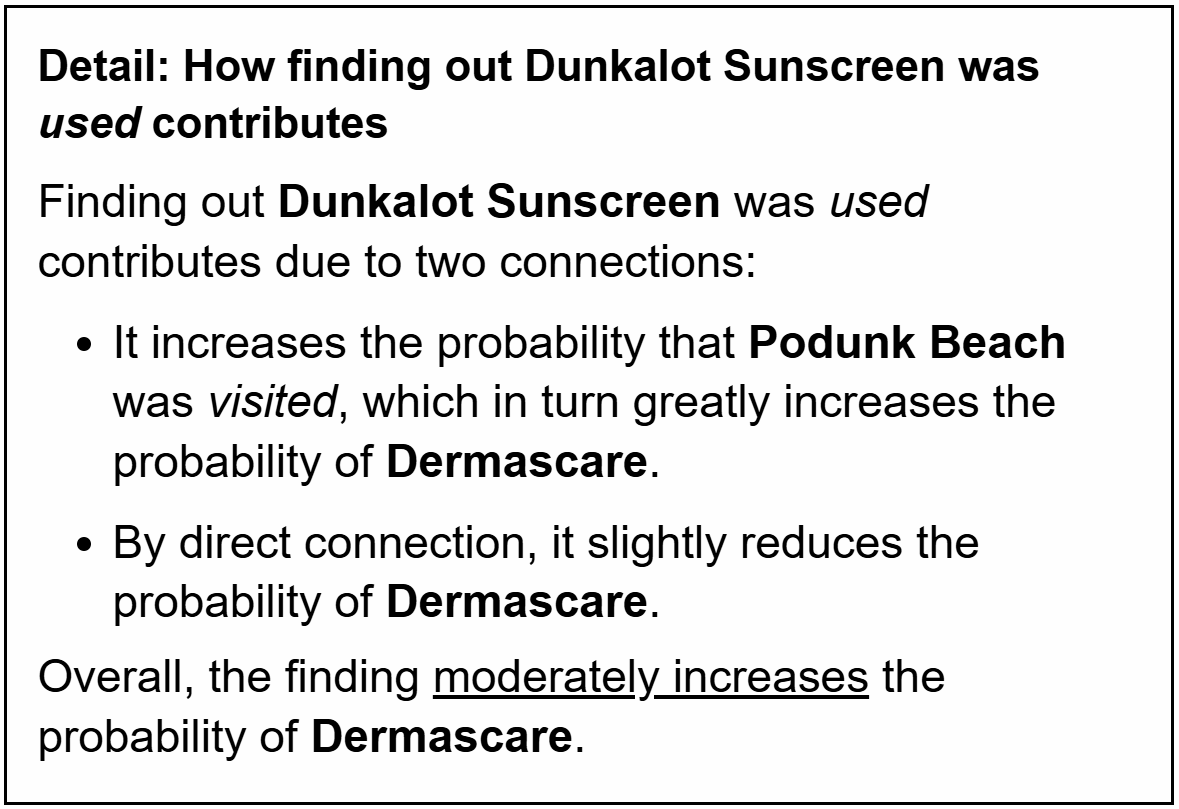}
        \vspace*{0.385\textwidth}
    \end{minipage}
    \caption{Verbal explanation\label{fig:podunk_sunscreen_verbal}}
\end{subfigure}
    \caption{Detailed explanation about Dunkalot sunscreen: (a)~scale, (b)~visual, and (c)~verbal explanation of the contribution of the \varft{Dunkalot$\,$Sunscreen} finding in the Podunk BN to the probability of \varft{Dermascare}.\label{fig:podunk_sunscreen}}
\end{figure}

\begin{figure}[!ht]
    \vspace{2ex}
    \centering
\begin{subfigure}[b]{0.17\textwidth}
    \begin{minipage}{\textwidth}
        \centering
        \includegraphics[width=\textwidth]{fig2a_scale_coloured-400}
        \vspace*{0.49\textwidth}
    \end{minipage}
    \caption{Contribution scale\label{fig:podunk_peeling_scale}}
\end{subfigure}
    \hfill
\begin{subfigure}[b]{0.45\textwidth}
    \begin{minipage}{\textwidth}
        \centering
        \includegraphics[width=\textwidth]{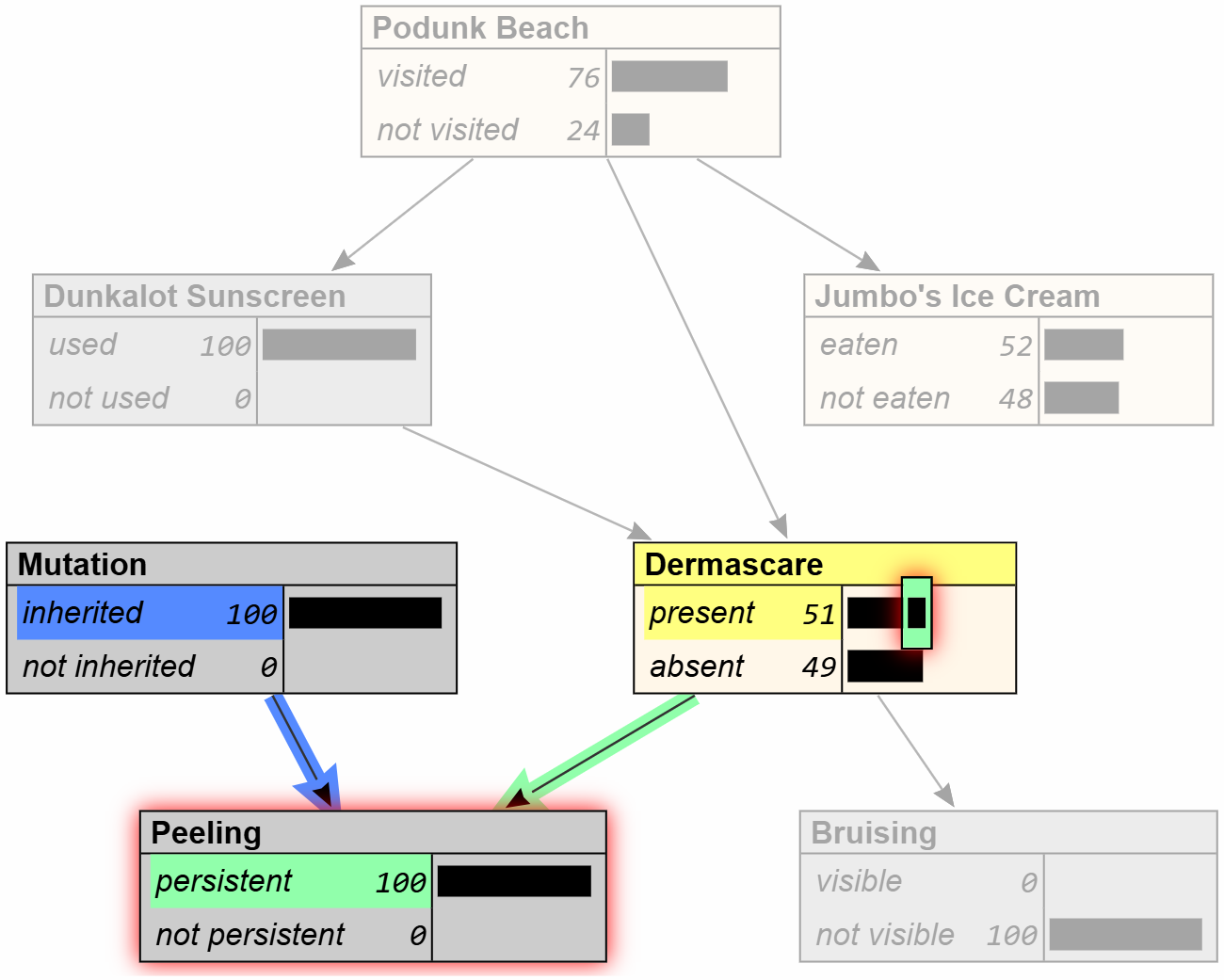}\vspace*{1ex}\\
        \includegraphics[width=\textwidth]{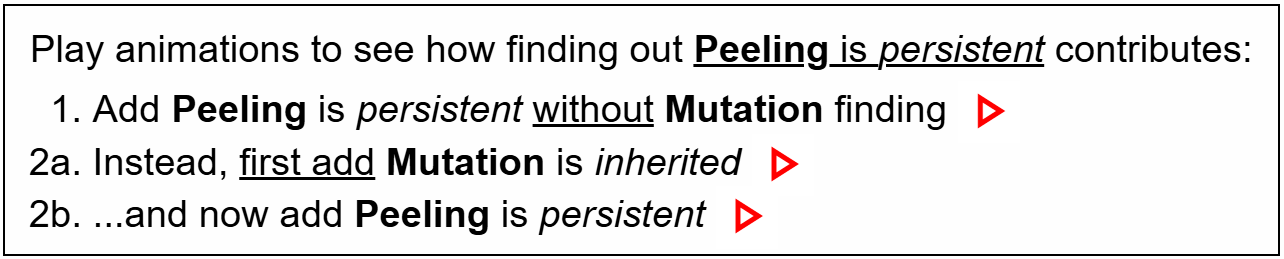}
    \end{minipage}
    \caption{Visual explanation\label{fig:podunk_peeling_visual}}
\end{subfigure}
    \hfill
\begin{subfigure}[b]{0.35\textwidth}
    \begin{minipage}{\textwidth}
        \centering
        \includegraphics[width=\textwidth]{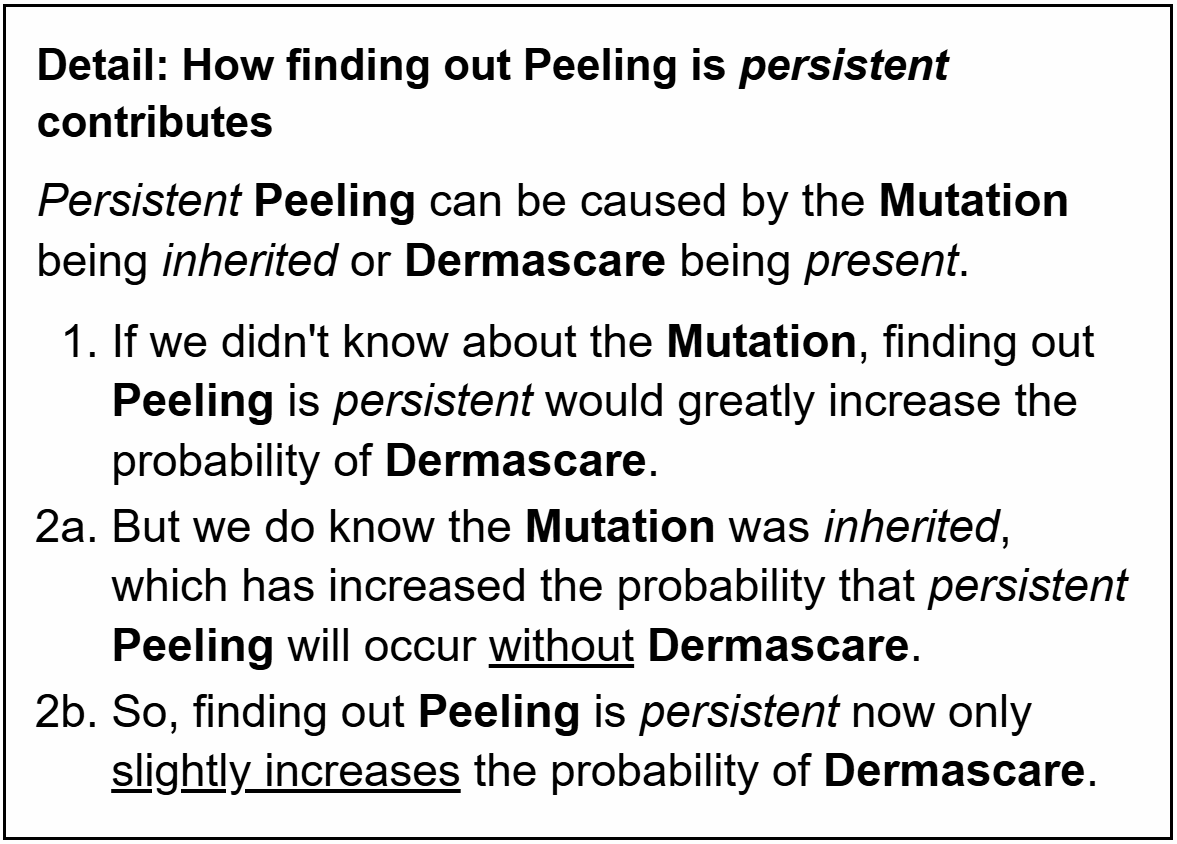}
        \vspace*{0.529\textwidth}
    \end{minipage}
    \caption{Verbal explanation\label{fig:podunk_peeling_verbal}}
\end{subfigure}
    \caption{Detailed explanation about Peeling: (a)~scale, (b)~visual, and 
    (c)~verbal explanation of the contribution of the \varft{Peeling} finding
    in the Podunk BN to the probability of \varft{Dermascare}.\label{fig:podunk_peeling}}
\end{figure}

\item
Animations are used to provide more detail. They mirror the verbal
explanations, and reveal features that the static visuals fail to
capture, such as individual paths that overlap with others,
or the 
way interacting findings change the finding's contribution. 
Each animation matches a 
path description from the text box, 
progressively highlighting the sequence of arcs and probability changes 
on intermediate variables to reveal their new contributions.
The animation for Figure~\ref{fig:podunk_sunscreen} shows the stronger and weaker paths highlighted in turn, before their combined contribution is highlighted on the \var{Dermascare} probability bar (similar to the animation snapshots in Figure~\ref{fig:rats_ani}).

The three animations for Figure~\ref{fig:podunk_peeling} illustrate (1)~a
counterfactual explanation of how the \var{Peeling} finding
contributes to \var{Dermascare} without the \var{Mutation} finding, 
then (2a)~how the latter changes the situation, before (2b)~showing how the 
\var{Peeling} finding now contributes to \var{Dermascare} (2b is similar to
the animation snapshots in Figure~\ref{fig:mutation_ani}). 
The interface provides controls to pause and replay each animation.
In our experiment, these were buttons in a text box below the BN along with text indicating what the animations show, as in Figures \ref{fig:podunk_sunscreen_visual} and \ref{fig:podunk_peeling_visual}.
\end{itemize}




Visual explanations for our user study were specified and automated according to the 
semantics and algorithms in \ref{appendix:visual}, yielding both static colouring and animations 
such as those in Figures~\ref{fig:podunk_summary_visual}, \ref{fig:podunk_sunscreen_visual} 
and~\ref{fig:podunk_peeling_visual}. Basic versions of these visual explanations have also 
been implemented in our prototype, where the two detailed explanations 
(Figures \ref{fig:podunk_sunscreen_visual} and~\ref{fig:podunk_peeling_visual})
can be obtained by clicking on the corresponding BN variable name in the summary 
explanation (Figure~\ref{fig:podunk_summary_visual}).

%% file: 4_experiment.tex
Our between-subjects experiment assesses (1)~users' performance in answering questions of
different types (\eg Findings, Path, Common effect) about two simple BNs 
under different explanatory conditions -- Verbal, Visual, Neither (standard 
BN UI) and Both (Visual and Verbal combined);
and (2)~users' views about the usability of the UI under each of these 
explanatory conditions, and optionally, their views about the 
difficulty of different parts of the experiment.\footnote{Full length PDF examples of our surveys, 
video clips of our survey animations,
user background information and all their responses for this experiment are publicly available at \href{https://doi.org/10.26180/30673622.v1}{doi.org/10.26180/30673622.v1}} As mentioned above, 
users' performance was not assessed in previous studies; rather,
their perceptions of their own performance was assessed.

The contribution scale (as in Figure~\ref{fig:podunk_summary_scale}, but without 
colour) was added to the baseline BN UI (Neither condition) to enable
users to correctly interpret the contribution terms in the assessment 
questions.
It is also worth noting that the training BN and the UIs for all the explanatory
conditions omit the standard UI functionality for adding findings or
inspecting CPTs, as all UIs were simulated on the Qualtrics platform.

\mysubsection{Materials} 
\paragraph{Question types}
We grouped our BN multiple-choice questions into the following six question 
types,
based on the information needed to answer them, and whether BN-based
inferences were required --- a correct answer received 1 mark
(Table~\ref{tab:qtypes} illustrates these question types for the Podunk BNs
in Figures~\ref{fig:podunk_intro}--\ref{fig:podunk_peeling}).

\begin{itemize}
    \myitem
\qtype{Control} and \qtype{ControlThink} questions require no information
from the explanations, but the latter requires inferences from the BN
(rows~1 and~2 in Table~\ref{tab:qtypes}).

\myitem
\qtype{Finding} questions require knowing the contribution of an
observation, \ie how adding it changes the target’s probability (row~3).


\myitem
\qtype{CommonEffect} questions require understanding the impact
of interacting findings within common-effect relations (row~4).


\myitem
\qtype{Path} and \qtype{PathThink} questions require understanding how
paths affect a finding's ability to contribute, but the latter requires
inferences from the BN (rows~5 and~6).
\end{itemize}

These questions cover the most common elements of a BN, including structural relationships and marginal probabilities (\qtype{Control}), 
evidence contributions (\qtype{Finding}) and the three-node patterns: chain and common cause (\qtype{Path}) and common effect (\qtype{CommonEffect}). The \qtype{Think} variants were designed to test differences in participants' abilities to perform counterfactual reasoning based on past explanations they received in their explanatory condition. These variants only exist for \qtype{Control} and \qtype{Path}, since although \qtype{Finding} and \qtype{CommonEffect} also include counterfactual questions, the answers are explicitly provided by our explanations (\eg\ Figure~\ref{fig:podunk_peeling_verbal}).

\input tab-question-type

\myparagraph{Test BNs}
Participants were tested using the Rats BN (Figure~\ref{fig:rats_intro}) 
followed by the Podunk BN (Figure~\ref{fig:podunk_intro}). Our BNs have only a small number
of binary variables, but they were carefully designed to include key BN
features and types of inferences. Our stories were designed to describe hypothetical 
scenarios that would be easy to grasp, but not enable participants to correctly answer 
our non-\qtype{Control} questions using only the story and their prior knowledge. 
We provided a button in the margin of every page allowing participants to re-read the story
at any time.

Our pilot study showed clear biases in participants' answers
resulting from guesswork based on common sense. To neutralise
such biases, we devised two versions of each of our BNs, $A$ 
and $B$, that had different underlying CPTs and different correct answers 
for all types of questions (Figures \ref{fig:rats_sum}--\ref{fig:rats_ecstazine}, \ref{fig:podunk_sum}--\ref{fig:podunk_mutation}), but had the same structure 
and similar probabilities with no findings (Figures \ref{fig:rats_intro}, \ref{fig:podunk_base}) 
and with all the findings (Figures \ref{fig:rats_sum}, \ref{fig:podunk_sum}). 
For example, the \var{Sunscreen} finding 
makes the same contribution to \var{Dermascare} and yields 
the same probabilities for the intermediate variable 
\var{Podunk$\,$Beach} in Podunk$\,A$ and Podunk$\,B$ 
(Figure~\ref{fig:podunk_sum}). 
However, in $A$, the indirect path via \var{Podunk$\,$Beach} makes the
larger difference, while in $B$, the direct path from 
\var{Sunscreen} to \var{Dermascare} makes the larger difference.
Since the value added by our explanations is 
measured by the average improvement in answers across versions
$A$ and $B$, it is not unduly influenced by which of the two 
answers participants would be more likely to guess.



\mysubsection{Hypotheses}
\label{section:hypotheses}
Our general hypothesis is that the explanatory condition and question
type will predict participants' performance, and will interact.
Additional specific hypotheses are as follows, where each hypothesis
comprises sub-hypotheses that are tested pairwise.

\begin{enumerate}[H1.]
\myitem
\label{item:H1}
Participants who receive explanations will outperform
participants in the \xcond{Neither} condition for any question type
that requires the information in the explanations (\qtype{Finding},
\qtype{CommonEffect}, \qtype{Path} and \qtype{PathThink}). Comprises twelve
sub-hypotheses: 3 explanation types (\xcond{Verbal}, \xcond{Visual},
\xcond{Both}) $\times$ 4 non-\qtype{Control} question types.

\myitem
\label{item:H2}
Participants who receive explanations will \emph{not} outperform
participants in the \xcond{Neither} condition for the \qtype{Control}
and \qtype{ControlThink} question types, for which the information in
the BN is sufficient. Comprises six sub-hypotheses: 3 explanation types
(\xcond{Verbal}, \xcond{Visual}, \xcond{Both}) $\times$ 2 \qtype{Control}
question types.

\myitem
\label{item:H3}
The effect sizes of \xcond{Visual}, \xcond{Verbal}
and \xcond{Both} explanations on \qtype{PathThink} questions
will be smaller than on \qtype{Path} questions. This is because
\qtype{PathThink} questions require additional inferences that rely on
BN knowledge. Comprises three sub-hypotheses: (Visual, Verbal, Both).


\myitem
\label{item:H4}
The effect size of \xcond{Both} explanations will be larger than
the effects of \xcond{Visual} or \xcond{Verbal} explanations alone. Comprises
eight sub-hypotheses: 2 (\xcond{Both} vs \xcond{Verbal}; \xcond{Both}
vs \xcond{Visual}) $\times$ 4 non-\qtype{Control} question types.
\end{enumerate}

We also compare \xcond{Visual} versus \xcond{Verbal} explanations for
\qtype{Finding}, \qtype{CommonEffect}, \qtype{Path} and \qtype{PathThink}
question types, but we do not formulate any hypotheses for these cases.


\mysubsection{Procedure}
In a Qualtrics survey, participants were first introduced to the experiment
(Figure~\ref{fig:explan_statement}, \ref{appendix:screenshots}), after which 
they completed a demographic questionnaire, and reported their familiarity with
computers and BNs. They then underwent assessments of skills relevant to
understanding BNs: numeracy was measured using \citet{FagerlinEtAl2007}'s 
\define{Subjective Numeracy Scale} (\define{SNS}\/), which correlates well 
with objective mathematical measures of numeracy, and spatial reasoning was
evaluated using the paper-folding task~\cite{EkstromEtAl1976}.
These tests are described in~\ref{appendix:skills}.


Participants then undertook a BN tutorial using the Car Battery BN 
(Figure~\ref{fig:car_intro}, \ref{appendix:BNs}), which included 
comprehension questions.
They had to answer at least 50\% of these questions correctly
in order to proceed to the main part of the experiment.

At this point, the surviving participants were randomly assigned to one of
four explanatory conditions according to the types of explanations they
would see during the actual study -- \xcond{Verbal}, \xcond{Visual},
\xcond{Both} or \xcond{Neither}. They then moved on to the Explanatory
Tools tutorial, where they were trained to interpret and use their
respective explanation types to answer a set of questions (the same for
all conditions), which were similar to questions in the main study. 
Performance on these training tasks was not used for participant selection,
and all participants could then proceed to the main study with the Rats 
and Podunk BNs (sample screenshots appear in Figures \ref{fig:rats_sum}--\ref{fig:mutation_ani}, \ref{appendix:screenshots}).
Throughout the study, in the \xcond{Visual} and \xcond{Both}
explanatory conditions, participants had to trigger the
animations before they were allowed to proceed to the
next screen (sample animation sequences appear in 
Figures \ref{fig:rats_ani} and \ref{fig:mutation_ani}, \ref{appendix:screenshots}).
Our survey concluded with the widely-used \define{System Usability
Scale} (\define{SUS}\/) questionnaire~\cite{Brooke1996,KortumPeres2014}, and
optional questions about participants' views regarding the difficulty of
each section in the survey (Figure~\ref{fig:feedback}, \ref{appendix:screenshots}).

\mysubsection{Participants}
\label{section:participants}
We used the crowd-sourcing platform Prolific Academic, where we recruited
participants from the UK whose primary language is English.

Participants were informed that they would earn \pounds 3 for completing
the Car Battery tutorial, but would need a passing grade on the questions in the tutorial
to continue to the main study --- there they could earn another \pounds 3
plus a bonus of 25p for each correct answer, for a total of up to \pounds
16. Of the initial 190 participants, 55 did not proceed to the main study
for the following reasons: they were not using a supported browser or platform (13), 
dropped out before (13) or during (17) the BN tutorial, or did not achieve 
the required 50\% on the tutorial (12). A further 11 dropped out during the
main study: 3 each for the Both, Verbal and Neither explanatory conditions, 
and 2 for the Visual condition. Table~\ref{tab:participants_section} shows 
descriptive statistics for the 124 surviving participants. 


\input tab-results-participants.tex

The number of participants assigned to each explanatory condition was
roughly equal, with 30--32 participants completing each condition. The
median duration of the experiment was 54 min. with an interquartile
range of 43--78 min; low non-completions (2--3) across conditions
suggest no survivorship bias.
There were no statistically significant differences among participants in
the four explanatory conditions for any demographic factor, SNS score
or $A/B$ BN version, for both Rats and Podunk ($\pvalue>0.05$, Fisher's
exact test for discrete variables and ANOVA for continuous variables).
However, there was one statistically significant difference in the folding
task (\xcond{Visual} $<$ \xcond{Neither}, $\pvalue<0.01$),
and one in the BN tutorial (\xcond{Neither} $<$ \xcond{Both},
$\pvalue<0.05$). Adjustment for these differences did not alter the
results for any hypothesis.

\mysubsection{Statistical model}
\label{section:stat_model}
Our main analysis of performance was via a mixed effects logistic
regression model of the probability that each participant would answer
each question correctly on the Rats and Podunk BNs. Effect sizes
are reported as the average mark difference per question between each
condition and \xcond{Neither}.

The main predictive variables of interest were explanatory condition
and question type. Participant ID and Question ID were included as
random effects, to adjust the predictions for individual variations in
participant ability and question difficulty. The model was also adjusted
for the participants' test score in the BN tutorial, as a fixed effect.
 
Other variables considered were demographic information, scores in both
skill tests and the Explanatory Tools tutorials, and completion time for
the Rats and Podunk sections. However, they were excluded from our model,
as they had no impact.

The effect of explanatory condition on SUS scores was tested separately.
A linear model was used, as SUS scores range from 0--100 and were normally distributed. The model
included an adjustment for the experiment score (\ie combined Rats and
Podunk section scores).

%% file: tab-question-type.tex
\begin{table*}[t]
\setlength{\tabcolsep}{6pt}
\caption{Question types: Number of questions per type, with examples and references to figures showing the corresponding view of the Podunk BN, and pointers to the necessary inferences or information in the \xcond{Verbal} or \xcond{Visual} explanations.\label{tab:qtypes}}
\centering
{\small
    \begin{tabular}{@{\extracolsep{\fill}}p{0.06\linewidth}P{0.01\linewidth}p{0.44\linewidth}p{0.17\linewidth}p{0.16\linewidth}p{0.02\linewidth}}
    \toprule
     \textbf{Type}\hspace*{-1mm} &
     \textbf{\#} &
     \textbf{Question text with answers [\correct{correct} / incorrect]} & \textbf{Verbal information} & \multicolumn{1}{c}{\textbf{Visual information}$\!\!$}
     & $\!\!$\textbf{Fig.}\\
     \midrule
     \qtype{Control} & 4 & According to the story, \guivartab{Jumbo's$\,$Ice$\,$Cream} is sold mainly [with chocolate sprinkles / to children /
     \correct{at the beach}].
& \multicolumn{2}{p{0.32\textwidth}}{none
     needed, since this is the only answer stated in the story}
& ~\ref{fig:podunk_intro}\\
     \midrule
    \qtype{Control Think} & 9 & Suppose we don't have any
    findings. Finding out \guivartab{Bruising} is \guistatetab{not~visible}
    would [increase / not change / \correct{reduce}] the probability of
    \guistatetab{persistent} \guivartab{Peeling}.
& \multicolumn{2}{p{0.32\textwidth}}{none needed, just infer
from the story that no \vartab{Bruising} reduces the probability of
\vartab{Dermascare}, and hence of \statetab{persistent} \vartab{Peeling}}
& ~\ref{fig:podunk_intro}\\
     \midrule
     \qtype{Finding} & 11 & Suppose we already knew three of the findings.
The one additional finding that would most increase the probability that
\guivartab{Dermascare} is \guistatetab{present} is
[\underline{\guivartab{Dunkalot$\,$Sunscreen} was \guistatetab{used}} / \guivartab{Mutation} is
\guistatetab{inherited} / \guivartab{Peeling} is \guistatetab{persistent} / \guivartab{Bruising} is
\guistatetab{not~visible}].
& the first dot point says that \vartab{Dunkalot} \vartab{Sunscreen} was \statetab{used} ``\underline{moderately increases}'' the target probability  & 
\vartab{Dunkalot$\,$Sunscreen} \statetab{used} is a darker green than any other state name & ~\ref{fig:podunk_summary}\\
         \midrule
     \qtype{Common Effect} & 7 & Imagine we didn't know about the
\guivartab{Mutation}. Finding out \guivartab{Peeling} is \guistatetab{persistent}
would [\correct{greatly increase} / moderately increase / slightly increase \dots]
the probability that \guivartab{Dermascare} is \guistatetab{present}.
    & ``greatly increases'' is used about this scenario in the first point
&  \hbox{\statetab{persistent} is dark} \hbox{green in the asso-} ciated animation & ~\ref{fig:podunk_peeling}, \hspace*{-3mm}\ref{fig:mutation_ani_1}\\
     \midrule
     \qtype{Path} & 6 & \guivartab{Dunkalot$\,$Sunscreen} was \guistatetab{used} contributes
due to two connections: by increasing the probability that \guivartab{Podunk$\,$Beach} was 
\guistatetab{visited}, it [\correct{greatly increases} / moderately increases /
slightly increases / \dots] the probability that \guivartab{Dermascare} is
\guistatetab{present}, and\dots
& the first dot point says this path ``greatly increases'' the target probability
 & the arcs on this path are coloured dark green & ~\ref{fig:podunk_sunscreen}\\
     \midrule
     \qtype{Path Think} & 3 & Now that we know all four findings, imagine that
we also find out \guivartab{Jumbo's$\,$Ice$\,$Cream} was \guistatetab{eaten}. This new
finding would [\correct{increase} / not change / reduce] the probability that
\guivartab{Dermascare} is \guistatetab{present}.
& \multicolumn{2}{p{0.32\textwidth}}{same as the line above, and infer from the story that 
finding out \vartab{Ice Cream} is \statetab{eaten} \text{increases} the probability that
\vartab{Podunk Beach} was \statetab{visited}} & ~\ref{fig:podunk_sunscreen}\\
     \bottomrule
    \end{tabular}
}
    \vspace{-3mm}
\end{table*}

%% file: tab-results-participants.tex
\begin{table*}[t]
\setlength{\tabcolsep}{9pt}
    \centering
    \small
    \caption{Descriptive statistics for the four explanatory conditions,
    \xcond{Neither}, \xcond{Verbal}, \xcond{Visual} and \xcond{Both} 
    (number of participants). The specified options have
    the most participants; computing skills and BN experience are self-assessed; 
    mean (standard deviation) and range are given for the SNS and folding task
    scores, and for the post-experiment SUS scores and difficulty ratings (100 participants).\label{tab:participants_section}}
    \begin{tabular}{llcccc}
        \toprule
        \multirow{2}{*}{\textbf{Question}} &  \multirow{2}{*}{\textbf{Option}} & \textbf{Neither} & \textbf{Verbal} & \textbf{Visual} & \textbf{Both} \\
        & & (32) & (31) & (31) & (30) \\
        \midrule
        Gender & Male / Female & 13 / 18  & 18 / 13  & 16 / 15  & 13 / 17 \\
        Age & 18-24 / 25-34 / 35-44 & 17 / 6 / 7  & 14 / 10 / 2  & 13 / 9 / 5  & 16 / 7 / 5 \\
        Nationality & UK / Nigeria & 26 / 3  & 24 / 2  & 21 / 5  & 26 / 1 \\
        Education & Bachelor / Graduate & 19 / 9  & 14 / 12  & 14 / 14  & 16 / 11 \\
        Employment & Full time / Part time & 14 / 5  & 11 / 13  & 17 / 8  & 12 / 9 \\
        Computing skills & Medium / High & 19 / 9  & 16 / 9  & 17 / 7  & 14 / 10 \\
        BN experience & None / Low / Medium & 15 / 10 / 7  & 17 / 6 / 8  & 17 / 9 / 5  & 9 / 12 / 8 \\
        \midrule
        SNS score & Mean (SD) [range: 1-6] & 4.76 (0.76)  & 4.54 (0.87)  & 4.67 (0.66)  & 4.53 (0.98) \\
        Folding task score & Mean (SD) [range: 0-1] & 0.62 (0.23)  & 0.54 (0.26)  & 0.43
(0.2)  & 0.60 (0.24) \\
        \midrule
        SUS score & Mean (SD) [range: 0-100] & 50.23 (20.54) & 57.74 (19.91) & 58.92 (24.44) & 60.25 (19.16) \\
        Difficulty rating & Mean (SD) [range: 1-5] & 3.08 (0.56) & 3.07 (0.65) & 3.09 (0.75) & 2.62 (0.77) \\
        \bottomrule
    \end{tabular}
\end{table*}

%% file: 5_results.tex
We report the results of our hypothesis tests, the SUS questionnaire,
and users' perceptions of section difficulty. \ref{appendix:results} shows
additional results describing the relationships between
experiment duration and marks per section, 
between key experimental variables, 
and between SUS and explanatory condition.

\begin{figure}
     \centering
     \hspace*{-0.1\linewidth}
     \includegraphics[width=0.5\linewidth]{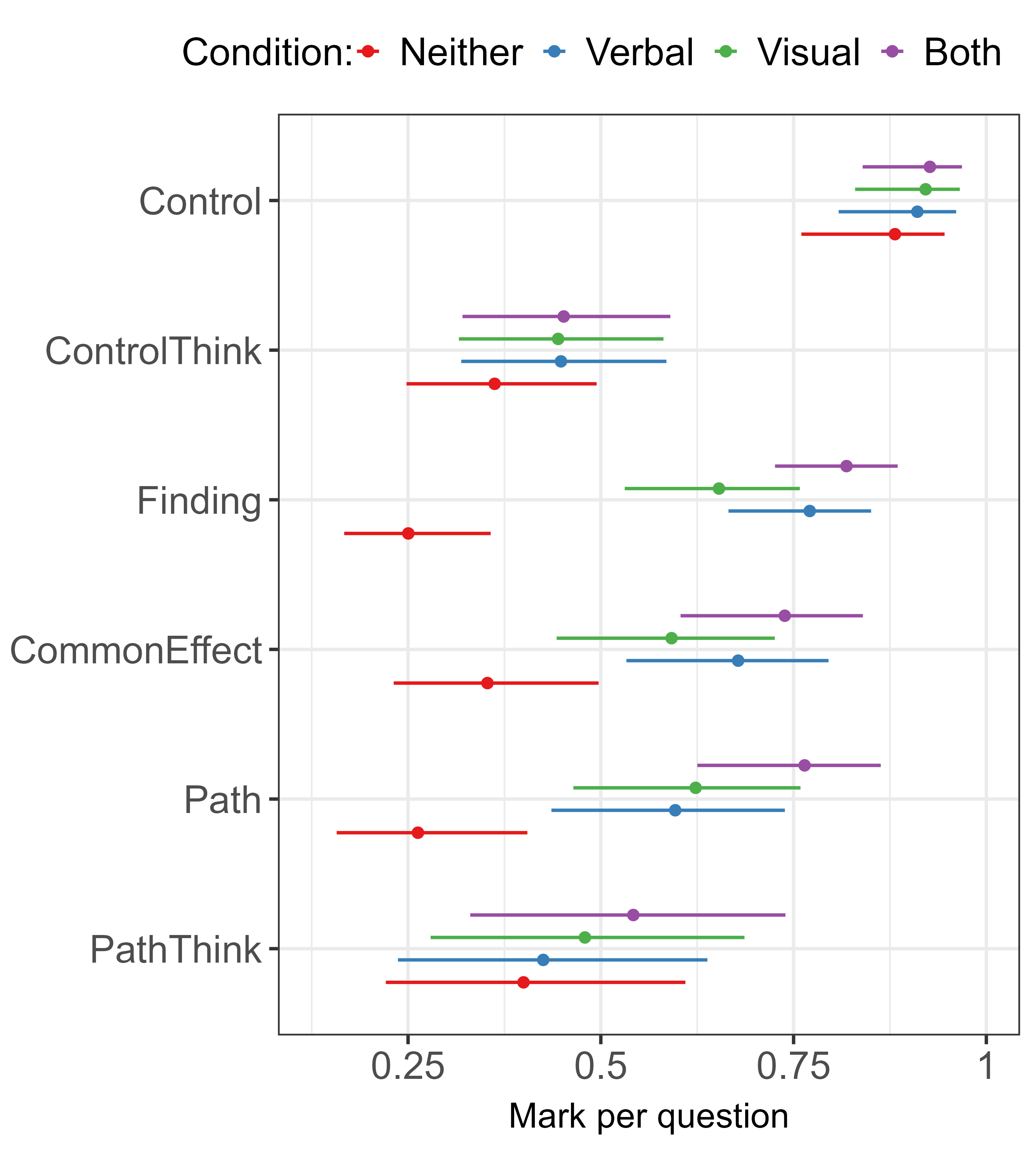}
     \caption{Average question mark by question type and condition.\label{fig:mark_by_qtype_and_cond}}
\end{figure}

\mysubsection{Hypotheses}
\label{section:hypotheses-results}
Figure~\ref{fig:mark_by_qtype_and_cond} shows the average mark ($x$ axis)
for each question type ($y$ axis) and explanatory condition (in different
colours). This average is estimated from the mixed effects model,
after adjusting for the effects of questions, participants and
performance on BN tutorial questions (Section~\ref{section:stat_model}).
Table~\ref{tab:stat_sig} shows the average difference in question marks, odds
ratios and the $\pvalue$s for each pairwise comparison between explanatory
conditions for \qtype{Finding}, \qtype{CommonEffect} and \qtype{Path} questions. 
Comparisons between explanatory conditions for other question types 
(\ie \qtype{Control}, \qtype{ControlThink} and \qtype{PathThink}\/) did not 
yield statistical significance, and have been omitted from the table.

\input tab-statsigV2
The results for the hypotheses in Section~\ref{section:hypotheses} are as follows.
\begin{enumerate}[H1.]
\myitem
\label{item:H1R} 
Participants in the \xcond{Verbal}, \xcond{Visual}
and \xcond{Both} explanatory conditions outperform those in the
\xcond{Neither} condition for \qtype{Finding}, \qtype{CommonEffect}
and \qtype{Path} question types, but not for \qtype{PathThink},
supporting nine of our twelve sub-hypotheses ($\pvalue \ll 0.001$ --
Table~\ref{tab:stat_sig} shows exact statistics).
Effect sizes (average mark difference versus \xcond{Neither}
per question type) ranged from 0.21 to 0.51 for significant results (a correct
answer earns 1 mark).

\myitem
\label{item:H2R}
We did not detect better performance by any of the
explanation conditions over \xcond{Neither} for the \qtype{Control}
or \qtype{ControlThink} question types ($\pvalue > 0.05$), consistent
with all our six sub-hypotheses.


\myitem
\label{item:H3R}
The effect sizes of \xcond{Visual}, \xcond{Verbal} and
\xcond{Both} explanations on \qtype{PathThink} questions were smaller
than on \qtype{Path} questions, supporting all our three sub-hypotheses
($\pvalue < 0.01$).


\myitem
\label{item:H4R}
The effect size of \xcond{Both} explanations was larger than
that of \xcond{Visual} ones only for \qtype{Finding} questions, and
larger than that of \xcond{Verbal} explanations only for \qtype{Path}
questions ($\pvalue\! <\!0.05$), supporting two out of eight sub-hypotheses.
Also, there are trends indicating that participants in the \xcond{Both}
condition outperform those in the \xcond{Visual} condition for
\qtype{CommonEffect} and \qtype{Path} questions ($\pvalue<0.05$ prior
to adjustment for multiple comparisons).

\end{enumerate}

Another trend is that \xcond{Verbal}-condition participants
outperformed \xcond{Visual}-condition ones for \qtype{Finding} questions
($\pvalue<0.05$ prior to adjustment).

\mysubsection{System Usability Scale (SUS)}
\label{section:SUS}
The average SUS scores for the conditions with explanations were similar
(57--60), while the average score for the \xcond{Neither} condition was
only 50 (Table~\ref{tab:participants_section}) --
a statistically significant difference ($\pvalue < 0.05$). Participants'
overall marks for the Rats and Podunk BNs are moderately positively correlated with
SUS score (Pearson $r=0.41$, $\pvalue\ll 0.001$).
We performed a mediation analysis (\ref{appendix:results}) that 
demonstrated no \emph{direct} effect of explanatory condition on SUS scores, 
but identified an indirect effect via participants' performance on the Rats 
and Podunk BNs. Notably, this effect exists despite participants not being 
informed of their marks, suggesting that participants' positive rating of the 
usability of an explanatory condition's tool depended only on the extent that 
it improved their actual performance.

\mysubsection{Perceptions of section difficulty}
Of the 124 participants, 100 completed the optional questions
on difficulty, with no detected completion bias across conditions
($\pvalue>0.1$) --- Table~\ref{tab:difficulty} shows
details of difficulty ratings across conditions. The Background section, which
includes demographic and familiarity questions and skill assessments,
was rated easiest overall, by between 0.7 and 1.4 points ($\pvalue <
0.001$). The Explanatory
Tools section was rated next easiest (easier than the BN tutorial, Rats
and Podunk sections) by between 0.4 and 0.7 points ($\pvalue < 0.05$).
Comparing the different explanatory conditions, there were no statistically 
significant differences in participant ratings of difficulty for either the 
BN tutorial or the Explanatory Tool tutorial. However,
participants in the \xcond{Neither} condition deemed the Rats and Podunk BNs
to be significantly harder than participants in the \xcond{Both} condition
($\pvalue <0.05$). 





\input tab-results-difficulty

%% file: tab-statsigV2.tex
\begin{table}[t]
\setlength{\tabcolsep}{8.5pt}
\caption{Average difference in mark per question (mark diff), odds
ratio, standard error and statistical significance for pairwise comparisons of
explanatory conditions across three question types (after adjustment for multiple comparisons): {\qtype{Finding}},
{\qtype{CommonEffect}} and {\qtype{Path}}. All the tests are one sided
per hypotheses H1--H4, except for Verbal vs Visual ($\dagger$),
where no hypothesis was formulated. A $\pvalue$ in italics indicates
the result was significant ($\pvalue < 0.05$) prior to adjustment
for multiple pairwise comparisons.  The other question types did
not yield statistically significant differences for any explanatory
condition.\label{tab:stat_sig}}

\centering
\small
\begin{tabular}{llrrrrrr}
\toprule
\multirow{2}{*}{\textbf{Question type}} & \multirow{2}{*}{\textbf{Comparison}} & {\textbf{Both vs}} & {\textbf{Verbal vs}} & {\textbf{Visual vs}} & {\textbf{Both vs}} & {\textbf{Both vs}} & {\textbf{$\dagger$ Verbal vs}}  \\
 & & {\textbf{Neither}} & {\textbf{Neither~}} & {\textbf{Neither~}} & {\textbf{Visual~}} & {\textbf{Verbal~}} & {\textbf{Visual~~}} \\ 
\midrule
\multirow{5}{*}{\qtype{Finding}} & Mark diff & 0.51 & 0.46 & 0.35 & 0.15 & 0.05  & 0.11\\
              & Odds ratio & 13.5 & 10.1 &  5.6 & 2.14 & 1.3 & 1.8\\
              & Std error & 3.70 & 2.68 & 1.47 & 0.64 & 0.37 & 0.48\\
              & $\pvalue$ & 0.000 & 0.000 & 0.000 & 0.003 & 0.613 & \textit{0.165}\\
              &           &  ***  & ***   & ***   & ** &  & \\
\midrule
\multirow{5}{*}{\qtype{CommonEffect}} & Mark diff & 0.34 & 0.29 & 0.21 & 0.13 & 0.06 & 0.08\\
               & Odds ratio & 5.2 & 3.9 & 2.7 & 1.9 & 1.3 & 1.5 \\
              & Std error & 1.52 & 1.10 & 0.75 & 0.57 & 0.40 & 0.42 \\
              & $\pvalue$ & 0.000 & 0.000 & 0.002 & \textit{0.066} & 0.661 & 0.736\\
              &           &  ***  & ***   & **   & $\cdot$ & & \\
\midrule
\multirow{5}{*}{\qtype{Path}} & Mark diff & 0.43 & 0.28 & 0.31 & 0.13 & 0.15 & $-0.02$\\
                          & Odds ratio & 9.1 & 4.1 & 4.6 & 2.0 & 2.2 & 0.9 \\
                          & Std error & 2.88 & 1.26 & 1.41 & 0.61 & 0.69 & 0.27 \\
                          & $\pvalue$ & 0.000 & 0.000 & 0.000 & \textit{0.087} & 0.038 & 0.999\\
              &           &  ***  & ***   & ***   & $\cdot$ & * & \\
\bottomrule
\end{tabular}
\end{table}

%% file: tab-results-difficulty.tex
\begin{table}[t]
\setlength{\tabcolsep}{9pt}
    \centering
    \small
    \caption{Mean (standard deviation) of difficulty ratings (range: 1--5) by survey section and explanatory condition \label{tab:difficulty}}
    \begin{tabular}{lllll}
        \toprule
        \textbf{Section} & \textbf{Neither} & \textbf{Verbal} & \textbf{Visual} & \textbf{Both} \\ 
        \midrule
        Background & 2.1 (1.0) & 2.2 (1.1) & 2.0 (1.0) & 1.9 (1.1) \\ 
        BN tutorial & 3.0 (1.0) & 3.4 (0.8) & 3.3 (1.1) & 3.0 (1.1) \\ 
        Explanatory tools & 2.8 (0.9) & 2.7 (1.0) & 3.1 (1.1) & 2.4 (1.0) \\ 
        Rats & 3.6 (0.8) & 3.5 (1.0) & 3.3 (1.0) & 2.8 (1.1) \\ 
        Podunk & 3.9 (0.8) & 3.5 (1.3) & 3.6 (0.9) & 3.0 (1.0) \\ 
        \bottomrule
    \end{tabular}
\end{table}

%% file: 6_llm.tex


Our aim in this paper was to introduce designs for verbal and visual 
explanations that work well for all basic BN structures. These designs 
are organised around shared explanatory concepts such as `evidence 
contributions', `pathway contributions', `common effects' and `target 
nodes', with the verbal designs being implemented by 
templates.
However, as noted in~\cite{Hennessy2020ExplainingBN}, template 
implementations of verbal explanations do not always produce natural sounding
explanations when scaled up to large BNs or ported to different domains.

Large Language Models (LLMs) appear to offer a promising approach for 
generating explanations for larger BNs. In this work, we validate their
use on our current BNs as a first step. The rationale is that if they 
perform well on smaller BNs, their applicability to larger BNs warrants 
further investigation; conversely, poor performance on smaller BNs would 
raise concerns about their suitability for more complex networks.

We consider two new research questions:
\begin{enumerate}[RQ1.]
\myitem
\label{item:RQ1}
Do LLMs generate explanations with a standardised structure that is consistent
with our general design and explanatory concepts, given minimal prompts?

\myitem
\label{item:RQ2}
If not, can LLMs do so when they are given explicit instructions?
\end{enumerate}


To investigate~RQ\ref{item:RQ1}, we assess the output of GPT-5.2
from multiple runs given a \define{\directquery}, which asks it to 
explain the role of evidence in changing the probability of a 
particular state in a BN. This prompt does not specify \emph{how}
the explanation should be generated, just \emph{what} is to be explained. 
The format of the \directquery\ for summary explanations is specified
in the left-hand column of Table~\ref{fig:direct-guidedquery}.

To investigate RQ\ref{item:RQ2}, we assess the output of an LLM from 
multiple runs given a \define{\guidedquery} that contains specific
instructions about how to format the output, in addition to the 
explanation request. 
These instructions were developed and refined iteratively.
The format of the \guidedquery\ for summary explanations appears
in the right-hand column of Table~\ref{fig:direct-guidedquery}.

\input tab-direct-guidedquery.tex

We tested the two types of prompts on four of the explanatory tasks 
from our user study:
Rats BN -- \qtype{Summary}, Podunk BN -- \qtype{Summary}, Rats BN --
\qtype{Path} and Podunk BN -- \qtype{Common Effect}. The prompts for
each task appear in~\ref{section:LLM_prompts}, and two sample
outputs for each task and prompt, together with our user-study texts, 
appear in~\ref{section:llm_scenarios}.
To minimise contamination, the domain was obfuscated by converting 
variable names to initials. The effectiveness of this 
obfuscation was informally tested by asking the LLM to identify the 
domain only from the details provided in the prompt for two of the explanatory
tasks, viz. Rats BN -- \qtype{Path} and Podunk BN -- \qtype{Common Effect}, which
are the most detailed tasks. The LLM failed to identify the domain
in two separate attempts for both explanatory tasks.

Each explanatory task was performed fifteen times using each type of prompt. 
For each task, we tested the effectiveness of the instructions, and the 
stability of the generated explanations. Instruction effectiveness was 
tested by computing the similarity of the generated explanations to the 
reference explanation used in the user study; and explanation stability 
was tested by computing the pairwise similarity of the generated 
explanations. For both tests, we employed three metrics, 
BLEU~\cite{PapineniEtAl2002Bleu}, METEOR~\cite{BanerjeeLavie2005Meteor} 
and BERTscore~\cite{ZhangEtAl2019Bertscore}, using the HuggingFace 
library implementations ({\small \url{https://huggingface.co}}). BLEU was
configured to use n-grams up to size 2 to handle short passages; METEOR 
was left at its default values; and BERTscore was set to English, and used
the standard rescaling of~\cite{ZhangEtAl2019Bertscore}.
It is worth noting that BLEU and METEOR were originally developed for
machine translation, whereas BERTscore measures semantic similarity 
between two texts (the scores obtained by these metrics are described
in~\ref{section:metric_interpretations}). Thus, while these metrics 
support relative comparisons between methods, the absolute values of 
BLEU and METEOR do not necessarily carry the same interpretation 
as in machine translation.

\input tab-llm-output-similarity-tests_v2

Table~\ref{tab:llm-similarities} summarises the similarity results
across the tested combinations, showing the mean and standard 
deviation for each metric in the fifteen runs. 
Table~\ref{tab:llm-similarities_a}(a) displays the similarity between 
each user-study explanation and the LLM-generated explanations for
the direct and the \guidedquery, and Table~\ref{tab:llm-similarities_b}(b)
shows the pairwise similarity between the LLM-generated explanations 
for the \directquery, and those for the \guidedquery. Overall, the pairwise 
similarities of the LLM-generated explanations to each other are 
notably higher than their similarities to the user-study explanations.
In addition, similarities (either to the user-study explanations or 
among the generated texts) are higher for the explanations generated 
by the \guidedquery\ than for those produced by the \directquery\ 
($\pvalue < 0.001$ for all differences, except the smallest two 
(italicised), where $\pvalue < 0.01$). Table~\ref{tab:llm-analysis}
characterises the main differences between sample explanations
generated by the direct and the guided prompts and the corresponding 
user-study explanations (\ref{section:llm_scenarios}). A noteworthy
phenomenon is that style can be highly sensitive to changes in BNs 
that should not be relevant to stylistic choices, \eg seven of
the fifteen summary explanations generated for the Rats BN with the 
\directquery\ contain ``**'' in the raw Markdown format used by the 
LLM (which indicates boldfaced text), while none of the fifteen summary
explanations generated for the Podunk BN contain ``**''. Similarly, all
the fifteen summary explanations generated with the \guidedquery\ for 
the Rats BN contain ``**'', while only four of the fifteen summary 
explanations generated for the Podunk BN contain ``**''.
In contrast, the (verbal) content generated with guided prompts is fairly 
stable, compared to the content generated with direct prompts.

\myparagraph{Similarity between LLM-generated explanations and 
user-study explanations}
Examining two sample outputs from the \directquery\ and two from the
\guidedquery\ in the four tasks (\ref{section:llm_scenarios}), we see 
a greater similarity to the user-study explanations when guidance is 
provided.

Table~\ref{tab:llm-similarities}(a) shows the similarity between the 
user-study explanations and the explanations generated by the direct 
and the \guidedquery.
Our results indicate that providing instructions (\ie \guidedquery\/) greatly 
increases the similarity for both \qtype{Summary} explanations, and to 
a lesser extent for the \qtype{Path} and \qtype{Common effect} explanations. 
Specifically, the BLEU scores of 0.1--0.3 achieved by direct prompts
would be considered poor to very poor, whereas the higher scores of 
0.4--0.5 for guided prompts would be considered high 
quality~\cite{google_bleu_docs}. METEOR and BERTscore suggest similar
levels of quality and relative improvement between direct and guided 
prompts.

Guidance had less effect for the longer and more structured explanations, 
as shown
by the smaller, yet statistically
significant, differences between the texts generated by the guided and 
the \directquery\ for Rats BN -- \qtype{Path} and Podunk BN -- \qtype{Common Effect} 
across all the metrics.  

\input tab-llm_analysis

\myparagraph{Pairwise similarity between LLM-generated explanations}
Table~\ref{tab:llm-similarities}(b) shows the average pairwise similarity
of the generated texts to each other for the same task and prompt type. 
For the \directquery, overall similarities begin in the range 0.4--0.5, 
suggesting the generated texts are similar enough in style and content 
to be considered good renditions of each other, even without specific
instructions. However, the average pairwise similarity is lower overall
for the \qtype{Common effect} explanation.

What is most evident in Table~\ref{tab:llm-similarities}(b) is that the 
instructions in the \guidedquery\ produce much stronger similarities 
between the generated texts than when instructions are absent, with the 
exception of the \qtype{Path} explanations, where the metric differences 
are smaller. With guidance, the metrics reach the top of their effective 
ranges (0.8--0.9) for the two \qtype{Summary} explanations, while they 
are slightly lower for the Podunk BN -- \qtype{Common effect} explanations,
and substantially lower for the Rats BN -- \qtype{Path} explanations.

\vspace*{2mm}
These brief investigations show that even for our simple BN explanatory 
tasks, LLM explanations generated by direct prompts exhibit wide 
variability, while those generated by guided prompts are less
variable. Notably, both types of explanations occasionally provide
erroneous information.

It is worth noting that some user groups may prefer the novelty of 
varied expression afforded by LLMs to a repetitive format. In
contrast, specific populations, such as blind and low vision users, 
prefer low variability for the same application~\cite{ReindersEtAl2026}.
Notwithstanding the target population, consistency in the terminology 
and structure of BN explanations is preferable for applications where 
inconsistent interpretations can lead to serious mistakes~\cite{StarmerEtAl2023}.
For example, traditional narrative radiology reports are associated with
excessive variability in language and style, reducing clarity, and making
it difficult for clinicians to identify key information needed for patient 
care~\cite{GaneshanEtAl2018}. Difficulties like these can be alleviated by
adopting standardised probabilistic terms~\cite{RishniwFreeman2023} (as 
required in US intelligence analysis~\cite{ICD-203-2015}) and report 
structures~\cite{SheaTowbin2019} --- practices that we have incorporated into 
our templates and the \guidedquery{}s. Nonetheless, as our exploratory results show, 
explanations generated by LLMs may deviate from their instructions, producing
the kind of variability and potentially consequential misinformation that we seek
to avoid.

%% file: tab-direct-guidedquery.tex
\begin{table}[!t]
\centering
\caption{Format of the direct and guided prompts for summary explanations. 
Elements enclosed in double braces are replaced before submitting the prompt to
the LLM.\label{fig:direct-guidedquery}}
\begin{tabular}{p{0.472\textwidth}p{0.472\textwidth}}
\toprule
\multicolumn{1}{c}{\textbf{\small Direct prompt}} & \multicolumn{1}{c}{\textbf{\small Guided prompt}}\\
\midrule
{\scriptsize
\begin{minipage}[t]{0.46\textwidth}
{
\begin{lstlisting}[breaklines=true,basicstyle=\ttfamily,keepspaces=true,breakindent=0pt]
We are talking about this causal Bayesian network and set of findings:

BN: {{JSON-encoded BN}}
Evidence: {{JSON-encoded evidence}}

Target: "{{target node name}}" is "{{target state name}}"

Summarise in ordinary language how the findings contribute to the probability that <Target>.

Instructions
------------
Use no more than {{wordcount}} words.
\end{lstlisting}
}
\end{minipage}
}
&
{\scriptsize
\begin{minipage}[t]{0.46\textwidth}
{
\begin{lstlisting}[breaklines=true,basicstyle=\ttfamily,keepspaces=true,breakindent=0pt]
We are talking about this causal Bayesian network and set of findings:

BN: {{JSON-encoded BN}}
Evidence: {{JSON-encoded evidence}}

Target: "{{target node name}}" is "{{target state name}}"

Summarise in ordinary language how the findings contribute to the probability that <Target>.

Instructions
------------
Use no more than {{wordcount}} words.

A "contribution" is defined as how much difference learning a finding (or findings) makes to the probability of a <Target> when all the other findings are already known.

Describe probability contributions only according to the following scale:
    greatly increases = 31% - 100%
    moderately increases = 16% - 30%
    slightly increases = 1% - 15%
    doesn't change = 0%
    slightly reduces = -1% - -15%
    moderately reduces = -16% - -30%
    greatly reduces = -31% - -100%

First, describe the contribution all the findings make together.

Second, list with dot points each finding's contribution (in order of decreasing contribution magnitude). For each finding, if not knowing another finding would change its contribution by at least 10%, then mention the other finding in this dot point.

Avoid pronouns if they're ambiguous.

Refer to findings with phrasing like "the finding that variable is state" or "finding out that variable is state".

Do not mention any other variables or any causal connections.
\end{lstlisting}
}
\end{minipage}
}\\
\bottomrule
\end{tabular}
\end{table}

%% file: tab-llm-output-similarity-tests_v2.tex
\begin{table}[t]
    \caption{Similarity of LLM responses to (a)~the user study explanation 
    and (b)~each other, by explanatory task, prompt type and scoring metric. 
    For each of the four explanatory tasks, response texts were generated by 
    fifteen runs of GPT~5.2 (with high reasoning and tool use) given the \directquery\, 
    and fifteen runs given the \guidedquery. Text similarities were measured by the 
    mean (SD) of BLEU, METEOR and BERTscore. On the left-hand side~(a), each 
    response was compared with the corresponding user-study reference explanation. 
    On the right-hand side~(b), each pair of \directquery\ responses was compared, 
    and each pair of \guidedquery\ responses was compared. 
    Differences between the performance of the \directquery\ 
    and \guidedquery\ are shown in each case, where \textbf{bold} 
    indicates significance with $\pvalue < 0.001$, and \textit{italics}
    indicates significance with $\pvalue < 0.01$. Metric scores
    are colour coded based on the gradient from~\cite{lavie2011mtevaluation}, but  
    with shades along the orange--blue spectrum indicating distances from 
    a minimum score of 0 and from a metric-specific maximum score 
    (0.8 for BLEU, 0.9 for METEOR and 0.82 for BERTscore respectively) --- we changed the
    colours to accommodate visual deficiencies in the green-red spectrum.\label{tab:llm-similarities}}
\setlength{\tabcolsep}{8pt}
\centering
\small
\begin{tabular}{lllllclll}
\toprule
\multirow{2}{5em}{\textbf{Explanatory task}} & \multirow{2}{2em}{\textbf{Prompt type}} & \multicolumn{3}{c}{\textbf{(a) Mean (SD) similarity to user study \label{tab:llm-similarities_a}}} & & \multicolumn{3}{c}{\textbf{(b) Mean (SD) pairwise self-similarity \label{tab:llm-similarities_b}}}\\
\cmidrule(lr){3-5} \cmidrule(lr){7-9}
& & BLEU & METEOR & BERTscore & & BLEU & METEOR & BERTscore \\
\midrule
\multirow{3}{4em}{Rats -- \qtype{Summary}}
& Direct & \colbleu{0.21} (0.03) &  \colmeteor{0.24} (0.05) &  \colbert{0.17} (0.03) & &  \colbleu{0.42} (0.10) &  \colmeteor{0.49} (0.12) &  \colbert{0.53} (0.09) \\
& Guided & \colbleu{0.47} (0.06) &  \colmeteor{0.56} (0.07) &  \colbert{0.51} (0.05) & &  \colbleu{0.76} (0.11) &  \colmeteor{0.84} (0.13) &  \colbert{0.82} (0.09) \\
& \textbf{Diff} & \textbf{+0.26} & \textbf{+0.32} & \textbf{+0.35} & & \textbf{+0.34} & \textbf{+0.35} &  \textbf{+0.29} \\
\midrule
\multirow{3}{4em}{Podunk -- \qtype{Summary}}
& Direct & \colbleu{0.14} (0.02) &  \colmeteor{0.23} (0.03) &  \colbert{0.17} (0.04) & &  \colbleu{0.41} (0.06) &  \colmeteor{0.43} (0.08) &  \colbert{0.41} (0.07) \\
& Guided & \colbleu{0.43} (0.04) &  \colmeteor{0.55} (0.03) &  \colbert{0.55} (0.03) & &  \colbleu{0.77} (0.08) &  \colmeteor{0.80} (0.10) &  \colbert{0.79} (0.07) \\
& \textbf{Diff} & \textbf{+0.28} & \textbf{+0.32} & \textbf{+0.39} & & \textbf{+0.36} & \textbf{+0.37} &  \textbf{+0.38} \\
\midrule
\multirow{3}{4em}{Rats -- \qtype{Path}}
& Direct & \colbleu{0.16} (0.03) &  \colmeteor{0.26} (0.05) &  \colbert{0.17} (0.03) & &  \colbleu{0.41} (0.06) &  \colmeteor{0.38} (0.07) &  \colbert{0.43} (0.06) \\
& Guided & \colbleu{0.27} (0.06) &  \colmeteor{0.52} (0.07) &  \colbert{0.33} (0.06) & &  \colbleu{0.51} (0.11) &  \colmeteor{0.59} (0.11) &  \colbert{0.52} (0.09) \\
& \textbf{Diff} & \textbf{+0.11} & \textbf{+0.26} & \textbf{+0.16} & & \textit{+0.10} & \textbf{+0.21} &  \textit{+0.09} \\
\midrule
\multirow{3}{4em}{Podunk -- \qtype{Common Effect}}
& Direct & \colbleu{0.16} (0.02) &  \colmeteor{0.26} (0.02) &  \colbert{0.23} (0.04) & &  \colbleu{0.33} (0.05) &  \colmeteor{0.39} (0.06) &  \colbert{0.34} (0.06) \\
& Guided & \colbleu{0.30} (0.02) &  \colmeteor{0.40} (0.03) &  \colbert{0.43} (0.02) & &  \colbleu{0.74} (0.05) &  \colmeteor{0.69} (0.07) &  \colbert{0.71} (0.06) \\
& \textbf{Diff} & \textbf{+0.14} & \textbf{+0.14} & \textbf{+0.20} & & \textbf{+0.40} & \textbf{+0.30} &  \textbf{+0.37} \\ 
\bottomrule
\end{tabular}
\end{table}

%% file: tab-llm_analysis.tex
\begin{table}
\caption{Analysis of two sample texts generated by the direct and the guided prompts.\label{tab:llm-analysis}}
\centering
{\small 
\begin{tabular}{p{16mm}p{0.41\textwidth}p{0.41\textwidth}}
\toprule
 & \multicolumn{1}{c}{\textbf{\small Direct prompt}} & \multicolumn{1}{c}{\textbf{\small Guided prompt}}\\
\midrule
\multirow{5}{*}{General} & $\bullet$ The decision to boldface text is arbitrary. & $\bullet$ The decision to boldface text is arbitrary.\\
 & $\bullet$ The choice to break lines is arbitrary. & $\bullet$ The explanations have bullet points.\\ 
& $\bullet$ There is a large variation in probabilistic terms, \eg odds, ``argues against'', ``points towards'', ``are more common'', ``less diagnostic''. & $\bullet$ The probabilistic terms follow the contribution scale.\\ 
\midrule
\hspace*{-1.7mm}\multirow{3}{*}{\begin{tabular}[c]{@{}l@{}}Rats --\\ \qtype{Summary}\\ \ref{section:LLM_rats_summary}\end{tabular}}  & 
$\bullet$ The first discussion point concerns an absolute probability, and the second concerns relative probabilities. & 
$\bullet$ The explanations match the user-study text.\\ 
 & $\bullet$ The second summary has numerical probabilities. & \\
\midrule
\hspace*{-1.7mm}\multirow{6}{*}{\begin{tabular}[c]{@{}l@{}}Podunk --\\ \qtype{Summary}\\ \ref{section:LLM_podunk_summary}\end{tabular}} & 
$\bullet$ The reasoning between \varft{Dermascare}, \varft{Peeling} \stateft{persistent} 
and \varft{Mutation} \stateft{inherited} is wrong. & $\bullet$ The explanations largely match the user-study text.\\
 & $\bullet$ The effect of \varft{Dunkalot$\,$Sunscreen} \stateft{used} is different in the two summaries: ``much more likely'' in the first versus the correct ``more likely'' in the second. &
$\bullet$ The effect of not knowing that \varft{Peeling} is \stateft{persistent} in light of the
\varft{Mutation} being \stateft{inherited} is different in the two explanations: ``much smaller effect'' in the first
versus the correct ``no effect'' in the second.\\ 
\midrule
\hspace*{-1.7mm}\multirow{7}{*}{\begin{tabular}[c]{@{}l@{}}Rats -- \qtype{Path}\\ \ref{section:LLM_rats_path}\end{tabular}} & 
$\bullet$ Both explanations mention \varft{Squeaking}, which is not required. & $\bullet$ The explanations largely match the user-study text. \\
 & $\bullet$ The first explanation does not mention the direct connection between \varft{Ecstazine} and \varft{Social$\,$Activity}. &
 $\bullet$ The phrase ``after accounting for the change in \varft{Neurofil}'' in the first explanation is imprecise.\\
& $\bullet$ The second explanation first treats \varft{Neurofil} as a separate variable, not related to \varft{Ecstazine}, which is incorrect; and then as a variable related to \varft{Ecstazine}. & 
$\bullet$ The second explanation gives numerical probabilities.\\
\midrule
\hspace*{-1.7mm}\multirow{5}{*}{\begin{tabular}[c]{@{}l@{}}Podunk --\\ \qtype{Common}\\ \qtype{effect}\\ \ref{section:LLM_podunk_collider}\end{tabular}} &
$\bullet$ The first explanation includes information about \varft{Bruising}, which is not required. & $\bullet$ The first explanation matches the user-study text, but gives numerical probabilities.\\
& $\bullet$ The second explanation matches the user-study text. & $\bullet$ The second explanation is unclear and dissimilar from the user-study text --- crucially, the explain-away relation is not conveyed.\\
\bottomrule 
\end{tabular}
}
\end{table}

%% file: 7_discussion.tex
From a BN perspective, this study is limited by (i)~using only a
few handmade, binary and small BNs; (ii)~removing the normal BN UI
functionality of adding findings or inspecting CPTs; and (iii)~recruiting
mainly participants with no prior BN experience (and providing them
with limited training). The third item is related to a common problem
with NLG evaluations --- using crowd-workers, instead of people who are
engaged with BNs.

Our explanatory approach incorporates novel visual aspects, such as
animations, and novel verbal aspects, such as counterfactual scenarios,
as well as combining visual and verbal modalities. In addition, our
evaluation focuses on task performance, which has not been done in previous
BN studies. Our results show that participants in the
\xcond{Verbal}, \xcond{Visual} and \xcond{Both} explanatory conditions
outperform those in the \xcond{Neither} condition for three key question
types -- \qtype{Finding}, \qtype{CommonEffect} and \qtype{Path}. However,
the limited training was insufficient for our participants to perform well in \qtype{PathThink}
questions under any explanatory condition. Still, our results are
promising enough to warrant further investigation.




Since our questions were formulated verbally, users can quite directly
derive answers from verbal explanations, which gives them a potential
advantage over users who saw only visual explanations. Hence, it is noteworthy
that participants in the \xcond{Visual} explanatory condition did \emph{not} 
perform much worse than those in the \xcond{Verbal} condition. 
In addition to the lack of BN experience, which is shared by all our participants, the slightly worse performance of those in the Visual condition may be due to their
relatively poor paper-folding performance (Section~\ref{section:participants}). 
Hence, we posit that it is worth considering different types of cohorts, \eg
participants with more familiarity with BNs, as well as tailoring explanation
types to users' abilities. 



Our current system is well suited for explaining fragments of a BN, as it 
explains the impact of every arc in a path, and shows counterfactual scenarios
to clarify common-effect interactions. There are important applications for
such small scale explanations, \eg teaching contexts and representing human 
mental models of causal systems~\cite{PilditchEtAl2025}. However, when applied
step-by-step to an entire BN, these fragment-based explanations may yield lengthy 
verbal explanations, and the visual explanations (with animations) may also become
unwieldy.
It is therefore worth investigating how to adapt our methods to larger explanatory 
tasks, \eg by omitting any variables or paths that make only minor contributions,
focusing on global impacts of variables, and further exploring the use of LLMs to
generate, adapt or summarise long explanations.

Finally, \citet{Hennessy2020ExplainingBN} note the disadvantages of
template-based explanation generation, and LLMs seem a natural next step for 
realising verbal explanations. However, our preliminary exploration of this
idea indicates that it is difficult to guarantee the quality, consistency and
accuracy of explanations produced by an LLM (Section~\ref{section:LLM}) --- a 
problem that may be mitigated with very specific and forceful prompts, possibly 
at the expense of text variability.

%% file: AppendixA-BNs.tex
\begin{figure}[ht]
\centering
\begin{subfigure}[b]{0.48\textwidth}
    \begin{minipage}{\textwidth}
        \centering
        \includegraphics[width=\textwidth]{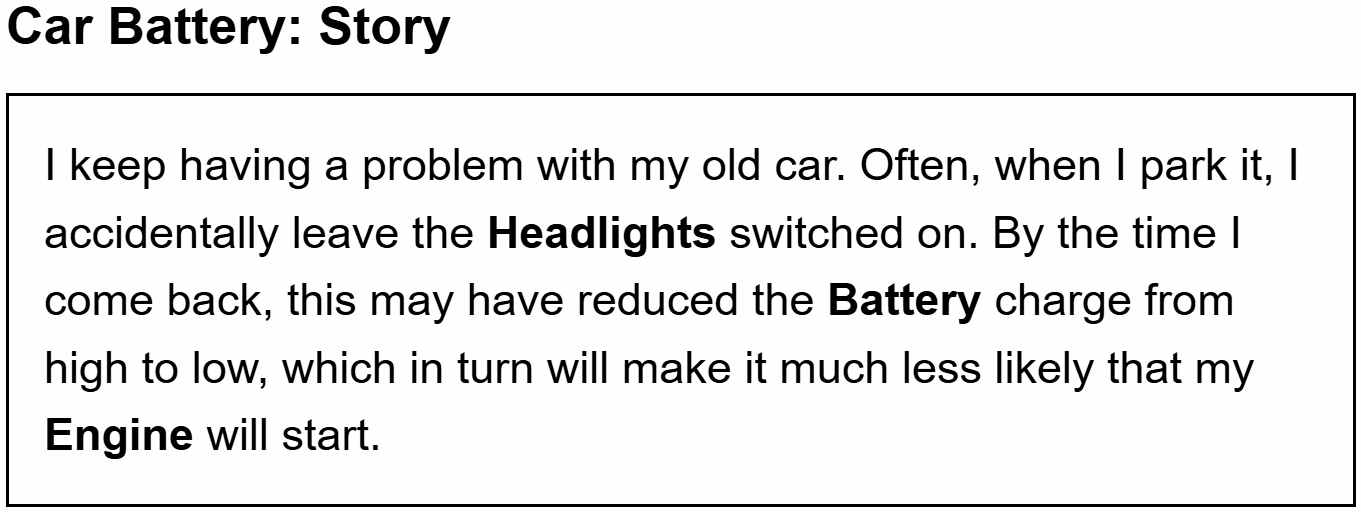}
        \vspace*{0.28\textwidth}
    \end{minipage}
    \caption{Car Battery story\label{fig:car_story}}
\end{subfigure}
    \hfill
\begin{subfigure}[b]{0.48\textwidth}
   \includegraphics[width=\textwidth]{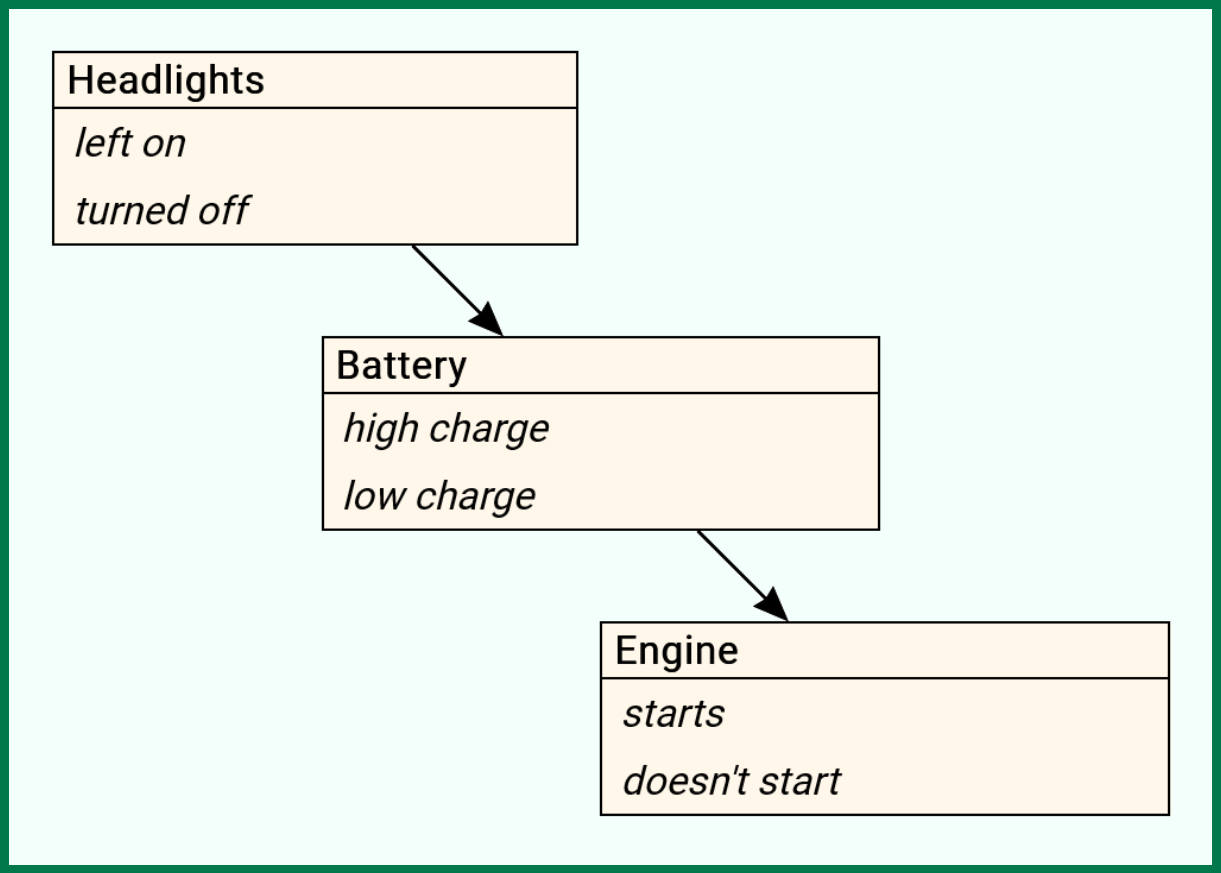}
    \caption{Car Battery BN structure\label{fig:car_base}}
\end{subfigure}
\caption{Car Battery: a simple example used in the BN tutorial.\label{fig:car_intro}}
\end{figure}
\begin{figure}[h]
\centering
\begin{subfigure}[b]{0.48\textwidth}
    \includegraphics[width=\textwidth]{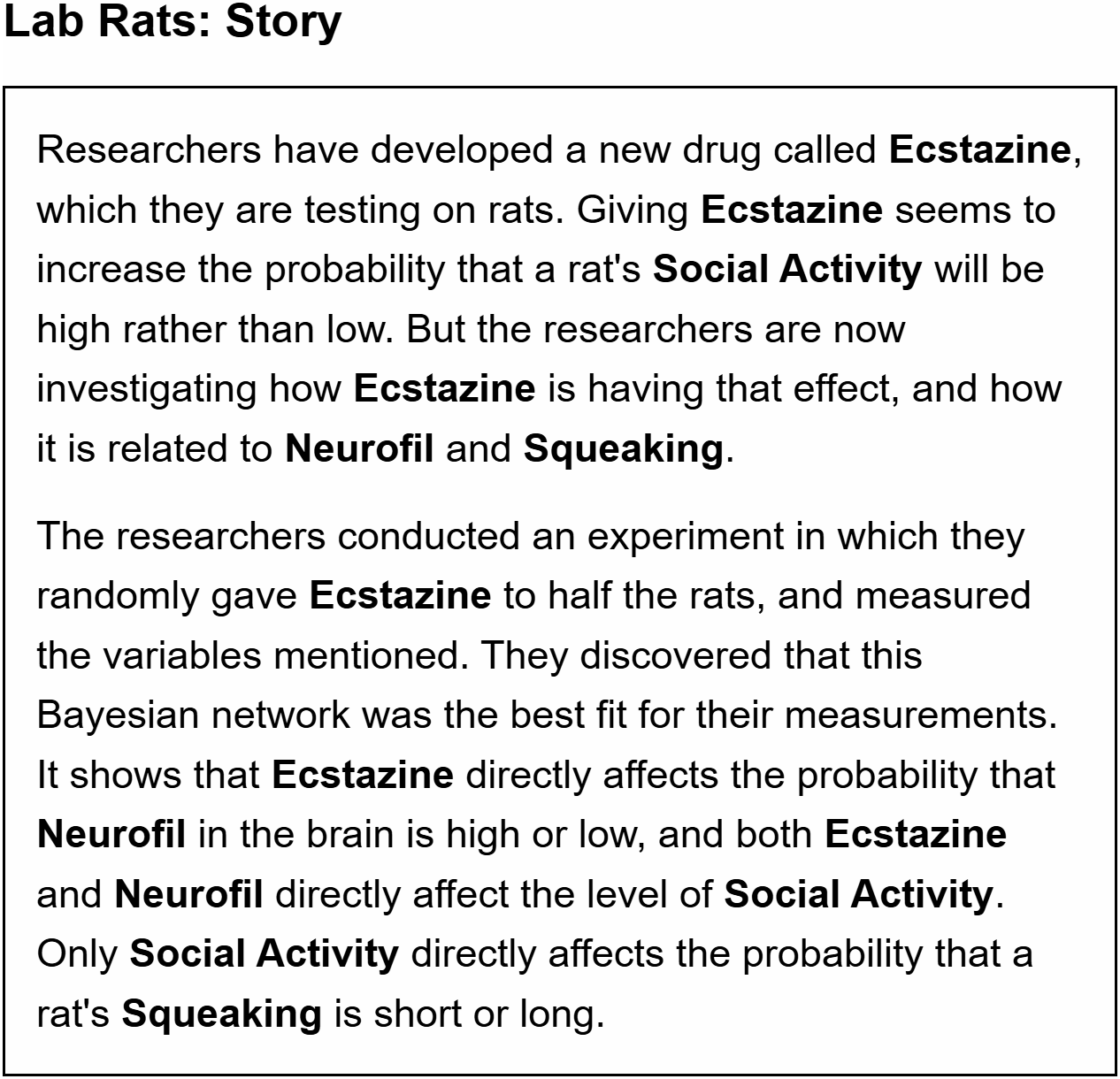}
    \caption{Rats story\label{fig:rats_story}}
    \end{subfigure}
    \hfill
\begin{subfigure}[b]{0.48\textwidth}
   \includegraphics[width=\textwidth]{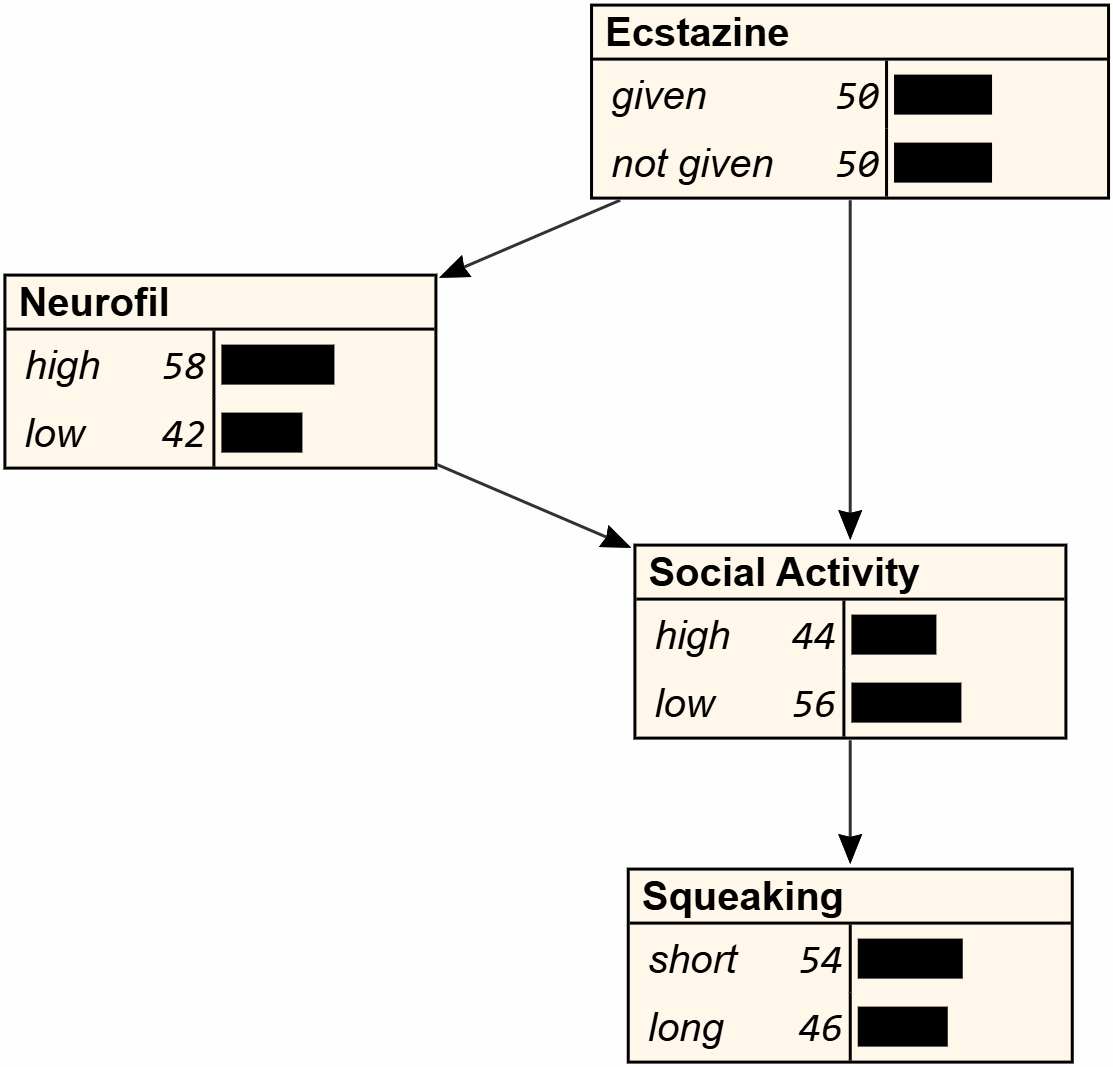}
    \caption{Rats BN\label{fig:rats_base}}
    \end{subfigure}
\caption{Rats narrative and BN with no instantiated variables.\label{fig:rats_intro}}
\end{figure}

Figure~\ref{fig:car_intro} displays the Car Battery example used for training, 
and Figure~\ref{fig:rats_intro} displays the Rats narrative and BN used for testing.

%% file: AppendixB-Skills.tex
\citet{Brase2021} found that Bayesian reasoning was best predicted
by measures of numerical literacy and visuospatial ability (including
the Paper Folding Task), as opposed to different measures of cognitive
thinking dispositions/styles, ability to conceptually model set-theoretic
relationships, or cognitive processing ability (working memory span).

\subsection{Subjective Numeracy Scale}
The \define{Subjective Numeracy Scale} (\define{SNS}\/), developed by
\citet{FagerlinEtAl2007}, consists of eight self-assessment numeracy
questions (Figure~\ref{fig:SNS_questions}). The answers are given on a
6-point Likert scale, where 1 indicates a low preference for numerical
information or a low proficiency in processing it, and 6 indicates a
high preference or proficiency. Participants' SNS score is the average
of their answers' scores in the SNS.

\input figB8_SNS_questions

\subsection{Paper Folding Task}
\begin{figure}[htb]
   \centering
   \includegraphics[width=\linewidth]{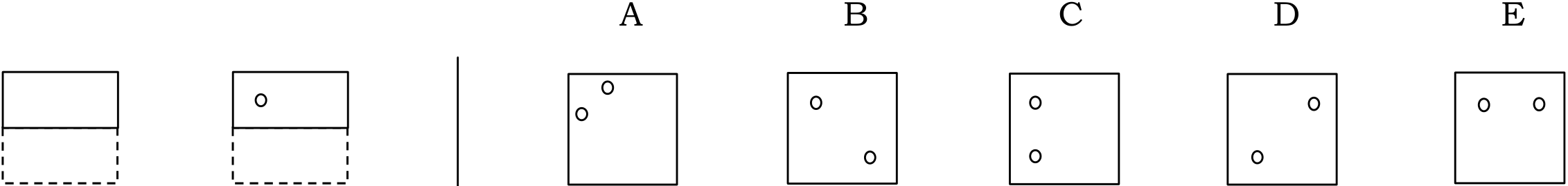}
   \caption{Example of a paper folding question. Among the five answers on the right, only C is correct.\label{fig:folding_eg}}
\end{figure}

The \define{Paper Folding Task} was developed by \citet{EkstromEtAl1976} as
a way to assess spatial ability. We used the first block of 10 multiple-choice 
questions, which has a 3 minute total
time limit. In each question, participants view diagrams of a square
piece of paper being folded and a hole punched through the folded
paper in a specific location. They must then select the corresponding
punch pattern when the paper is unfolded from five options presented.
Figure~\ref{fig:folding_eg} shows the training question for our task.

%% file: figB8_SNS_questions.tex
\begin{figure}[ht]
\centering
\vspace*{-1mm}
\begin{tabular}{|p{0.92\textwidth}|}\hline
\begin{minipage}{0.91\textwidth}
\vspace*{2ex}
\begin{enumerate}
\item
Please indicate how good you are at each of the tasks listed below:\\
{\scriptsize $\bullet$}~~Working with fractions\\
{\scriptsize $\bullet$}~~Working with percentages\\
{\scriptsize $\bullet$}~~Calculating a 15\% tip\\
{\scriptsize $\bullet$}~~Figuring out the price of a shirt that is 25\% off
\myitem
When reading the newspaper, how helpful do you find tables and graphs that are part of a story?
\myitem
When people tell you the chance of something happening, do you prefer that they use words (``it rarely happens'') or numbers (``there's a 1\% chance'')?
\myitem
When you hear a weather forecast, do you prefer predictions using percentages (\eg ``there will be a 20\% chance of rain today'') or predictions using only words 
(\eg ``there is a small chance of rain today'')?
\myitem
How often do you find numerical information useful?
\vspace*{2ex}
\end{enumerate}
\end{minipage}\\
\hline
\end{tabular}
\caption{Questions in the Subjective Numeracy Scale. Answers are on a
6-point Likert scale.\label{fig:SNS_questions}}
\end{figure}

%% file: AppendixC-Templates.tex
Figure~\ref{fig:template-contrib} shows the template for our
detailed explanation of the contribution of a finding \find\ when
there are multiple paths connecting it to the target variable.
Table~\ref{tab:template-features} interprets the symbolic expressions
and lists example values,
which produce the verbal output in
Figure~\ref{fig:podunk_sunscreen_verbal}(c) (replicated in
Figure~\ref{fig:detail-contrib}).

The paths are listed from strongest to weakest, and described arc by arc. 
The functions $\say(\find)$ and $\say(C_G(\find, \BNpath_i, \Finds))$ name 
a finding and identify its contribution via a path $i$ respectively. 
Similarly, \linebreak
$\say(D_G(\find, \var{var}_{j}, \BNpath_i, \Finds))$
selectively expresses differences a finding makes to the 
state probabilities of an intermediate variable $j$ on path $i$,
and $\say(\var{var}_{j}, \BNpath_i)$ names those states.

Some rules for $\say(\find)$ and $\say(\var{var}_{j}, \BNpath_i)$, 
with example outputs coloured green, are: 
\begin{description}
\leftmargin=0pt \itemindent=0pt
\myitem[Default:]
Use the standard phrase `that \var{variable} is \state{state}', 
\eg the probability \inst{that \guivar{Peeling} is \guistate{persistent}}.

\myitem[Concise:]
Use concise versions where available.
\begin{itemize}
\myitem
Use \state{state} followed by \var{variable} when they can be
reversed, \eg the probability \inst{of \guistate{persistent} \guivar{Peeling}}.
\item
Omit \state{state} if it
is \state{present} and the only alternative is \state{absent}, 
\eg the probability \inst{of \guivar{Dermascare}}.
\item
Use short names (if any are specified), \eg \inst{\guivar{Sunscreen}} instead of \inst{\guivar{Dunkalot$\,$Sunscreen}}.
\item
But use the default version in the first sentence of Summary explanations (on the impact of all the evidence).

\end{itemize}

\myitem[Tense:]
Replace `is' with `was' if $\find$ is known to be earlier than $\target$ in the BN variable order.
This may be determined by a causal path from $\find$ to $\target$, \eg from \var{Sunscreen} to \var{Dermascare}, or any supplementary information, \eg if the BN builder specified that the \var{Mutation} occurs before \var{Dermascare}.
\end{description}

\input figC10_say_contrib_template
\vspace*{1ex}
\input figC11_say_contrib_detail

\input tab-template-rubric

%% file: figC10_say_contrib_template.tex
\begin{figure}[!h]
\centering
\begin{tabular}{|p{0.7\textwidth}|}
    \hline \vspace*{-0.5ex}
    \textbf{Detail: How finding out $\say(\find)$ contributes} \vspace*{1ex}\\
    Finding out $\say(\find)$ contributes due to $\say(|\bnpaths|)$
    connections:\vspace*{-1ex}
     \begin{itemize}
        \myitem It $\say(D_G(\find, \var{var}_{j}, \BNpath_i, \Finds))$ the probability $\say(\var{var}_{j}, \BNpath_i)$, 
        which in turn $\say(C_G(\find, \BNpath_i, \Finds))$ the probability $\say(\target)$.\vspace*{-1ex}
        \myitem By direct connection, it $\say(C_G(\find, \BNpath_k, \Finds))$ the probability $\say(\target)$.\vspace*{-1ex}
     \end{itemize}
     Overall, the finding $\say(C_G(\find, \Finds))$ the probability $\say(\target)$. \vspace*{1ex}\\
     \hline
\end{tabular}
\caption{Template for saying how \varft{finding} contributes to the
\varft{target} probability via multiple paths; the pseudocode functions and variables are in small caps and monospaced font respectively.\label{fig:template-contrib}}
\vspace*{-2ex}
\end{figure}

%% file: figC11_say_contrib_detail.tex
\begin{figure}[!h]
\centering
\begin{tabular}{|p{0.7\textwidth}|}
    \hline \vspace*{-0.5ex}
    \textbf{Detail: How finding out \inst{\guivar{Dunkalot$\,$Sunscreen} was \guistate{used}} contributes} \vspace*{1ex}\\
    Finding out \inst{\guivar{Dunkalot$\,$Sunscreen} was \guistate{used}} contributes due to \inst{two} connections:\vspace*{-1ex}
     \begin{itemize}
        \myitem It \inst{increases} the probability \inst{that \guivar{Podunk$\,$Beach} was \guistate{visited}}, 
        which in turn  \inst{greatly increases} the probability \inst{of \guivar{Dermascare}}.\vspace*{-1ex}
        \myitem By direct connection, it  \inst{slightly reduces} the probability \inst{of \guivar{Dermascare}}.\vspace*{-1ex}
     \end{itemize}
      Overall, the finding \inst{\underline{moderately increases}} the probability \inst{of \guivar{Dermascare}}. \vspace*{1ex}\\
    \hline
\end{tabular}
\caption{Detailed explanation for how finding out about 
\varft{Dunkalot$\,$Sunscreen} contributes to the probability 
of \varft{Dermascare}, generated from the template in 
Figure~\ref{fig:template-contrib} and values in
Table~\ref{tab:template-features}; the $\say$ function outputs are coloured green.\label{fig:detail-contrib}}
\end{figure}

%% file: tab-template-rubric.tex
\begin{table*}[t]
\vspace*{-5mm}
\setlength{\tabcolsep}{6pt}
\caption{Expressions for the verbal template in
Figure~\ref{fig:template-contrib}, and values which yield the explanation
in Figure~\ref{fig:detail-contrib}.\label{tab:template-features}}
\centering
\vspace*{-2mm}
{\small \raggedright
    \begin{tabular}{l l l}
    \toprule
    \textbf{Symbols} & \textbf{Semantics} & \textbf{Sample Value}\\
    \midrule
   \varft{finding} & BN \varft{finding} variable has \statetab{state}  & 
        {\varft{Dunkalot$\,$Sunscreen} $=$ \statetab{used}}\\
    \varft{target} & a {\statetab{state}} of \varft{target} BN variable  & \varft{Dermascare}, \statetab{present}\\ 
    $\BNpathft_i$ & BN path $i$ from \varft{finding} to \varft{target} & 
        \mbox{\varft{Dunkalot$\,$Sunscreen}~$\leftarrow$~\varft{Podunk$\,$Beach}}~$\rightarrow$~\varft{Dermascare}\\
    $\BNpathft_k$ & BN path $k$ from \varft{finding} to \varft{target} & 
        \varft{Dunkalot$\,$Sunscreen}~$\rightarrow$~\varft{Dermascare}\\
    $\varsc{var}_{j}, \BNpathft_i$ & BN variable $j$ on path $i$ & \varft{Podunk$\,$Beach}\\
    $|\varft{Paths}|$ & number of paths & 2\\
    $C_G(\varft{finding}, \Finds)$ & the contribution of \varft{finding} & $+28\%$, moderately increases\\
    $C_G(\varft{finding}, \BNpathft_i, \Finds)$ & the contribution of \varft{finding} via $\BNpathft_i$ & $+32\%$, greatly increases\\
    $C_G(\varft{finding}, \BNpathft_k, \Finds)$ & the contribution of \varft{finding} via $\BNpathft_k$  & $-4\%$, slightly reduces\\
    \multirow{2}{*}{$D_G(\varft{finding}, \varft{var}_j, \BNpathft_i, \Finds)$} & {the differences \varft{finding} makes to the} & 
        {$+51\%$, greatly increases [the probability that }\\
         &   probabilities of $\varft{var}_j$ states on $\BNpathft_i$ & \varft{Podunk$\,$Beach} was \statetab{visited}]\\
    \bottomrule
    \end{tabular}
}
    \vspace*{15mm}
\end{table*}

%% file: Appendix-Visual.tex
Table~\ref{tab:visual_semantics} lists key functions in our visual code, specifies their semantics and use, and gives an example of their output. We also note how each visual element is added in animations, where any gradual addition is always completed before adding the next element. Algorithms \ref{alg:summarise}--\ref{alg:detail_ani_2b} were sufficient for all the BNs in our experiment.

\input tab-visual-spec


\newpage
\begin{algorithm}[H]
\caption{Summary of all findings}
\label{alg:summarise}
\small
\begin{algorithmic}
    \LineComment{Show individual and total contributions of a set of findings $\Finds$ to \target\ using static markup}
    \Function{Summarise}{$\Finds$, \target}
        \State \Call{RemoveMarkup}{}
        \Comment{Remove any existing markup and highlight target state and variable}
        \State \Call{Highlight}{\target}

        \Statex
        \ForAll{\find\ $\in \mathbf{F}$}
            \Comment{Highlight each finding's state name with its contribution colour}
            \State \Call{Highlight}{\find, $C_G(\find, \mathbf{F})$}
        \EndFor

        \Statex
        \State \Call{Diffbox}{$C_G(\mathbf{F})$}
        \Comment{Draw box with the total probability difference due to all findings}
    \EndFunction
\end{algorithmic}
\end{algorithm}

\begin{algorithm}[H]
\caption{Detail about a finding}
\label{alg:detail_static}
\small
\begin{algorithmic}
    \LineComment{Explain \find's contribution to \target\ via the \bnpaths\ connecting them,
    and explain interacting elements}
    \Function{Detail}{\find, \target}
        \State \bnpaths $\gets$ \Call{GetPathsBetween}{\find, \target}
        \Comment{Identify all connecting and interacting paths (store in \apaths)}
        \State $\ipaths \gets $ \Call{GetInteractingPaths}{\bnpaths}
        \State $\apaths \gets \bnpaths \cup \ipaths$

        \Statex
        \State \Call{RemoveMarkup}{}
        \Comment{Remove any existing markup and highlight target state and variable}
        \State \Call{Highlight}{\target}

        \Statex
        \State $H \gets G \setminus \apaths$
        \Comment{Fade non-interacting elements (variables and arcs)}
        \State \Call{Fade}{$H$}

        \Statex
        \State \Call{Glow}{\find}
        \Comment{Add glow to \find\ variable and highlight its state name with its contribution colour}


        \Statex
        \ForAll{$\apath \in \apaths$}
            \Comment{For every active path, highlight contribution of each path element.}
            \ForAll{$\var{el} \in \apath$}
                \CommentCont{\var{path} sequence is: finding, arc, variable, arc, ..., target} 
                \State \Call{Highlight}{$\var{el}$, $C_G(\var{el}, \mathbf{F}))$}
            \EndFor
        \EndFor
        
        \Statex
        \State \Call{Diffbox}{$C_G(\find, \mathbf{F})$, \var{glow}=T}
        \Comment{Add diffbox with glow on target state showing contribution due to \find}
    \EndFunction
\end{algorithmic}
\end{algorithm}

\begin{algorithm}[H]
\caption{Animated detail about a finding}
\label{alg:detail_ani_2b}
\small
\begin{algorithmic}
    \LineComment{Explain \find's contribution to \target\ using animated changes along \bnpaths,
    and explain contributions from interactions}
    \Function{Animate\_2b}{\find, \target}
        \State $\bnpaths \gets$ \Call{GetPathsBetween}{\find, \target}
        \Comment{Initialise scene per \Call{Detail}{\find, \target}}
        \State $\ipaths \gets $ \Call{GetInteractingPaths}{\{\BNpath\}}
        \State $\apaths \gets \bnpaths \cup \ipaths$

        \Statex
        \State \Call{RemoveMarkup}{}
        \State \Call{Highlight}{\target}

        \Statex
        \State $H \gets G \setminus \apaths$
        \State \Call{Fade}{$H$}

        \Statex

        \State \Call{RemoveEvidenceFromBN}{\find}
        \Comment{First, show (instantly) contributions on interacting paths \emph{without} \find}
        \ForAll{$\ipath \in \ipaths$}
            \ForAll{$\var{el} \in \ipath$}
                \State \Call{Highlight}{$\var{el}$, $C_G(\var{el}, \mathbf{F}))$}
            \EndFor
        \EndFor

        \Statex
        \If{\var{prevfinding} exists}
            \Comment{Instantly show previous finding's diffbox (if any), then animate removal}
            \State \Call{Diffbox}{$C_G(\var{prevfinding}, \mathbf{F}))$, \var{glow}=T}
            \State \Call{RemoveDiffbox}{\var{anim}=T}
        \EndIf

        \Statex
        \State \Call{Glow}{\find, \var{anim}=T}
        \Comment{Flash the \find\ variable, add its evidence and highlight finding}
        \State \Call{AddEvidenceToBN}{\find}
        \State \Call{Highlight}{\find, $C_G(\find, \mathbf{F}))$}

        \Statex
        \ForAll{$\bnpath \in \bnpaths$ ordered by $C_G(\BNpath, \mathbf{F}))$ descending}
            \Comment{Animate the contributions along the connecting paths}
            \ForAll{$\var{el} \in \bnpath$ in sequence}
                \CommentCont{in descending order of path contribution}
                \State \textbf{if} \var{el} $\in \{\find,\target\}$ \textbf{then} \textbf{next} \var{el}
                \State \Call{Highlight}{$\var{el}$, $C_G(\var{el}, \mathbf{F}))$, \var{anim}=T}
                \Comment{\var{path} sequence is: arc, variable, arc, variable, ..., arc}
            \EndFor
        \EndFor



        \Statex
        \State \Call{Diffbox}{$C_G(\find, \mathbf{F}))$, \var{glow}=T, \var{anim}=T}
        \Comment{Animated, so box grows in the direction that the probability changes}
    \EndFunction
\end{algorithmic}
\end{algorithm}







%% file: tab-visual-spec.tex
\begin{table}[H]
\setlength{\tabcolsep}{4pt}
    \centering
    \small
    \caption{Visual functions, their semantics, example output and animation behaviour.\label{tab:visual_semantics}}
    \begin{tabular}{p{0.24\textwidth}p{0.33\textwidth}>{\centering\arraybackslash}p{0.2\textwidth}p{0.16\textwidth}}
        \toprule
        \textbf{Function} & \textbf{Semantics} & \textbf{Example} & \textbf{In Animations} \\
        \midrule
        \addlinespace[2ex]
        
        {$\sho(\vartab{target})$} & Highlight the names of \vartab{target} state and variable in yellow. Used to make \vartab{target} salient with a unique colour.  & \includegraphics[width=0.7\linewidth, valign=t]{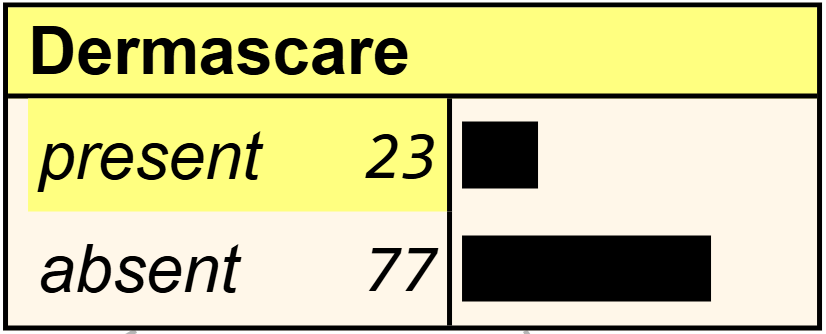} & instant \\
        \addlinespace[2ex]    
        
        {$\sho(\vartab{finding}, \vartab{contrib})$} & Highlight the state name of the \vartab{finding} variable in the colour of contribution size \vartab{contrib}. Used to show the contribution of \vartab{finding}. & \includegraphics[width=0.55\linewidth, valign=t]{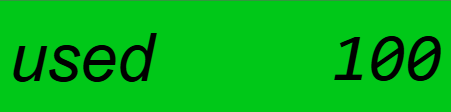} & fades in (gradually replacing any existing highlight) \\ 
        \addlinespace[2ex]
        
        $\sho(\vartab{arc}, \vartab{contrib})$ & Highlight \vartab{arc} in the colour of the contribution size \vartab{contrib}. Used to show the contribution of \vartab{arc}. & \includegraphics[width=0.6\linewidth, valign=t]{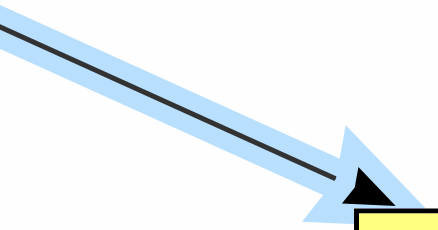} & drawn towards the target on each arc \\
        \addlinespace[2ex]

        & Adding/changing colour shows a contribution difference of \vartab{arc}, \eg due to adding \vartab{finding}. & \includegraphics[width=0.65\linewidth, valign=t]{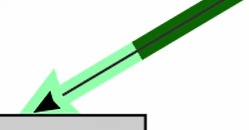} & paints over any previous highlight \\ 
        \addlinespace[2ex]
        
        $\sho(\vartab{interVar}, \vartab{contrib})$ & Highlight any increase in a probability bar on the intermediate variable \vartab{interVar} in the colour of contribution size \vartab{contrib}. Used to show the difference \vartab{finding} makes to \vartab{interVar}'s probability distribution, and hence \vartab{target} probability. & \includegraphics[width=0.5\linewidth, valign=t]{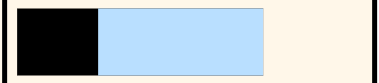} & drawn left to right to mimic the increase, which here slightly reduces \vartab{target} probability \\ 
        \addlinespace[2ex]
        
        {$\diffbox(\vartab{target}, \vartab{contrib})$} & Draw a box on \vartab{target} state's probability bar around a probability difference (increase or reduction) of size \vartab{contrib}, in the colour of that contribution size. Used to show the contribution of all findings, $C_G(\Finds)$. & \includegraphics[width=0.27\linewidth, valign=t]{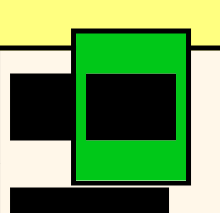} & drawn left to right to mimic increases (green); reversed for reductions (blue) \\ 
        \addlinespace[2ex]
        
        & With \vartab{glow=T}: include red glow around diffbox. Used 
        to indicate this is only the contribution of \vartab{finding}, not of all findings $\Finds$. & \includegraphics[width=0.35\linewidth, valign=t]{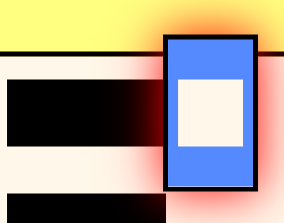} & matches the size of \text{diffbox} as it is drawn \\
        \addlinespace[2ex]        

        $\glow(\vartab{var})$ & Add red glow around the variable \vartab{var}. Used to make the \vartab{finding} variable more salient in a detailed explanation. & \includegraphics[width=0.8\linewidth, valign=t]{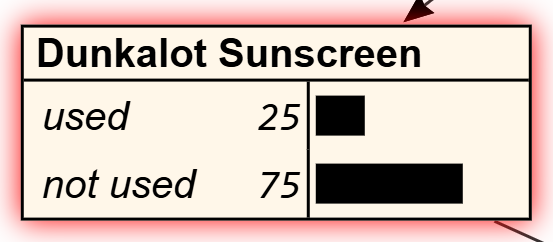} & flashes repeatedly (by cycling through quick fades in and out), then stays on \\ 
        \addlinespace[2ex]    
        
        $\fade(\vartab{elements})$ & Fade the BN components in \vartab{elements}. Used to indicate variables and arcs that do not interact with \vartab{finding}.  & \includegraphics[width=0.8\linewidth, valign=t]{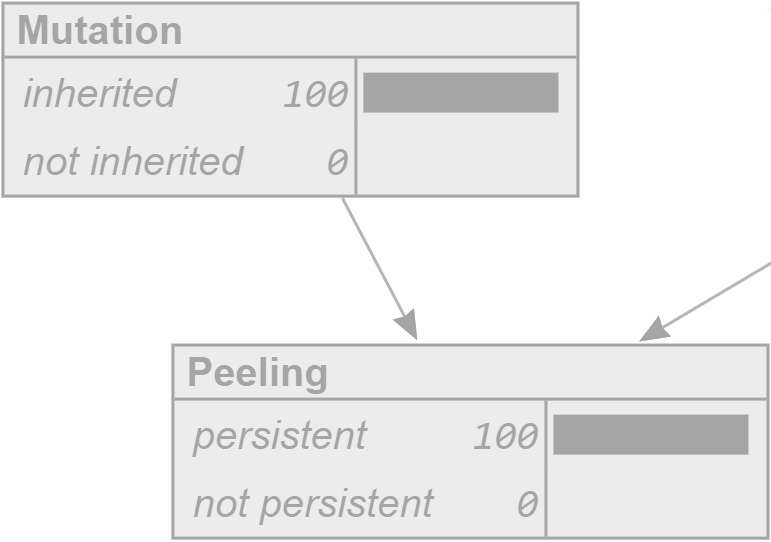} & instant \\ 

        \vartab{anim=T} & Argument that specifies the function should be run as an animation (when set to true). 
        & & switches on all the behaviours above \\ 
        \addlinespace[2ex]
        
        \bottomrule
    \end{tabular}
\end{table}

%% file: AppendixD-Screenshots.tex
\begin{figure}[!h]
    \centering
    \includegraphics[width=0.7\textwidth]{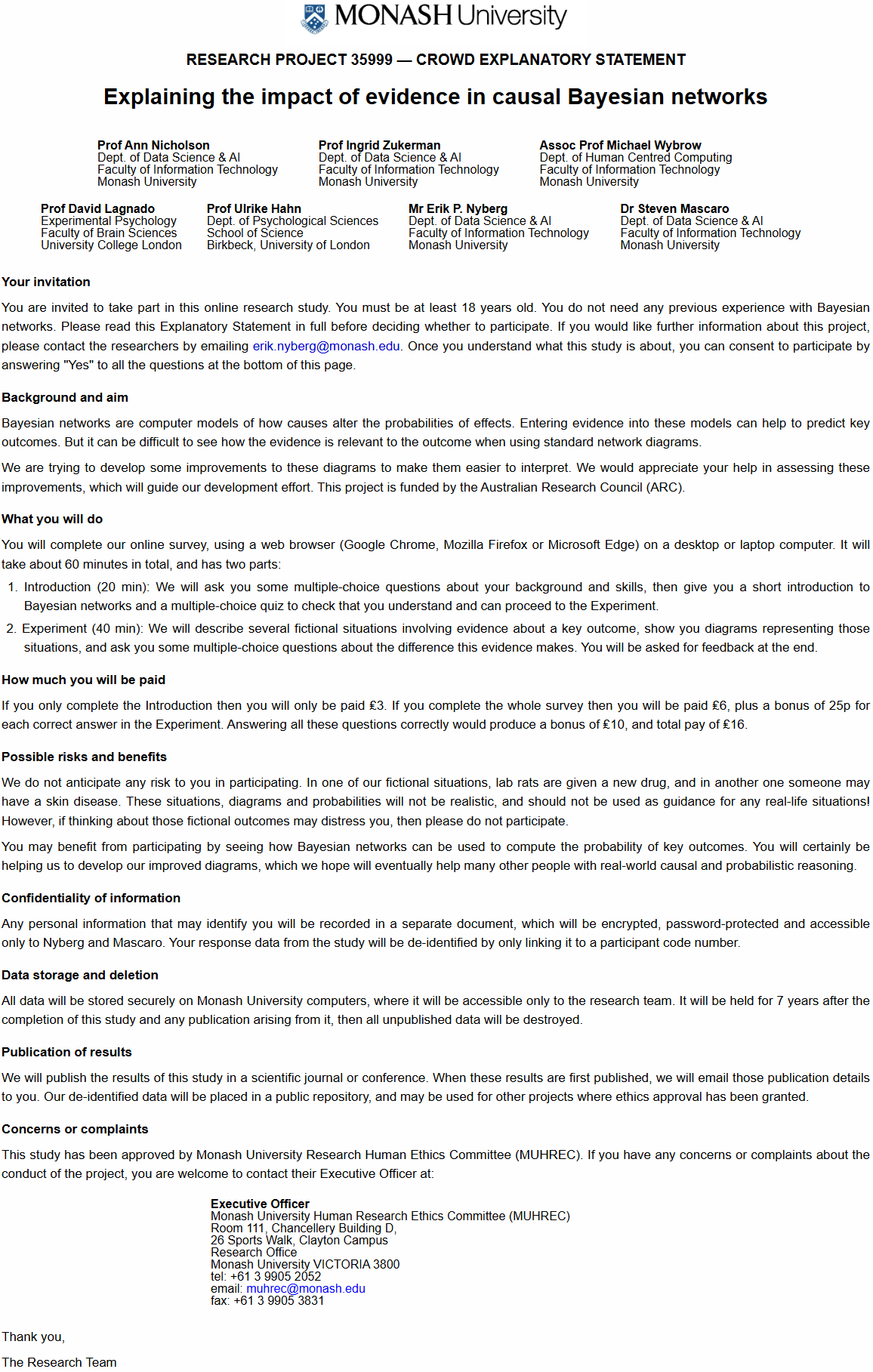}
    \caption{Introduction and explanatory statement: overview of the research project on explaining evidence in Bayesian networks; details of the research team, study aim, participant requirements, and two-part study procedure including a background quiz and main experiment.\label{fig:explan_statement}}
\end{figure}
\begin{figure}[tbp]
\centering
\begin{subfigure}[t]{0.9\textwidth}
    \includegraphics[trim={0 0 0 15mm},clip,width=\textwidth]{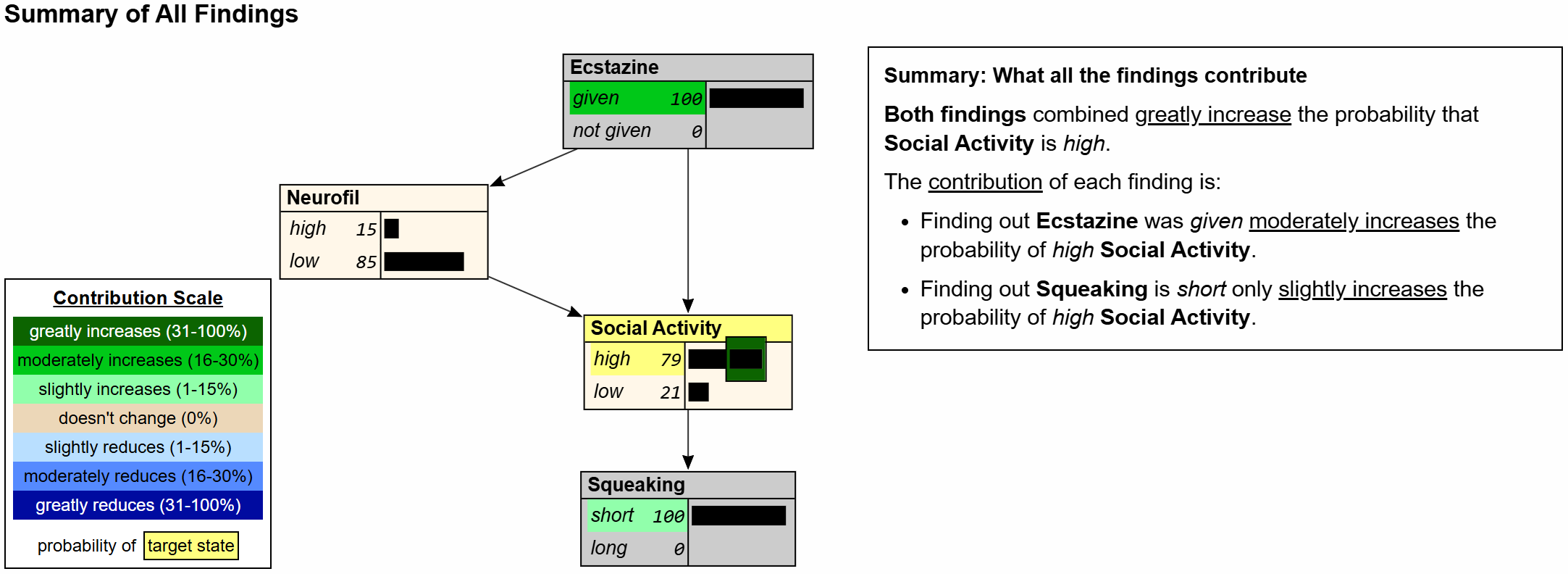}\vspace*{2ex}\\
        \vspace*{-0.37\textwidth}\\ {\huge\color{red}{$A$}} \vspace*{0.30\textwidth}\\
    \includegraphics[width=0.6\textwidth]{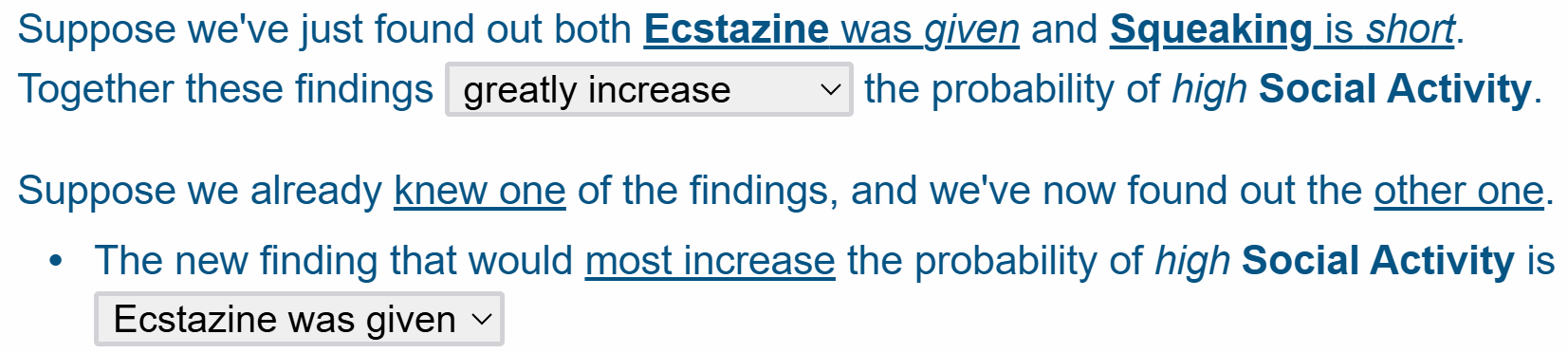}\\
        \vspace*{-0.152\textwidth}\\
        \hspace*{0.65\textwidth}\qtypelabel{Control}\vspace*{0.05\textwidth}\\
        \hspace*{0.65\textwidth}\qtypelabel{Finding}\vspace*{1ex}
    \caption{\label{fig:rats_sum_A}Version $A$: The \varsc{Ecstazine} finding increases the probability of high \varsc{Social$\,$Activity} more than the \varsc{Squeaking} finding.}
\end{subfigure}

\vspace*{4ex}
\begin{subfigure}[t]{0.9\textwidth}
    \includegraphics[trim={0 0 0 15mm},clip,width=\textwidth]{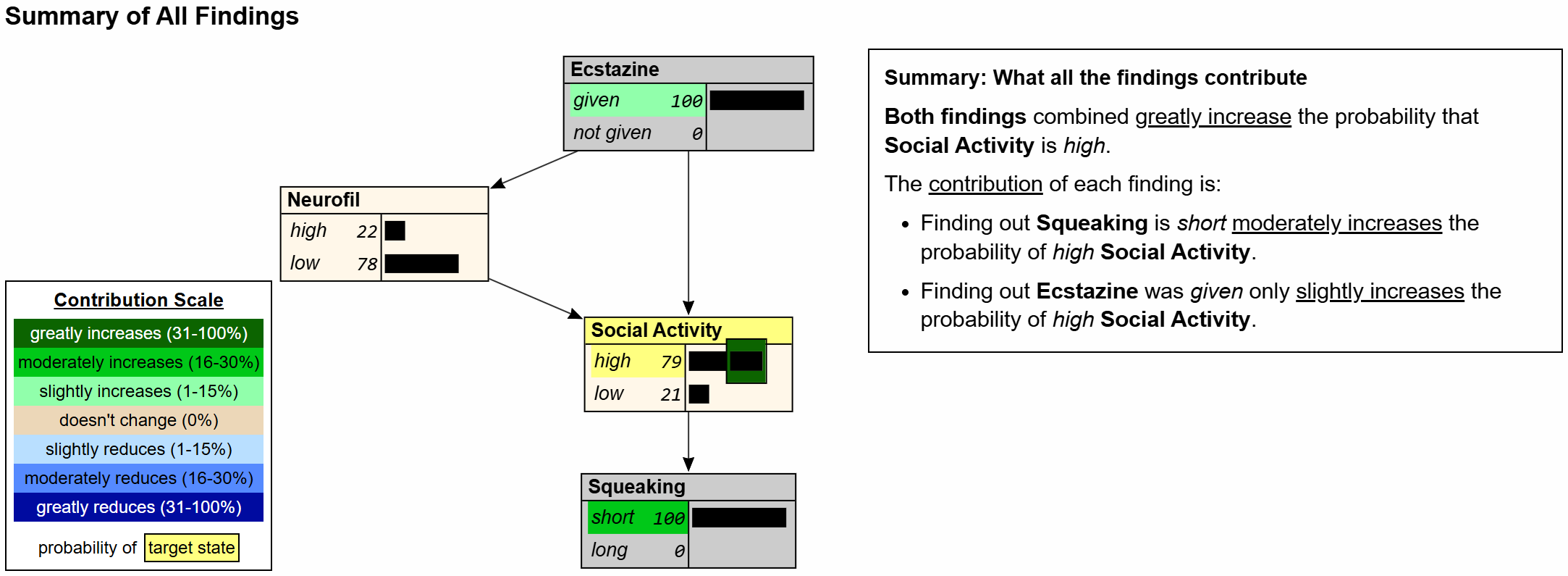}\vspace*{2ex}\\
        \vspace*{-0.37\textwidth}\\ {\huge\color{red}{$B$}} \vspace*{0.30\textwidth}\\
    \includegraphics[width=0.6\textwidth]{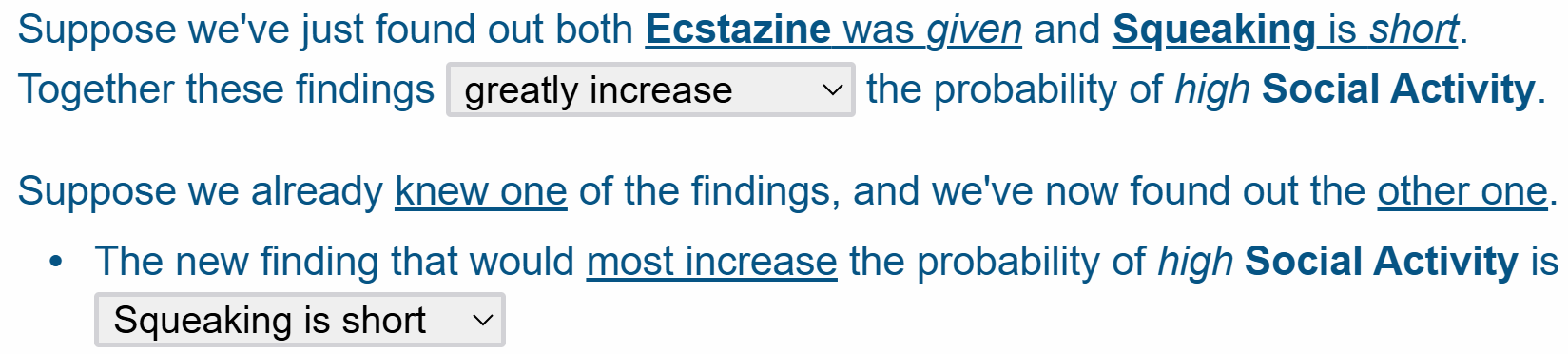}\\
        \vspace*{-0.152\textwidth}\\
        \hspace*{0.65\textwidth}\qtypelabel{Control}\vspace*{0.05\textwidth}\\
        \hspace*{0.65\textwidth}\qtypelabel{Finding}\vspace*{1ex}
    \caption{\label{fig:rats_sum_B}Version $B$: The \varsc{Squeaking} finding increases the probability of high \varsc{Social$\,$Activity} more than the \varsc{Ecstazine} finding.}
\end{subfigure}
\caption{\label{fig:rats_sum}Rats section, Summary page: This shows the contributions of all findings to the probability of \varft{Social$\,$Activity}. The visual and verbal explanations, and the answers to one test question, differ between versions $A$ and  $B$. Red markup indicates the question types: \qtype{Control} and \qtype{Finding}.}
\end{figure}
\begin{figure}[tbp]
\centering
\begin{subfigure}[t]{0.9\textwidth}
    \includegraphics[trim={0 29mm 0 15mm},clip,width=\textwidth]{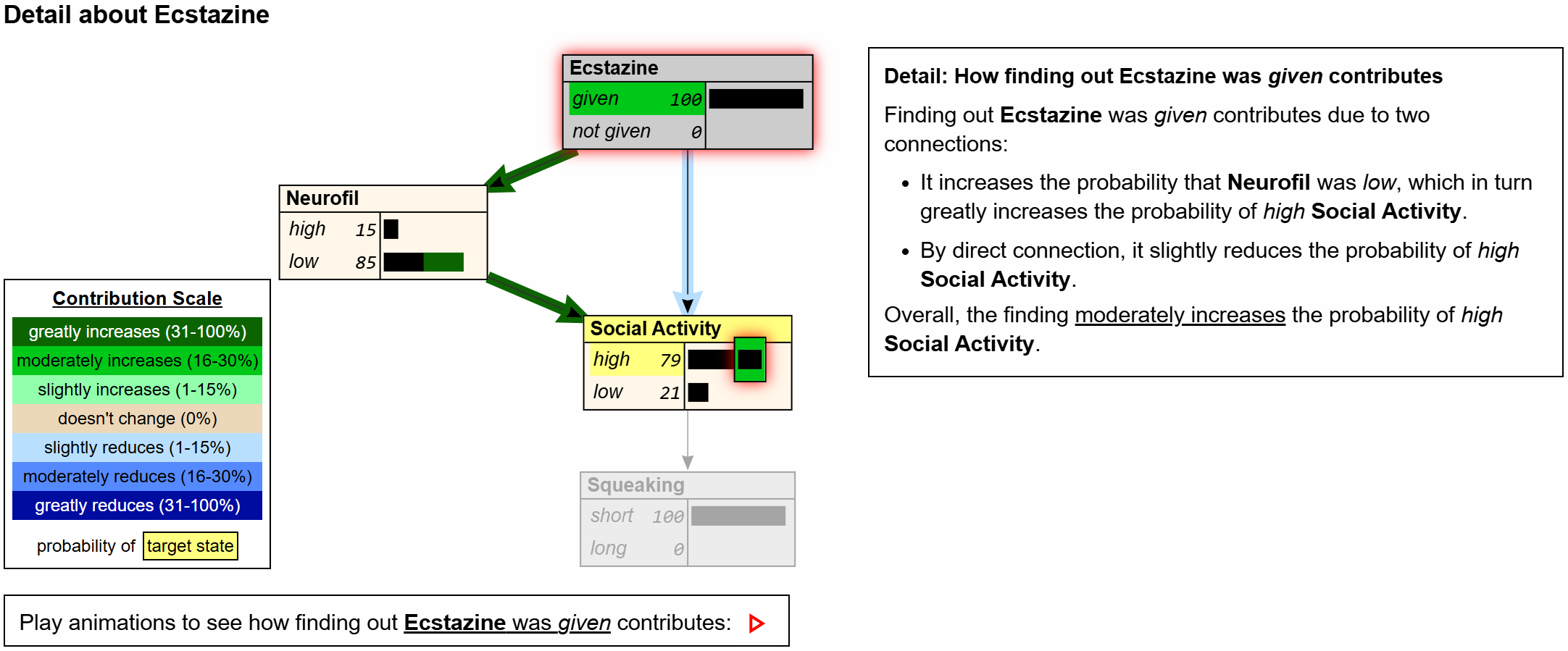}\vspace*{2ex}\\
        \vspace*{-0.36\textwidth}\\ {\huge\color{red}{$A$}} \vspace*{0.29\textwidth}\\
    \includegraphics[width=0.7\textwidth]{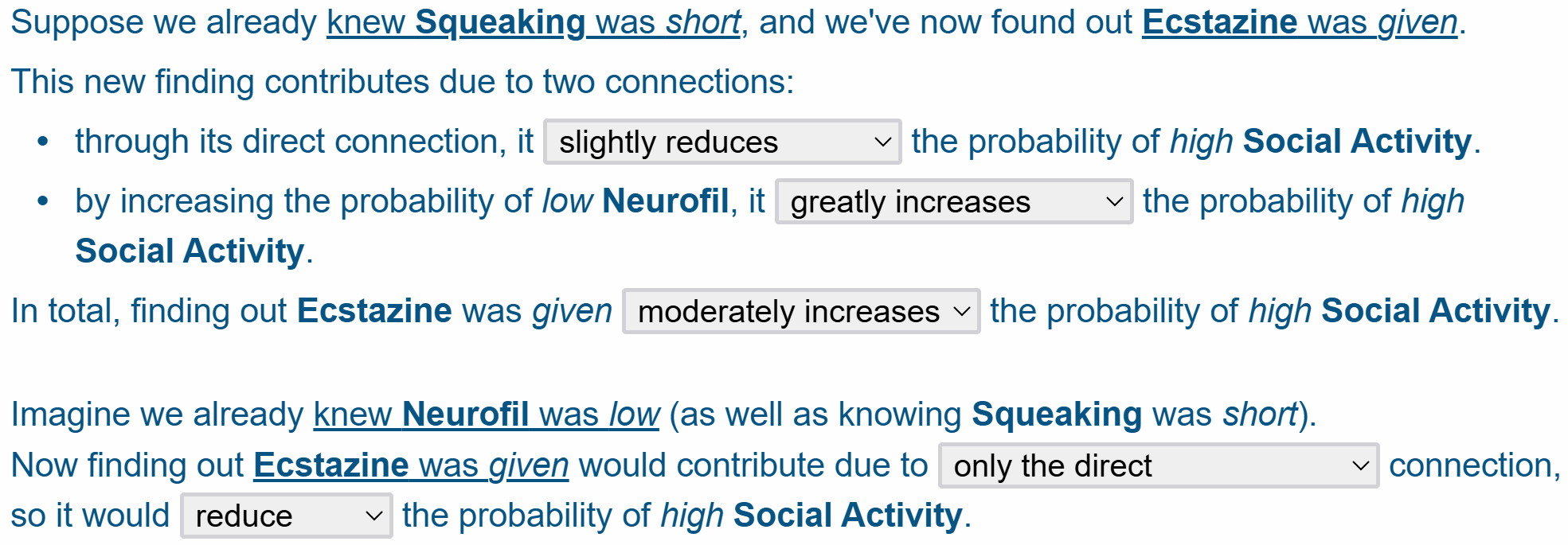}\\
        \vspace*{-0.233\textwidth}\\
        \hspace*{0.75\textwidth}\qtypelabel{Path}\vspace*{-0.001\textwidth}\\
        \hspace*{0.75\textwidth}\qtypelabel{Path}\vspace*{0.02\textwidth}\\
        \hspace*{0.75\textwidth}\qtypelabel{Finding}\vspace*{0.04\textwidth}\\
        \hspace*{0.75\textwidth}\qtypelabel{ControlThink}\vspace*{-0.005\textwidth}\\
        \hspace*{0.75\textwidth}\qtypelabel{PathThink}\vspace*{1ex}
    \caption{\label{fig:rats_ecstazine_A}Version $A$: The indirect path 
    is stronger than the direct path, and the two paths combined moderately increase the probability of high \varsc{Social$\,$Activity}.}
\end{subfigure}

\vspace*{4ex}
\begin{subfigure}[t]{0.9\textwidth}
    \includegraphics[trim={0 29mm 0 15mm},clip,width=\textwidth]{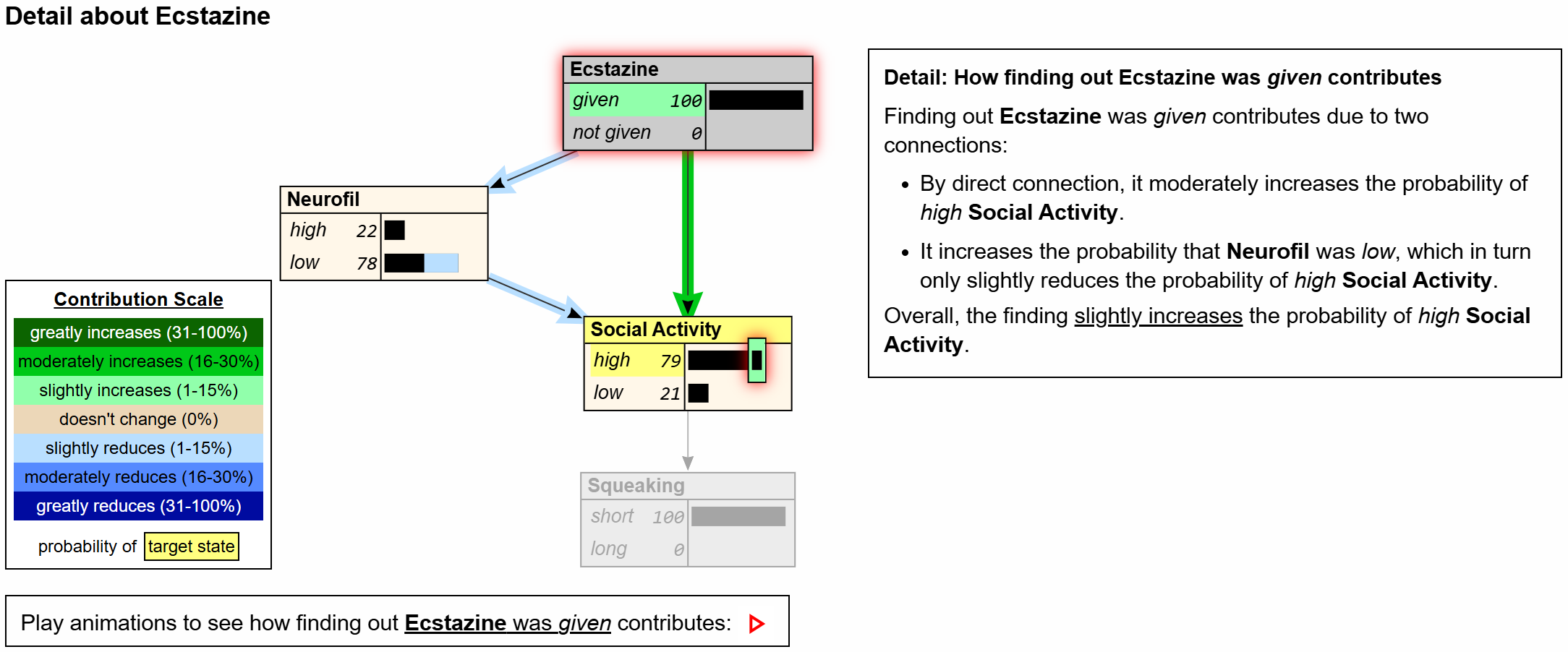}\vspace*{2ex}\\
        \vspace*{-0.36\textwidth}\\ {\huge\color{red}{$B$}} \vspace*{0.29\textwidth}\\
    \includegraphics[width=0.7\textwidth]{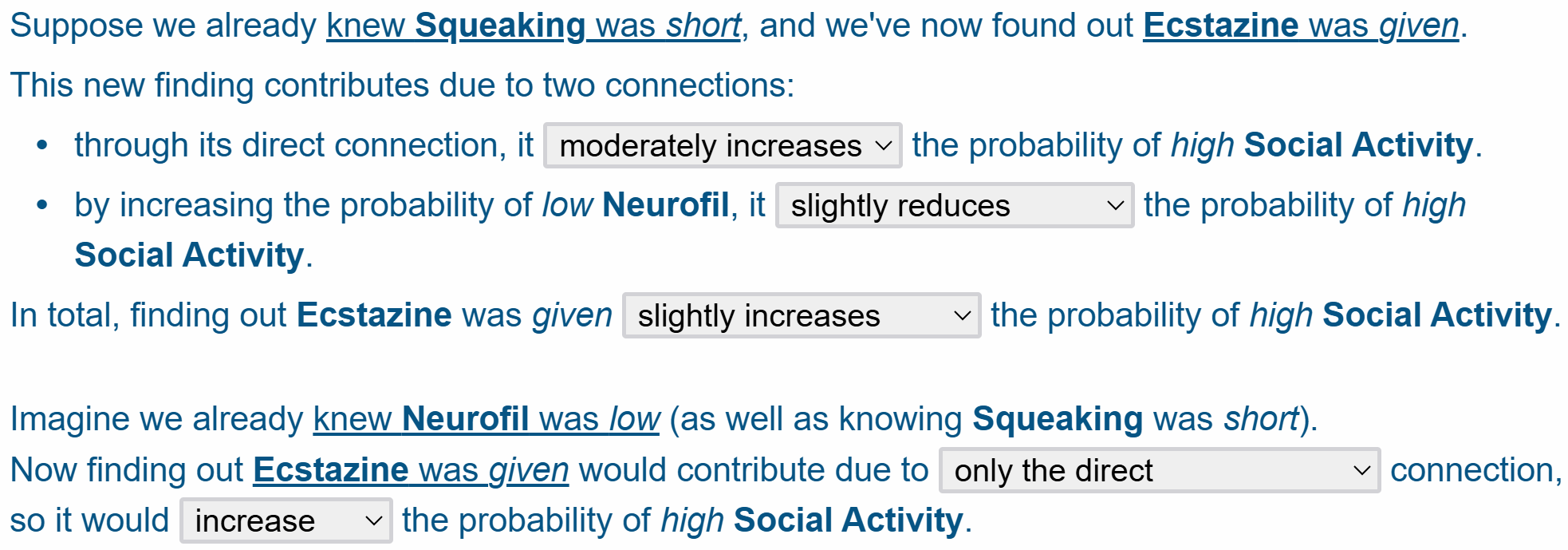}\\
        \vspace*{-0.233\textwidth}\\
        \hspace*{0.75\textwidth}\qtypelabel{Path}\vspace*{-0.001\textwidth}\\
        \hspace*{0.75\textwidth}\qtypelabel{Path}\vspace*{0.02\textwidth}\\
        \hspace*{0.75\textwidth}\qtypelabel{Finding}\vspace*{0.04\textwidth}\\
        \hspace*{0.75\textwidth}\qtypelabel{ControlThink}\vspace*{-0.005\textwidth}\\
        \hspace*{0.75\textwidth}\qtypelabel{PathThink}\vspace*{1ex}
    \caption{\label{fig:rats_ecstazine_B}Version $B$: The indirect path 
    is weaker than the direct path, and the two paths combined slightly increase the probability of high \varsc{Social$\,$Activity}.}
\end{subfigure}
\caption{\label{fig:rats_ecstazine}Rats section, \varft{Ecstazine} page: The visual and verbal explanations for the contribution of \varft{Ecstazine}, and the answers to four test questions, differ between versions $A$ and  $B$. The animation button is not shown. Red markup indicates the question types:  \qtype{Path}, \qtype{Finding}, \qtype{ControlThink} and \qtype{PathThink}.}
\end{figure}
\begin{figure}[htbp]
\centering
\begin{subfigure}[t]{0.37\textwidth}
    \includegraphics[width=\textwidth]{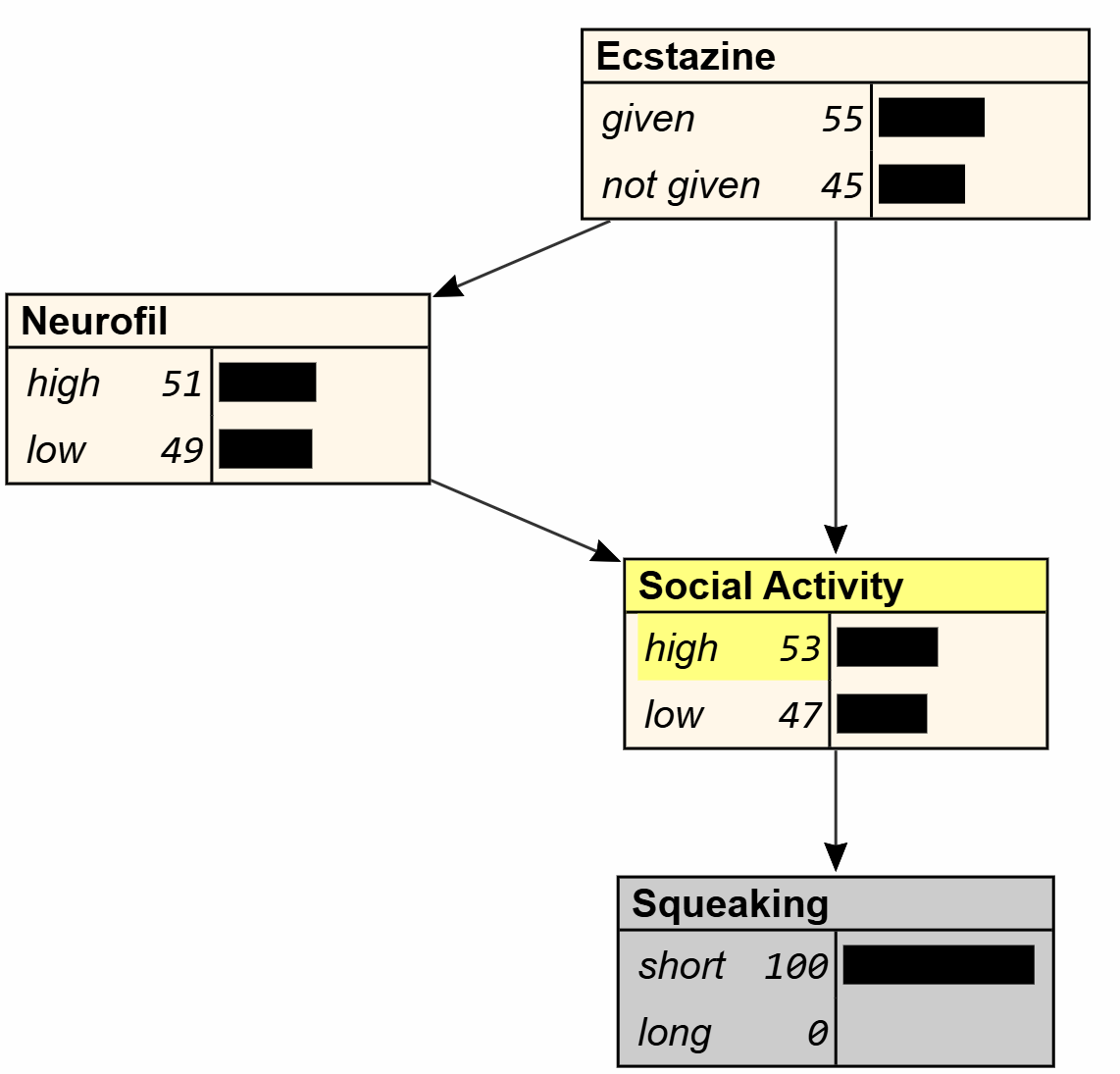}
    \caption{\label{fig:rats_ani_1}starting without the finding or focus on \varsc{Ecstazine}} 
    \end{subfigure}
    \hspace*{0.13\textwidth}
\begin{subfigure}[t]{0.37\textwidth}
    \includegraphics[width=\textwidth]{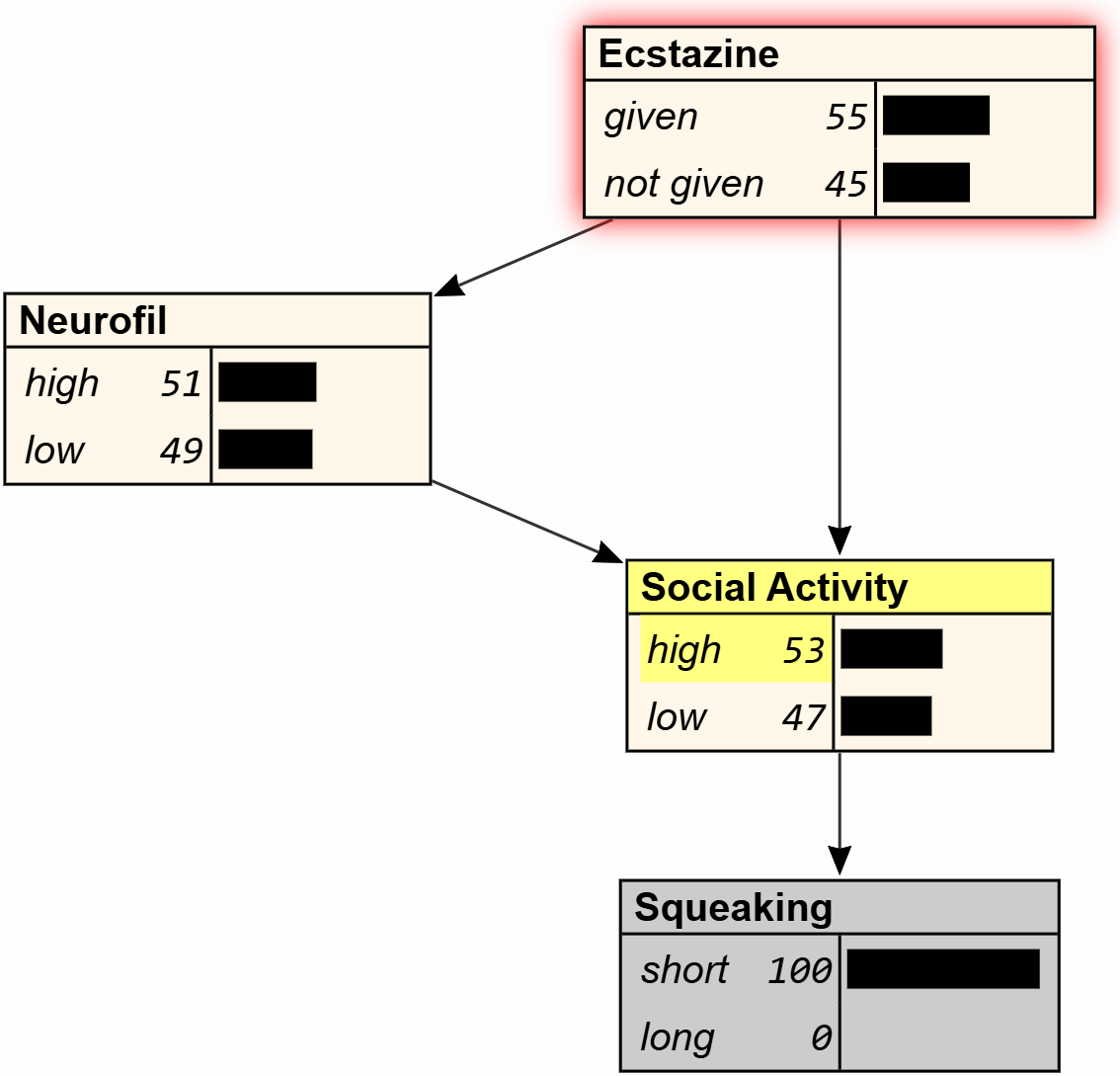}
    \caption{\label{fig:rats_ani_2}with focus highlighting added to \varsc{Ecstazine}} 
    \end{subfigure}\vspace*{5mm}
    
\begin{subfigure}[t]{0.37\textwidth}
    \includegraphics[width=\textwidth]{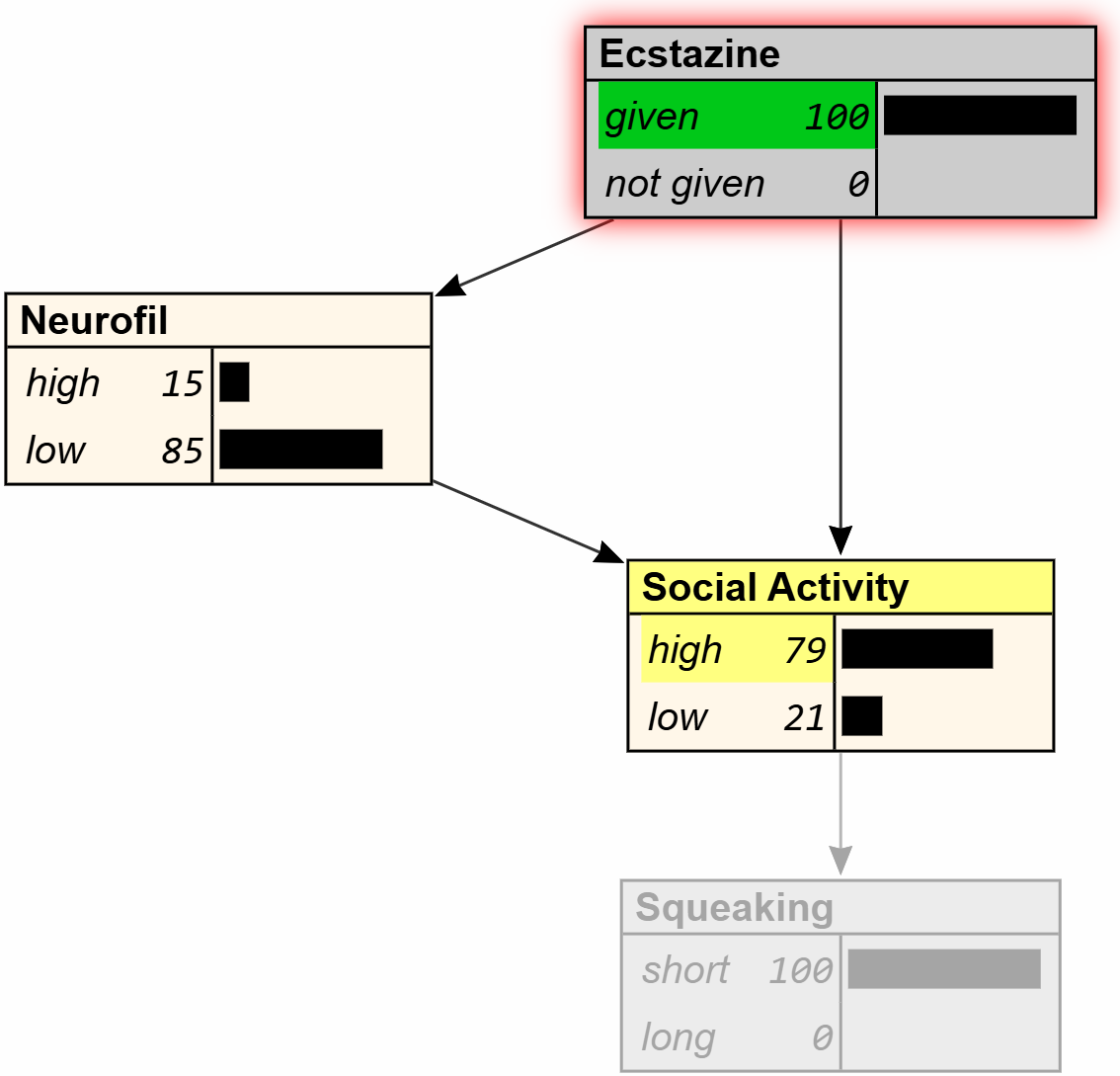}
    \caption{\label{fig:rats_ani_3}with the finding that \varsc{Ecstazine} was given added and highlighted}
    \end{subfigure}
    \hspace*{0.13\textwidth}
\begin{subfigure}[t]{0.37\textwidth}
    \includegraphics[width=\textwidth]{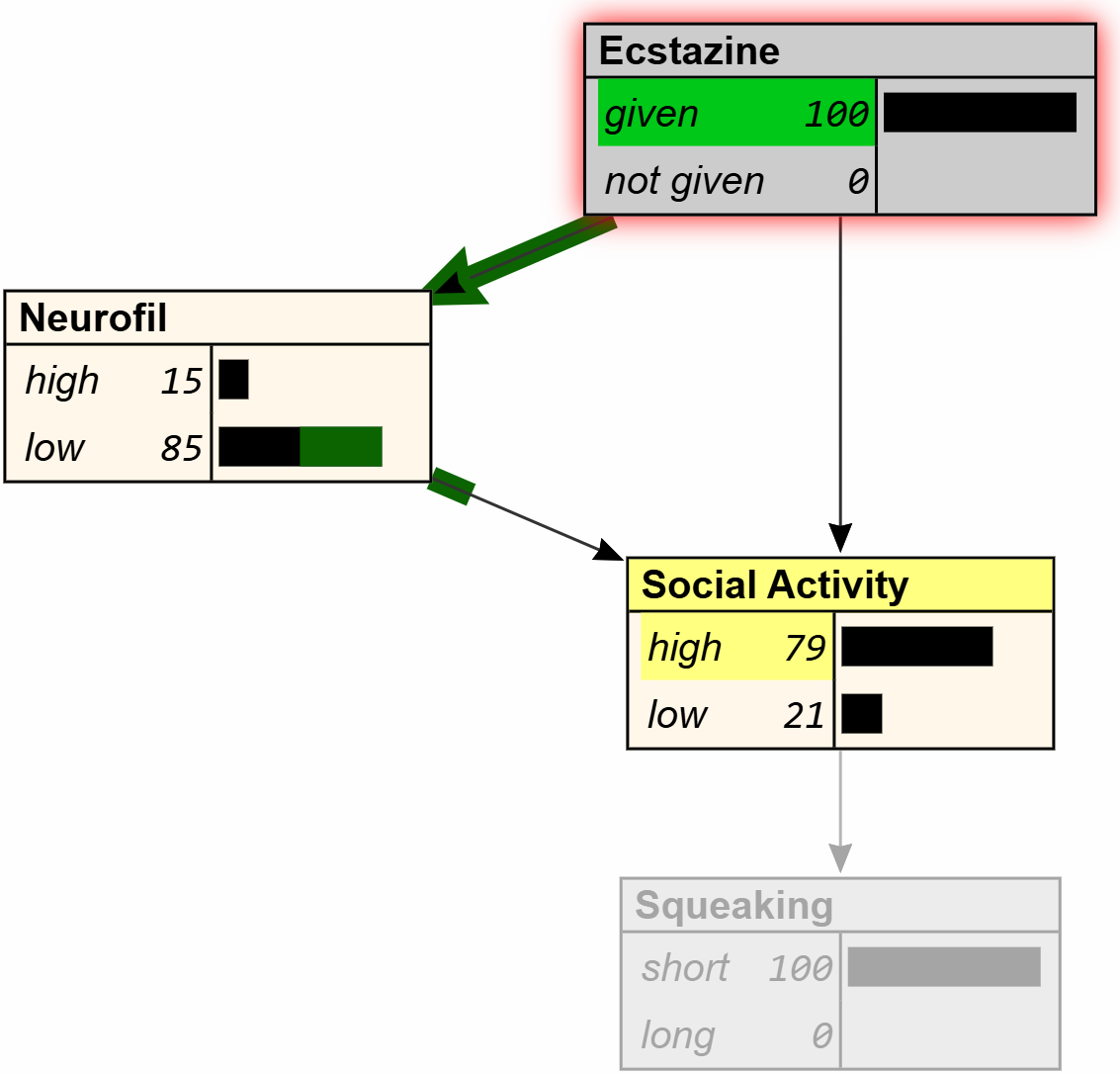}
   \caption{\label{fig:rats_ani_4}highlighting the stronger path from \varsc{Ecstazine} through \varsc{Neurofil}} 
    \end{subfigure}\vspace*{5mm}
   
\begin{subfigure}[t]{0.37\textwidth}
    \includegraphics[width=\textwidth]{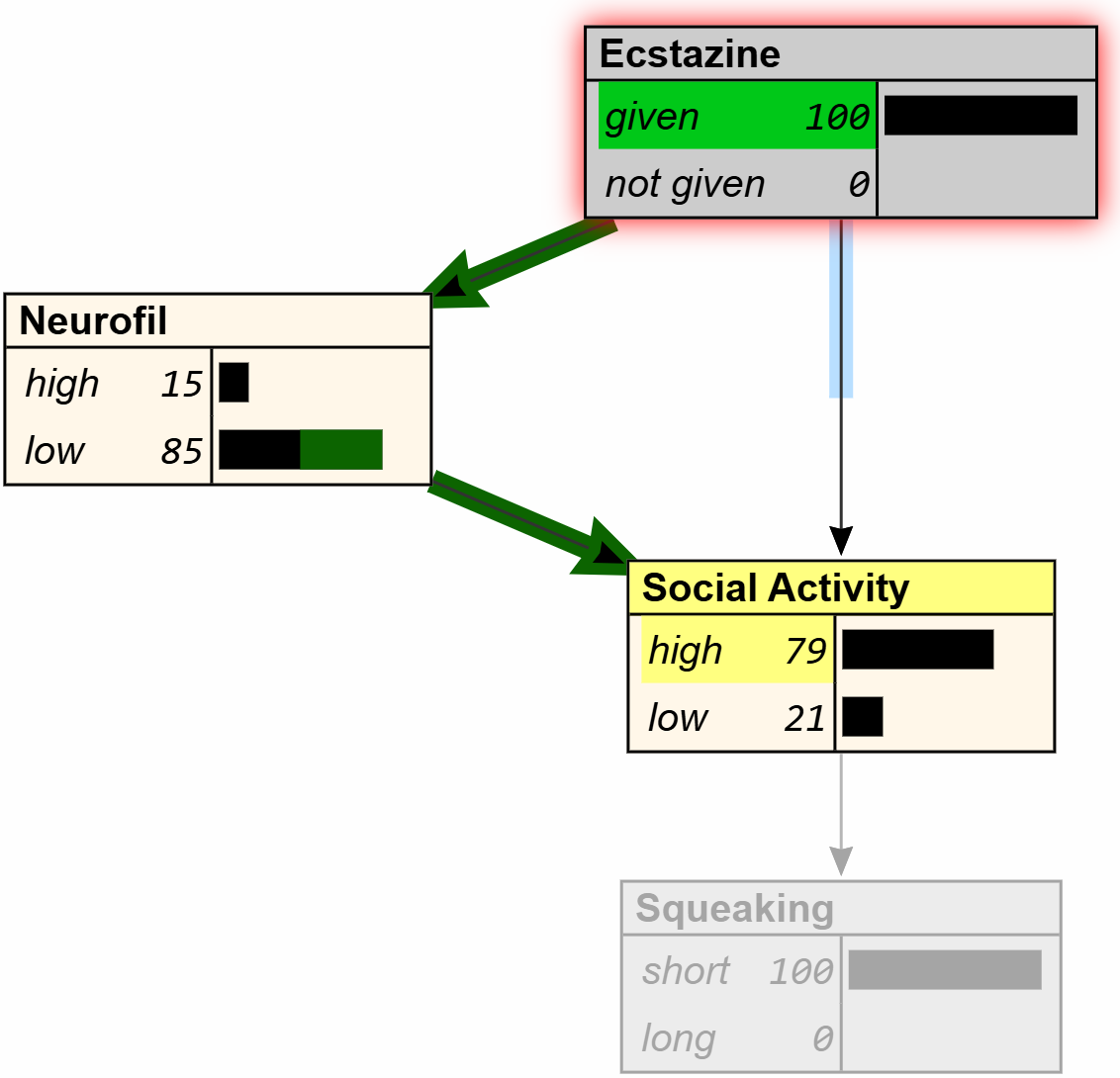}
    \caption{\label{fig:rats_ani_5}after completing the stronger path, highlighting the weaker path directly to \varsc{Social$\,$Activity}}
    \end{subfigure}
    \hspace*{0.13\textwidth}
\begin{subfigure}[t]{0.37\textwidth}
    \includegraphics[width=\textwidth]{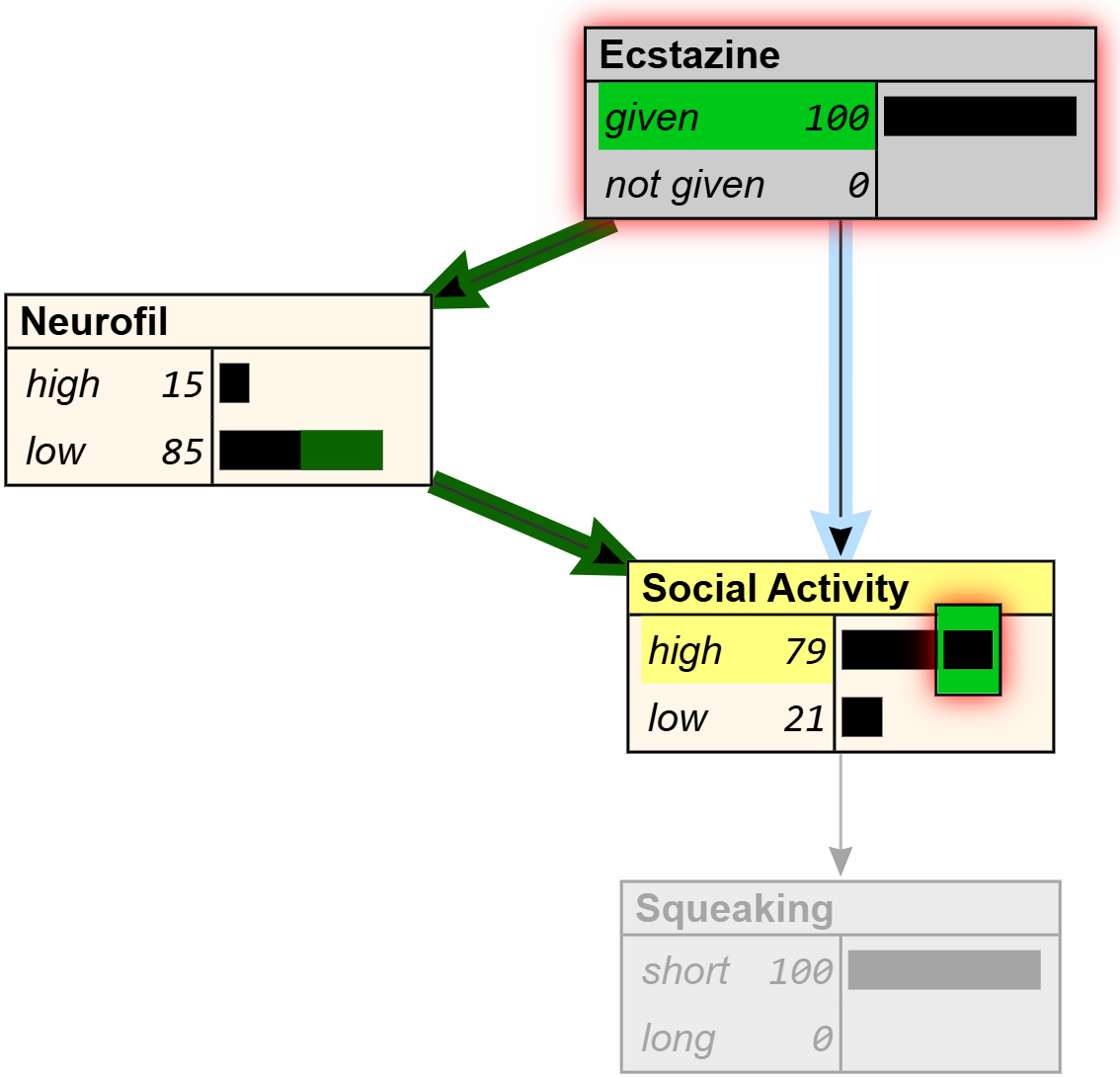}
    \caption{\label{fig:rats_ani_6}with the total probability change due to both paths highlighted on the target state's probability bar} 
    \end{subfigure}
  \caption{\label{fig:rats_ani}Snapshots from the animation that shows how finding out that \varft{Ecstazine} was given contributes. The animation starts when the play button is pressed, and ends at (f) after progressively redrawing the finding and the associated highlighting shown in Figure~\ref{fig:rats_ecstazine_A}.}
\end{figure}
\begin{figure}[tbp]
\centering
\begin{subfigure}[t]{0.9\textwidth}
    \includegraphics[trim={0 0 0 15mm},clip,width=\textwidth]{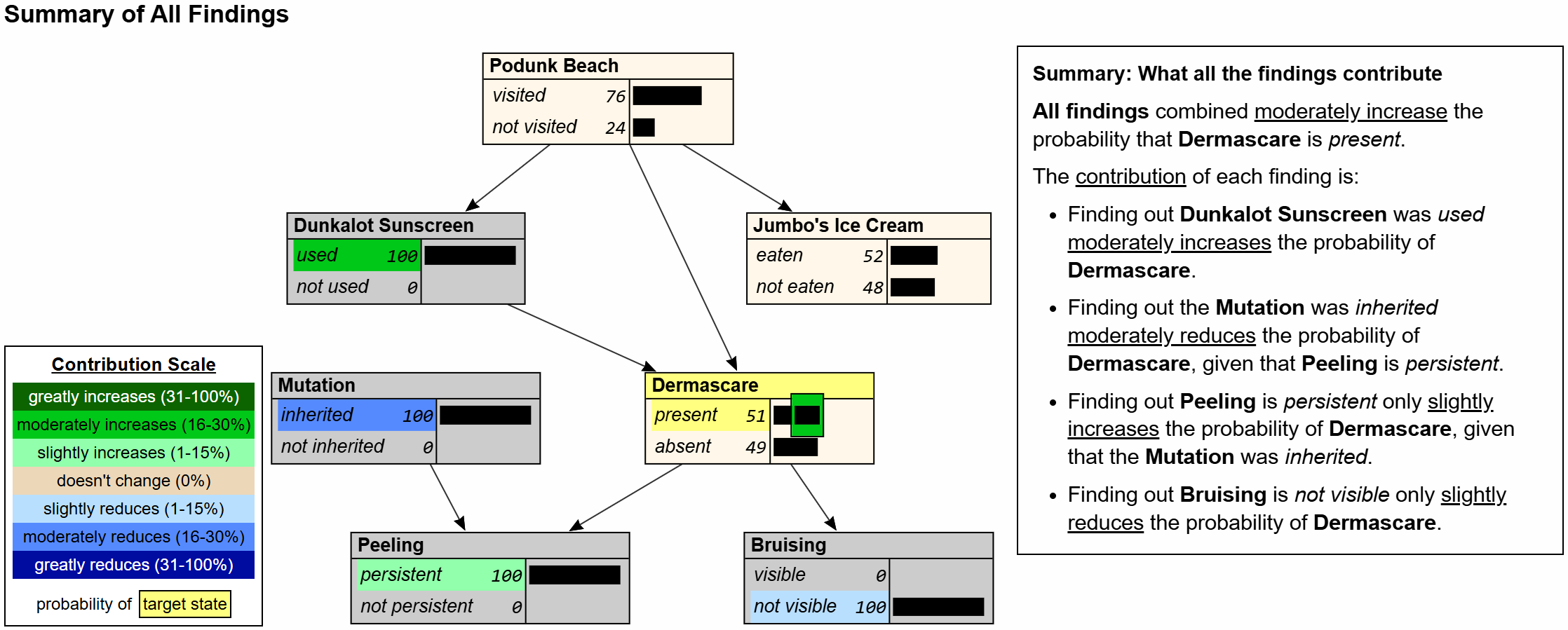}\vspace*{2ex}\\
        \vspace*{-0.40\textwidth}\\ {\huge\color{red}{$A$}} \vspace*{0.33\textwidth}\\
    \includegraphics[width=0.82\textwidth]{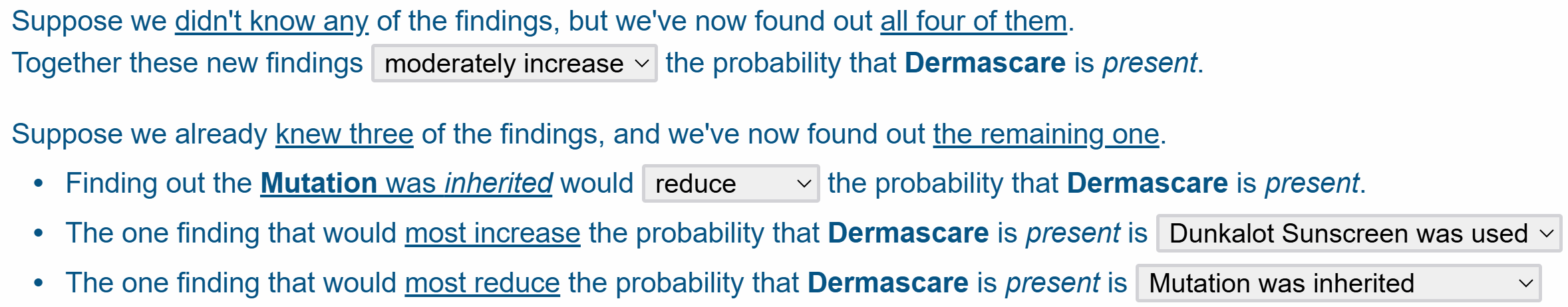}\\
        \vspace*{-0.18\textwidth}\\
        \hspace*{0.85\textwidth}\qtypelabel{Control}\vspace*{0.035\textwidth}\\
        \hspace*{0.85\textwidth}\qtypelabel{Finding}\vspace*{-0.003\textwidth}\\
        \hspace*{0.85\textwidth}\qtypelabel{Finding}\vspace*{-0.0025\textwidth}\\
        \hspace*{0.85\textwidth}\qtypelabel{Finding}\vspace*{1ex}    
    \caption{\label{fig:podunk_sum_A}Version $A$: Inheriting the \varsc{Mutation} explains away persistent \varsc{Peeling}, which makes persistent \varsc{Peeling} weaker evidence than using \varsc{Sunscreen}. Inheriting the \varsc{Mutation} thereby \emph{reduces} the probability of \varsc{Dermascare}, and does so more than the lack of visible \varsc{Bruising}.}
\end{subfigure}

\vspace*{4ex}
\begin{subfigure}[t]{0.9\textwidth}
    \includegraphics[trim={0 0 0 15mm},clip,width=\textwidth]{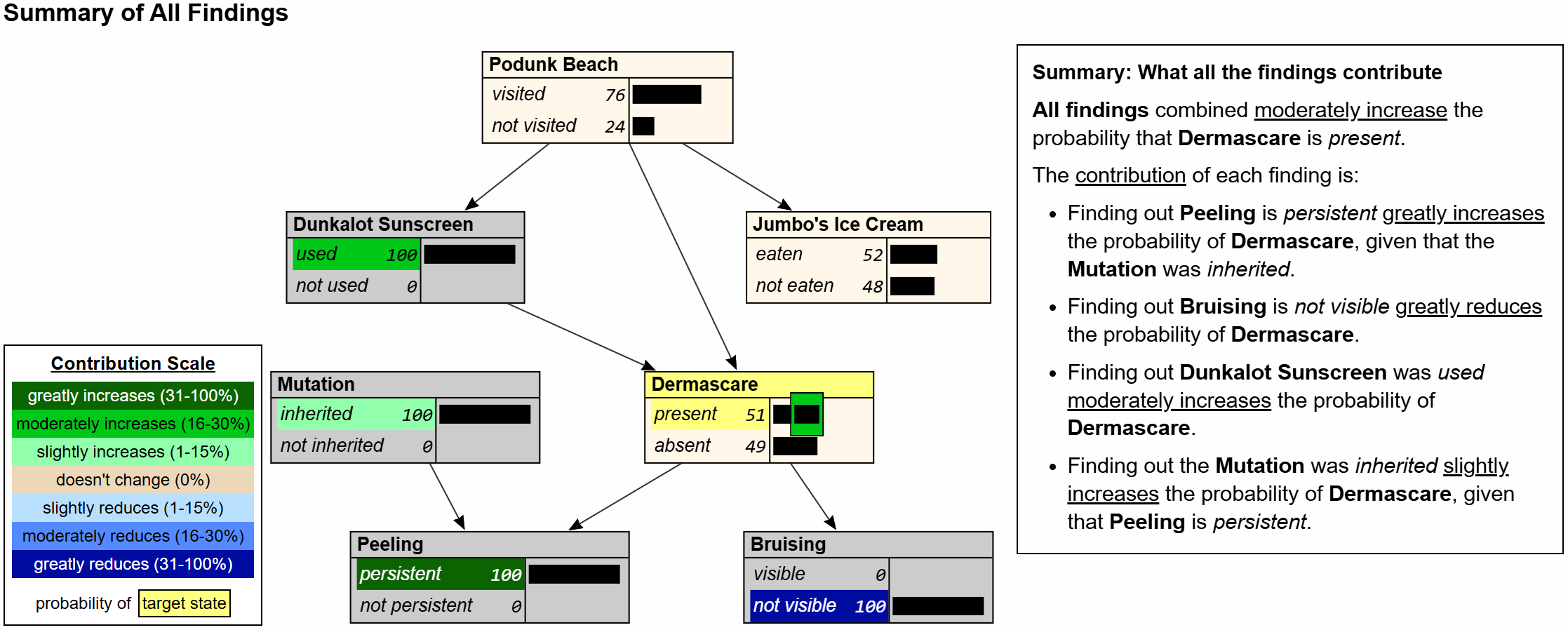}\vspace*{2ex}\\
        \vspace*{-0.40\textwidth}\\ {\huge\color{red}{$B$}} \vspace*{0.33\textwidth}\\
    \includegraphics[width=0.82\textwidth]{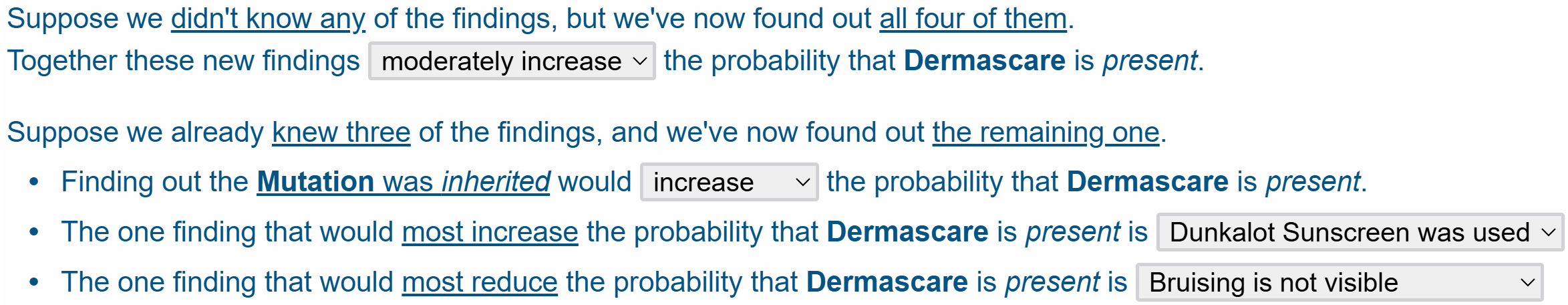}\\
        \vspace*{-0.18\textwidth}\\
        \hspace*{0.85\textwidth}\qtypelabel{Control}\vspace*{0.035\textwidth}\\
        \hspace*{0.85\textwidth}\qtypelabel{Finding}\vspace*{-0.003\textwidth}\\
        \hspace*{0.85\textwidth}\qtypelabel{Finding}\vspace*{-0.0025\textwidth}\\
        \hspace*{0.85\textwidth}\qtypelabel{Finding}\vspace*{1ex}    
    \caption{\label{fig:podunk_sum_B}Version $B$: Inheriting the \varsc{Mutation} increases the significance of persistent \varsc{Peeling}, which makes persistent \varsc{Peeling} stronger evidence than using \varsc{Sunscreen}. Inheriting the \varsc{Mutation} thereby \emph{increases} the probability of \varsc{Dermascare}, which is now only reduced by the lack of visible \varsc{Bruising}.}
\end{subfigure}
\caption{\label{fig:podunk_sum}Podunk section, Summary page: This shows the contributions of all findings to the probability of \varft{Dermascare}. The visual and verbal explanations, and the answers to three test questions, differ between versions $A$ and  $B$. Red markup indicates the question types: \qtype{Control} and \qtype{Finding}.}
\end{figure}
\begin{figure}[tbp]
\centering
\begin{subfigure}[t]{0.9\textwidth}
    \includegraphics[trim={0 30mm 0 15mm},clip,width=\textwidth]{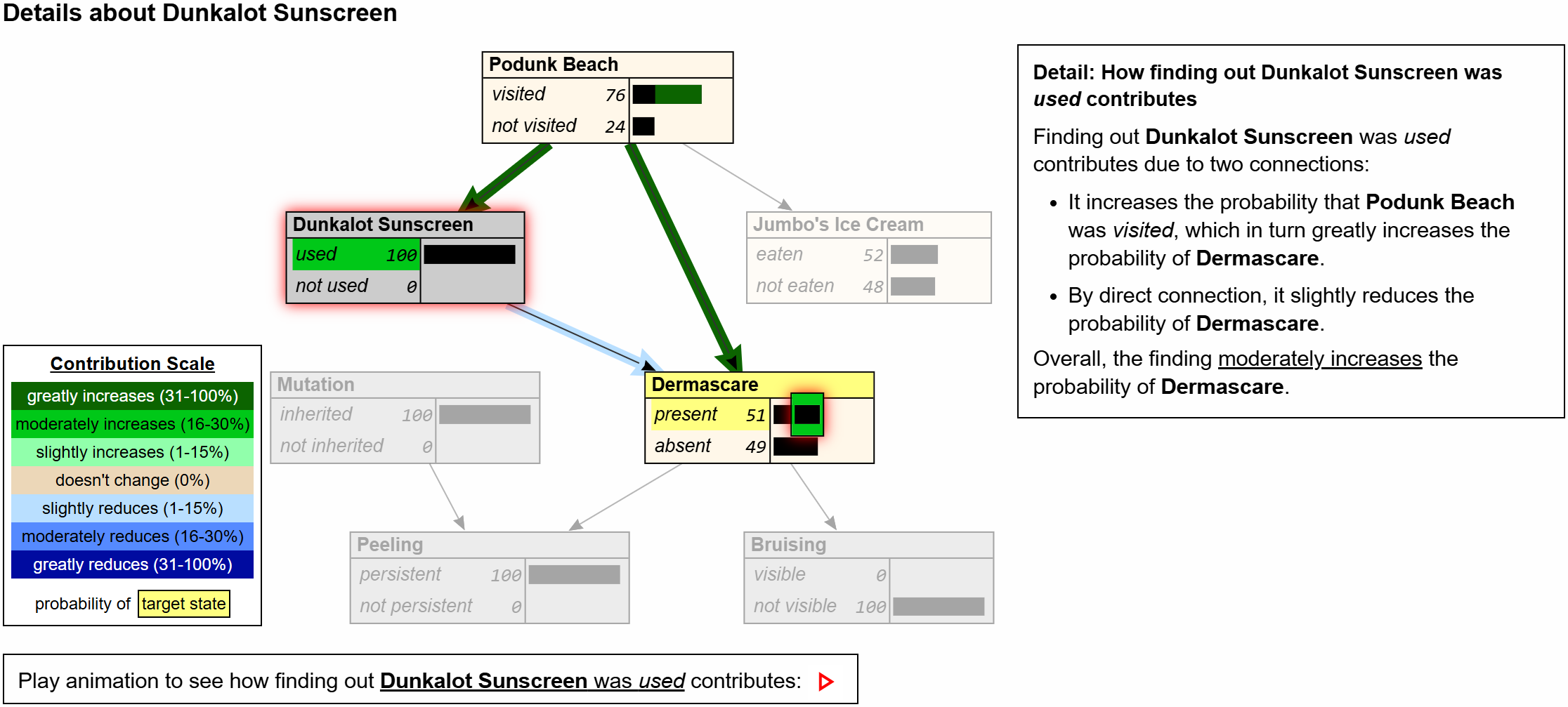}\vspace*{2ex}\\
        \vspace*{-0.41\textwidth}\\ {\huge\color{red}{$A$}} \vspace*{0.34\textwidth}\\
    \includegraphics[width=0.65\textwidth]{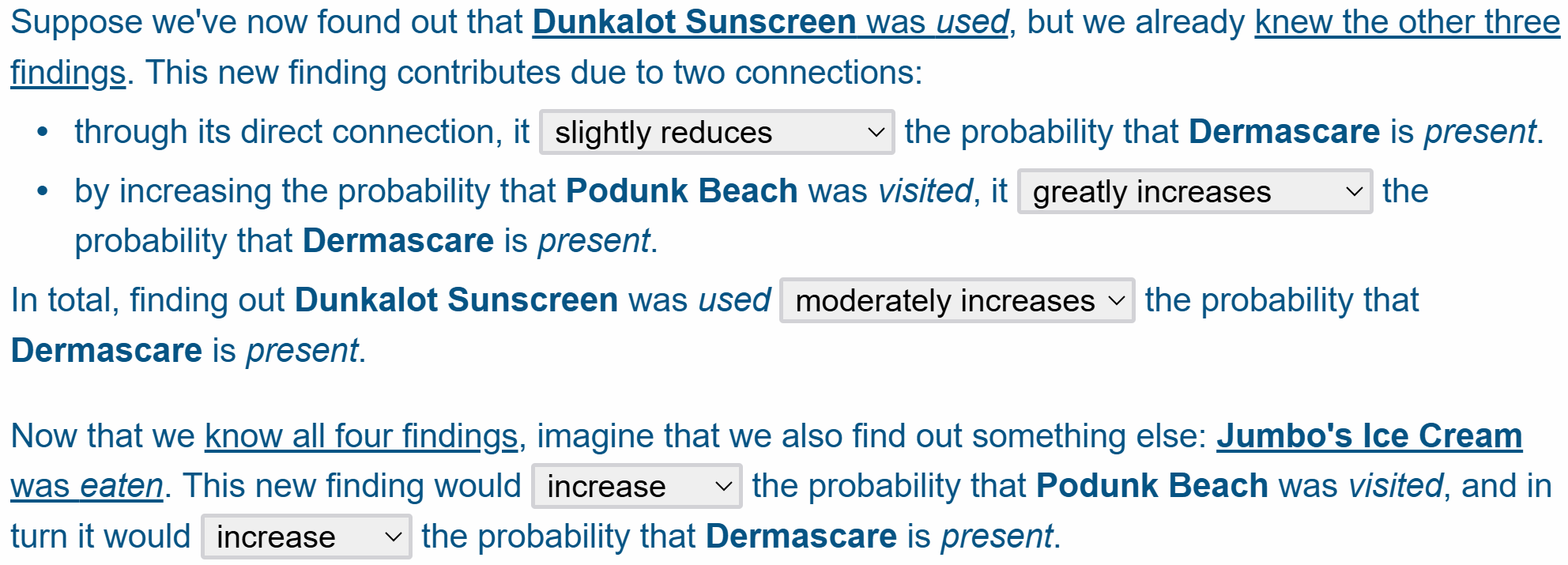}\\
        \vspace*{-0.2318\textwidth}\\
        \hspace*{0.7\textwidth}\qtypelabel{Path}\vspace*{-0.003\textwidth}\\
        \hspace*{0.7\textwidth}\qtypelabel{Path}\vspace*{0.016\textwidth}\\
        \hspace*{0.7\textwidth}\qtypelabel{Finding}\vspace*{0.048\textwidth}\\
        \hspace*{0.7\textwidth}\qtypelabel{ControlThink}\vspace*{-0.006\textwidth}\\
        \hspace*{0.7\textwidth}\qtypelabel{PathThink}\vspace*{1ex}
    \caption{\label{fig:podunk_sunscreen_A}Version $A$: The indirect path via \varsc{Podunk Beach} 
    is stronger than the direct path.} 
\end{subfigure}

\vspace*{4ex}
\begin{subfigure}[t]{0.9\textwidth}
    \includegraphics[trim={0 30mm 0 15mm},clip,width=\textwidth]{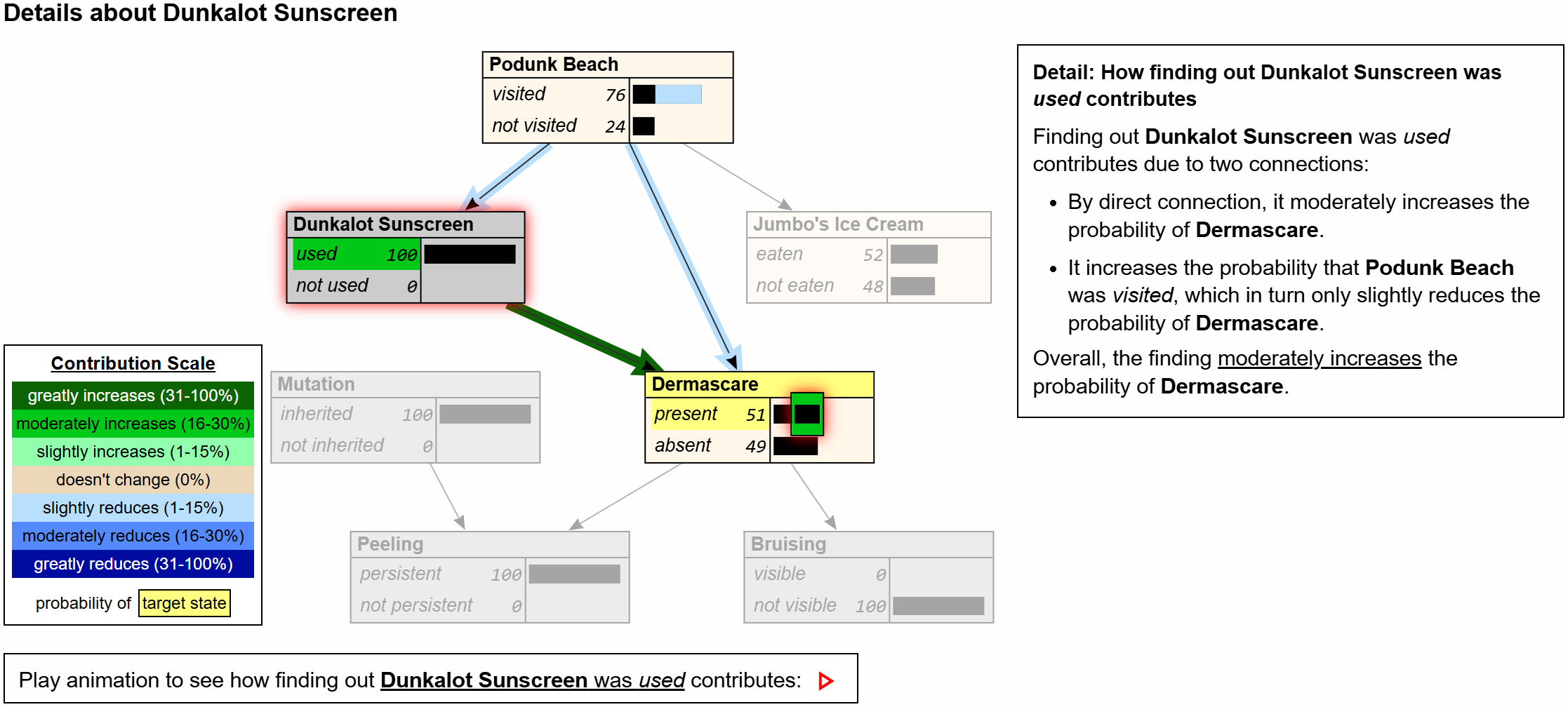}\vspace*{2ex}\\
    \vspace*{-0.41\textwidth}\\ {\huge\color{red}{$B$}} \vspace*{0.34\textwidth}\\
    \includegraphics[width=0.65\textwidth]{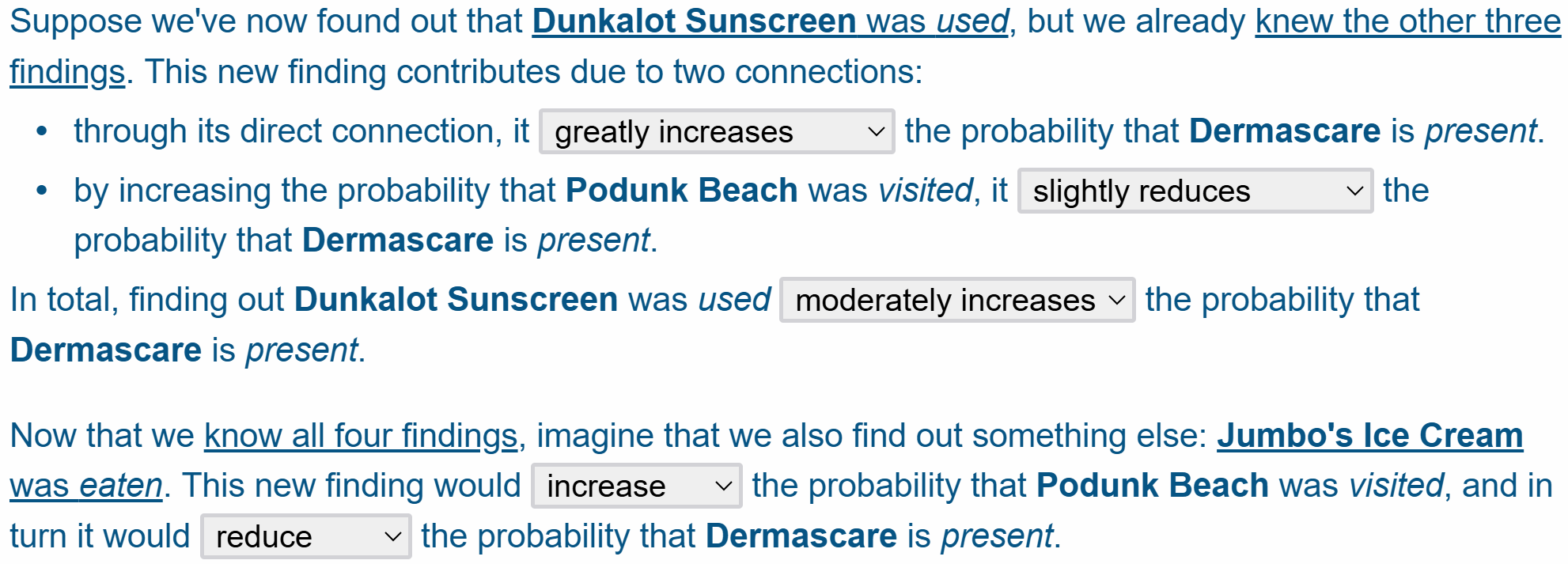}\\
        \vspace*{-0.2318\textwidth}\\
        \hspace*{0.7\textwidth}\qtypelabel{Path}\vspace*{-0.003\textwidth}\\
        \hspace*{0.7\textwidth}\qtypelabel{Path}\vspace*{0.016\textwidth}\\
        \hspace*{0.7\textwidth}\qtypelabel{Finding}\vspace*{0.048\textwidth}\\
        \hspace*{0.7\textwidth}\qtypelabel{ControlThink}\vspace*{-0.006\textwidth}\\
        \hspace*{0.7\textwidth}\qtypelabel{PathThink}\vspace*{1ex}
    \caption{\label{fig:podunk_sunscreen_B}Version $B$: The direct path is stronger than the indirect path via \varsc{Podunk Beach}.} 
\end{subfigure}
\caption{\label{fig:podunk_sunscreen_AB}Dermascare section, \varft{Dunkalot Sunscreen} page: The visual and verbal explanations for the contribution of \varft{Sunscreen}, and the answers to four test questions, differ between versions $A$ and  $B$. The animation button is not shown. Red markup indicates the question types:  \qtype{Path}, \qtype{Finding}, \qtype{ControlThink} and \qtype{PathThink}.}
\end{figure}
\begin{figure}[tbp]
\centering
\begin{subfigure}[t]{0.9\textwidth}
    \includegraphics[trim={0 0 0 15mm},clip,width=\textwidth]{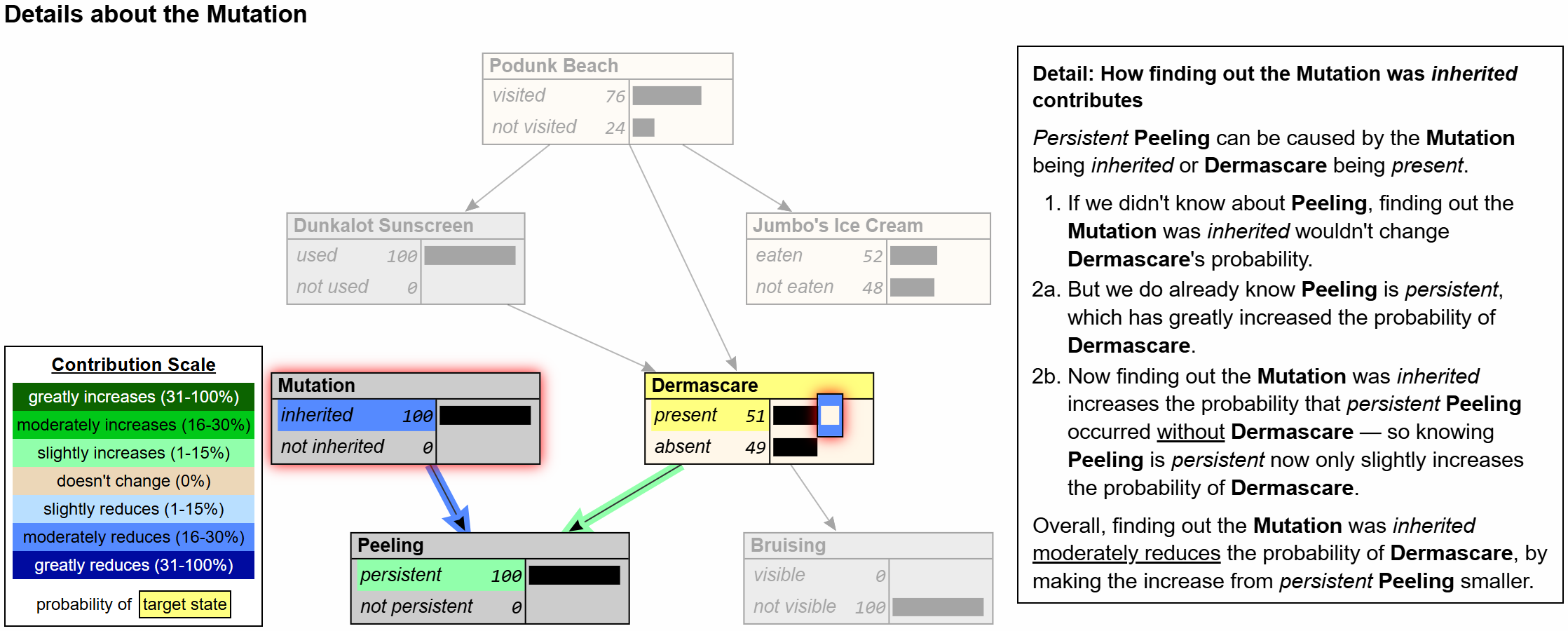}\vspace*{2ex}\\
        \vspace*{-0.41\textwidth}\\ {\huge\color{red}{$A$}} \vspace*{0.34\textwidth}\\
    \includegraphics[width=0.7\textwidth]{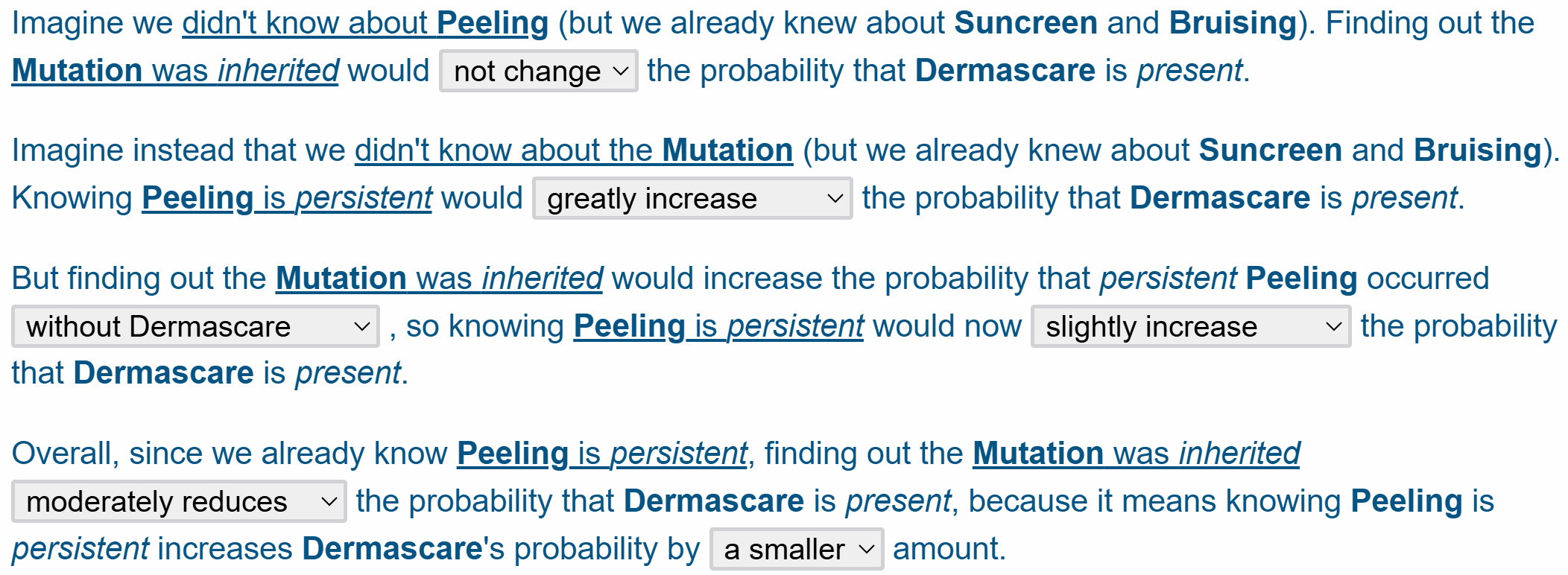}\\
        \vspace*{-0.278\textwidth}\\
        \hspace*{0.75\textwidth}\qtypelabel{ControlThink}\vspace*{0.028\textwidth}\\
        \hspace*{0.75\textwidth}\qtypelabel{CommonEffect}\vspace*{0.0285\textwidth}\\
        \hspace*{0.75\textwidth}\qtypelabel{CommonEffect, Finding}\vspace*{0.050\textwidth}\\
        \hspace*{0.75\textwidth}\qtypelabel{Finding}\vspace*{-0.007\textwidth}\\
        \hspace*{0.75\textwidth}\qtypelabel{CommonEffect}\vspace*{1ex}
    \caption{\label{fig:podunk_mutation_A}Version $A$: Inheriting the \varsc{Mutation} \emph{reduces} the probability of \varsc{Dermascare} by \emph{weakening} (via explaining away) the persistent \varsc{Peeling} evidence.}
\end{subfigure}

\vspace*{1.5ex}
\begin{subfigure}[t]{0.9\textwidth}
    \includegraphics[trim={0 0 0 15mm},clip,width=\textwidth]{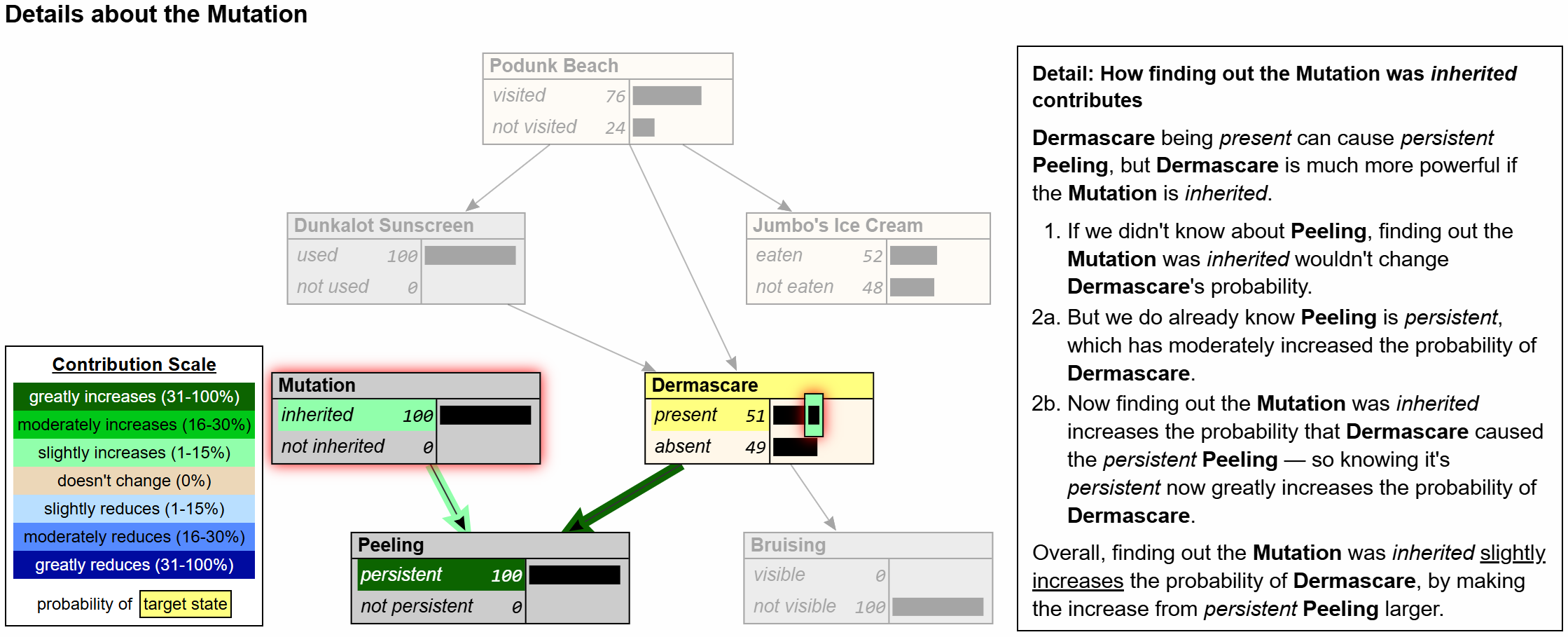}\vspace*{2ex}\\
        \vspace*{-0.41\textwidth}\\ {\huge\color{red}{$B$}} \vspace*{0.34\textwidth}\\
    \includegraphics[width=0.7\textwidth]{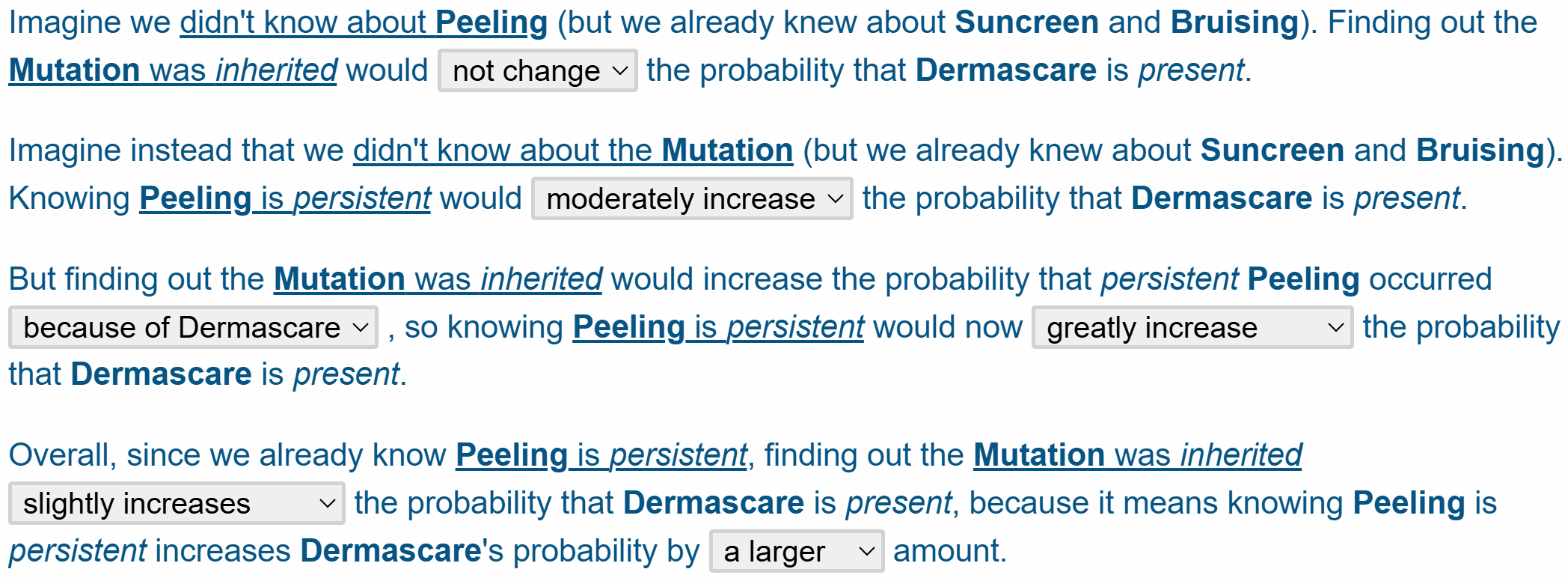}\\
        \vspace*{-0.278\textwidth}\\
        \hspace*{0.75\textwidth}\qtypelabel{ControlThink}\vspace*{0.028\textwidth}\\
        \hspace*{0.75\textwidth}\qtypelabel{CommonEffect}\vspace*{0.0285\textwidth}\\
        \hspace*{0.75\textwidth}\qtypelabel{CommonEffect, Finding}\vspace*{0.050\textwidth}\\
        \hspace*{0.75\textwidth}\qtypelabel{Finding}\vspace*{-0.007\textwidth}\\
        \hspace*{0.75\textwidth}\qtypelabel{CommonEffect}\vspace*{1ex}
    \caption{\label{fig:podunk_mutation_B}Version $B$: Inheriting the \varsc{Mutation} \emph{increases} the probability of \varsc{Dermascare} by \emph{strengthening} (via enabling) the persistent \varsc{Peeling} evidence.}
\end{subfigure}
\caption{\label{fig:podunk_mutation}Dermascare section, \varft{Mutation} page: The visual and verbal explanations for the contribution of \varft{Mutation}, and the answers to five test questions, differ between versions $A$ and  $B$. The animation buttons are not shown.
Red markup indicates the question types: \qtype{ControlThink}, \qtype{CommonEffect} and \qtype{Finding}.}
\end{figure}
\begin{figure}[htbp]
\centering
\begin{subfigure}[t]{0.45\textwidth}
    \includegraphics[width=\textwidth]{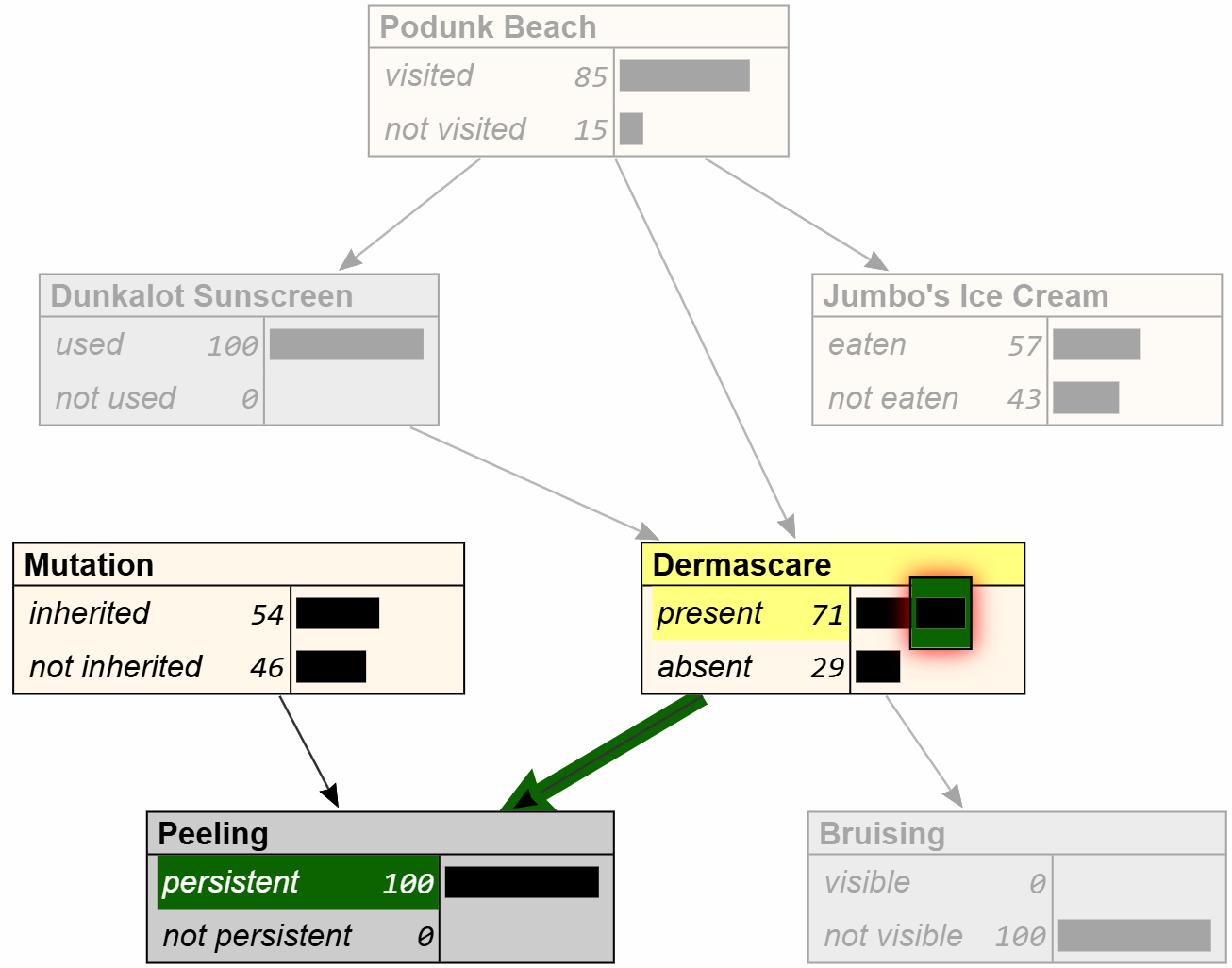}
    \caption{\label{fig:mutation_ani_1}starting without the finding or focus on \varsc{Mutation}} 
    \end{subfigure}
    \hspace*{0.05\textwidth}
\begin{subfigure}[t]{0.45\textwidth}
    \includegraphics[width=\textwidth]{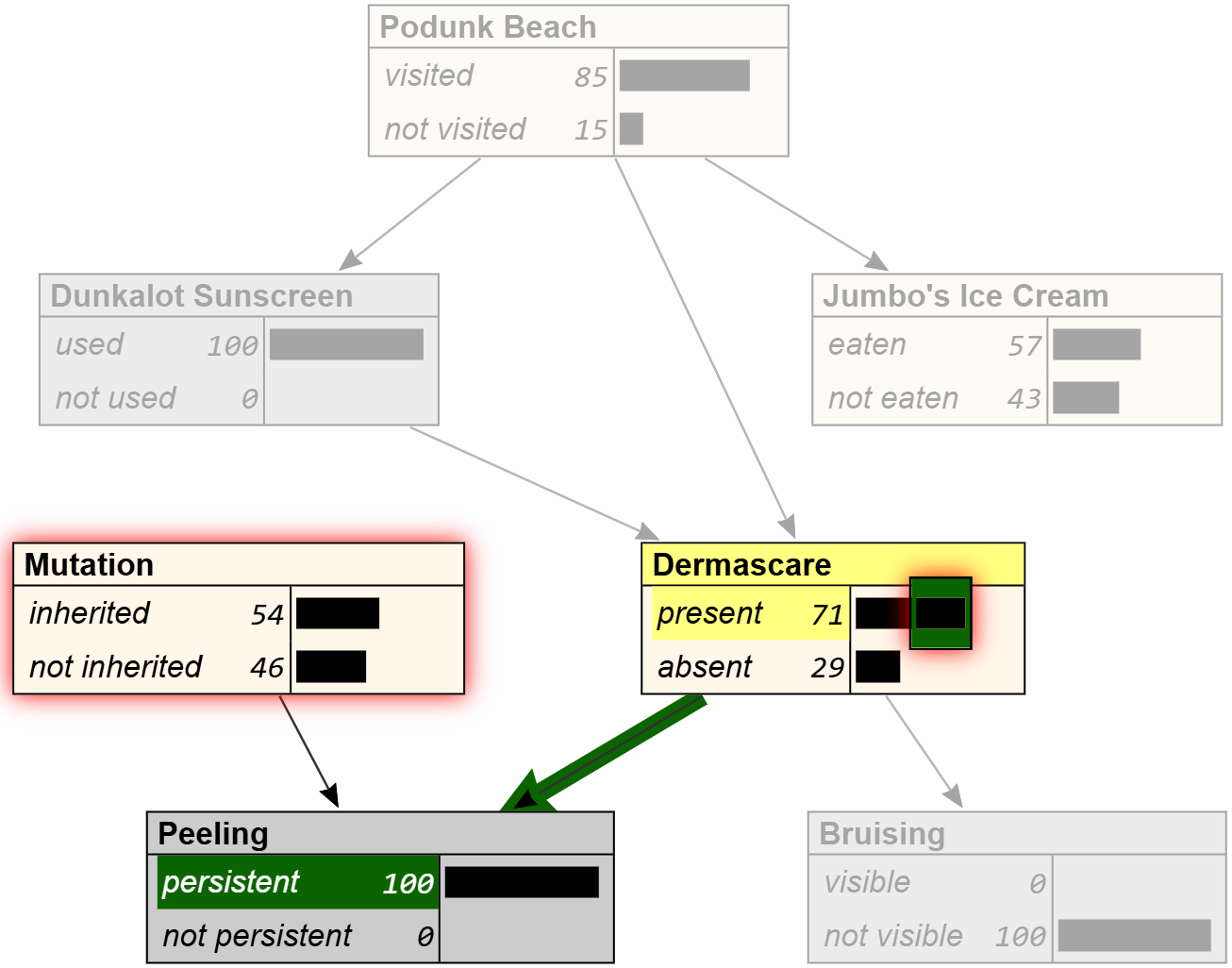}
    \caption{\label{fig:mutation_ani_2}with focus highlighting added to \varsc{Mutation}} 
     \end{subfigure}\vspace*{4mm}
    
\begin{subfigure}[t]{0.45\textwidth}
    \includegraphics[width=\textwidth]{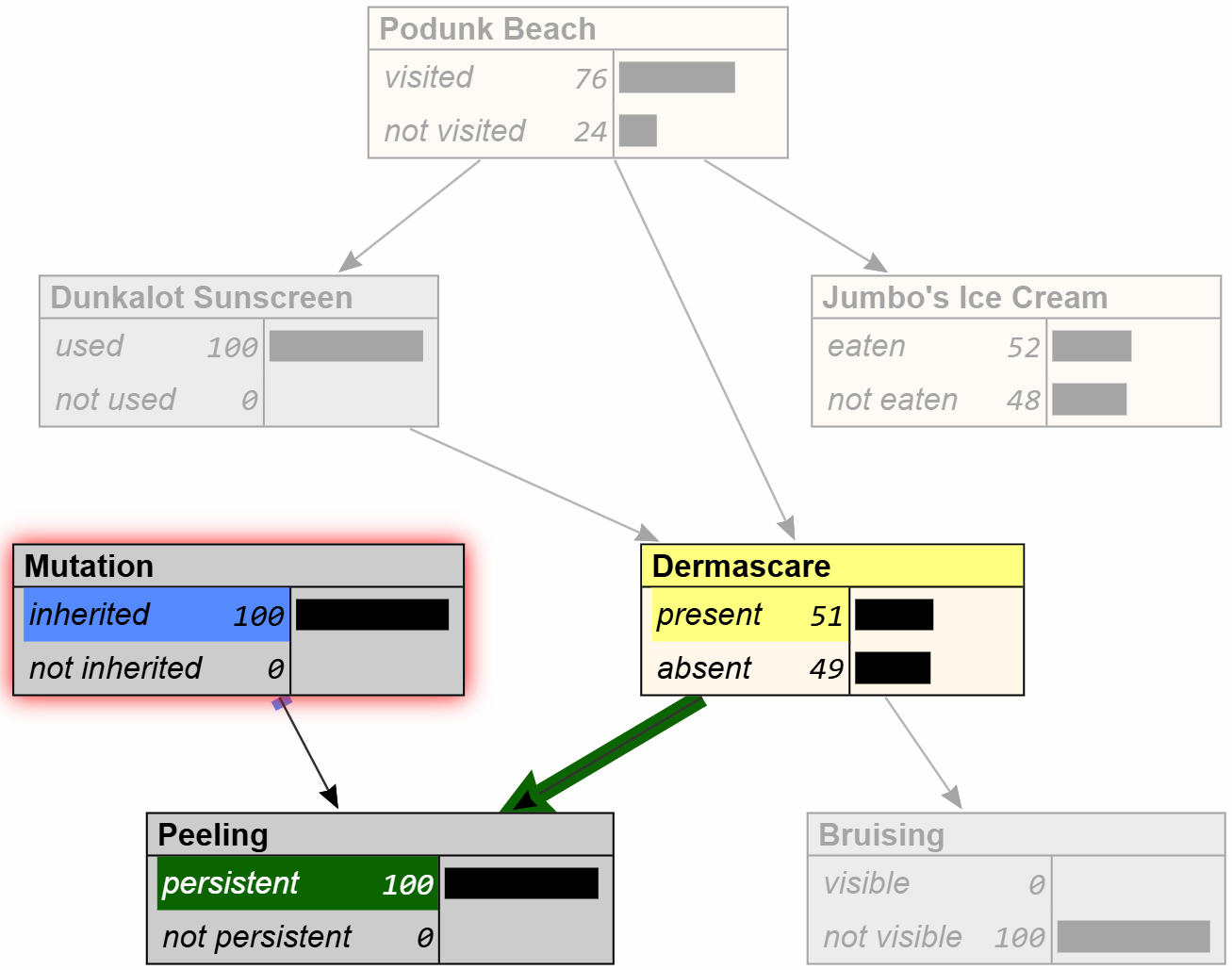}
    \caption{\label{fig:mutation_ani_3}with the finding that \varsc{Mutation} was inherited added and highlighted}
    \end{subfigure}
    \hspace*{0.05\textwidth}
\begin{subfigure}[t]{0.45\textwidth}
    \includegraphics[width=\textwidth]{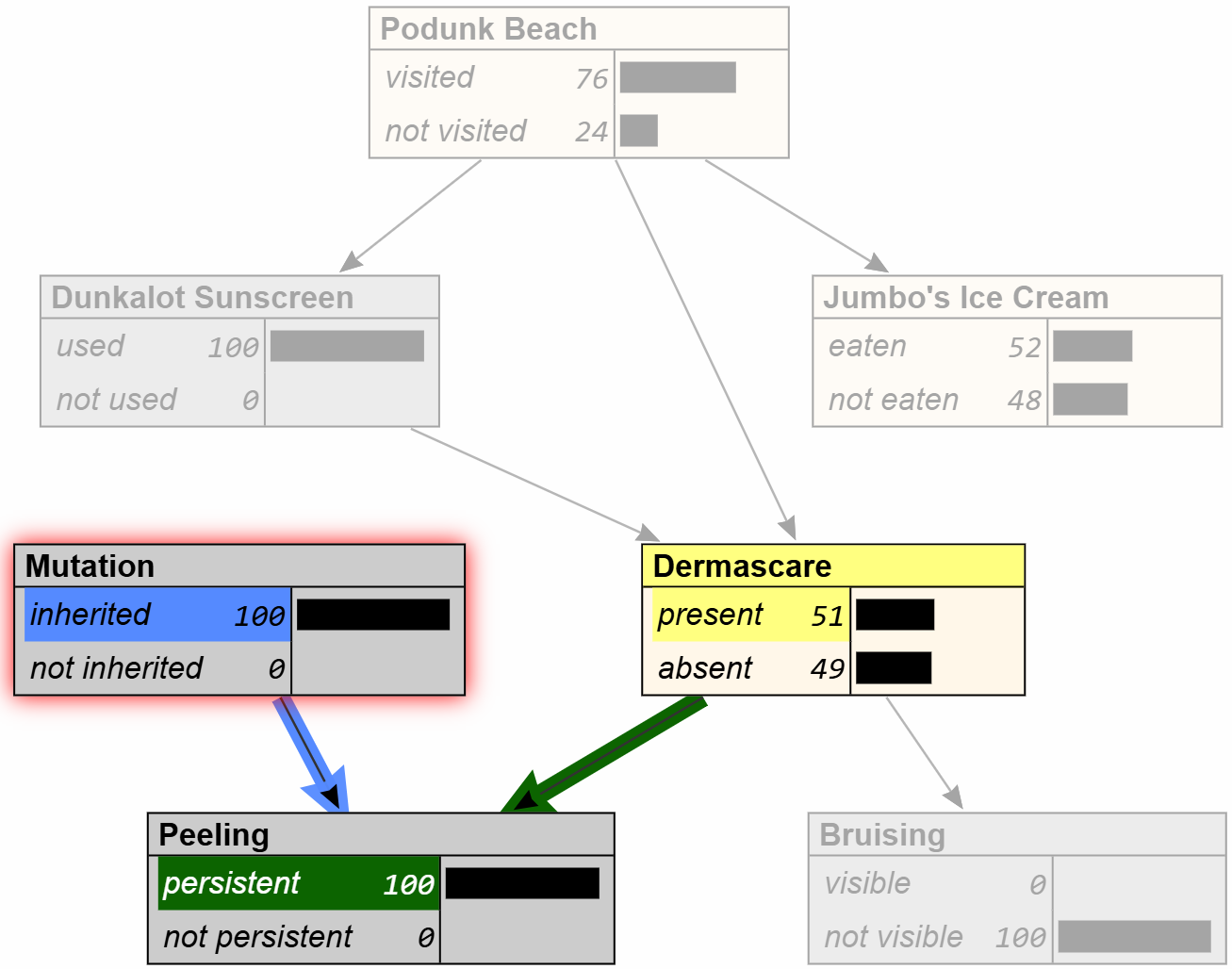}
   \caption{\label{fig:mutation_ani_4}with the arc from \varsc{Mutation} to \varsc{Peeling} highlighted}
   \end{subfigure}\vspace*{4mm}
   
\begin{subfigure}[t]{0.45\textwidth}
    \includegraphics[width=\textwidth]{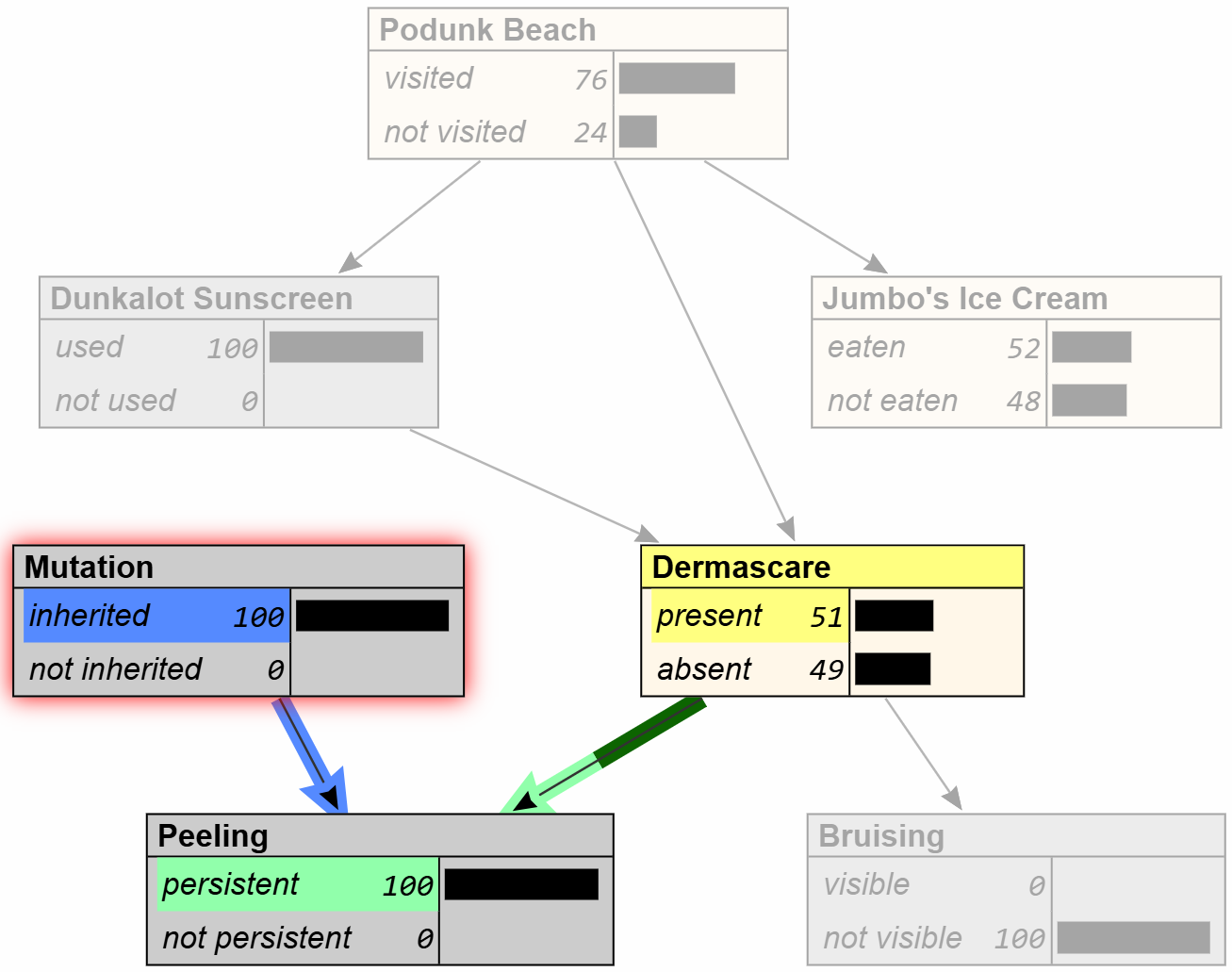}
    \caption{\label{fig:mutation_ani_5}overwriting the highlighting on \varsc{Peeling} and its path to \varsc{Dermascare}}
    \end{subfigure}
    \hspace*{0.05\textwidth}
\begin{subfigure}[t]{0.45\textwidth}
    \includegraphics[width=\textwidth]{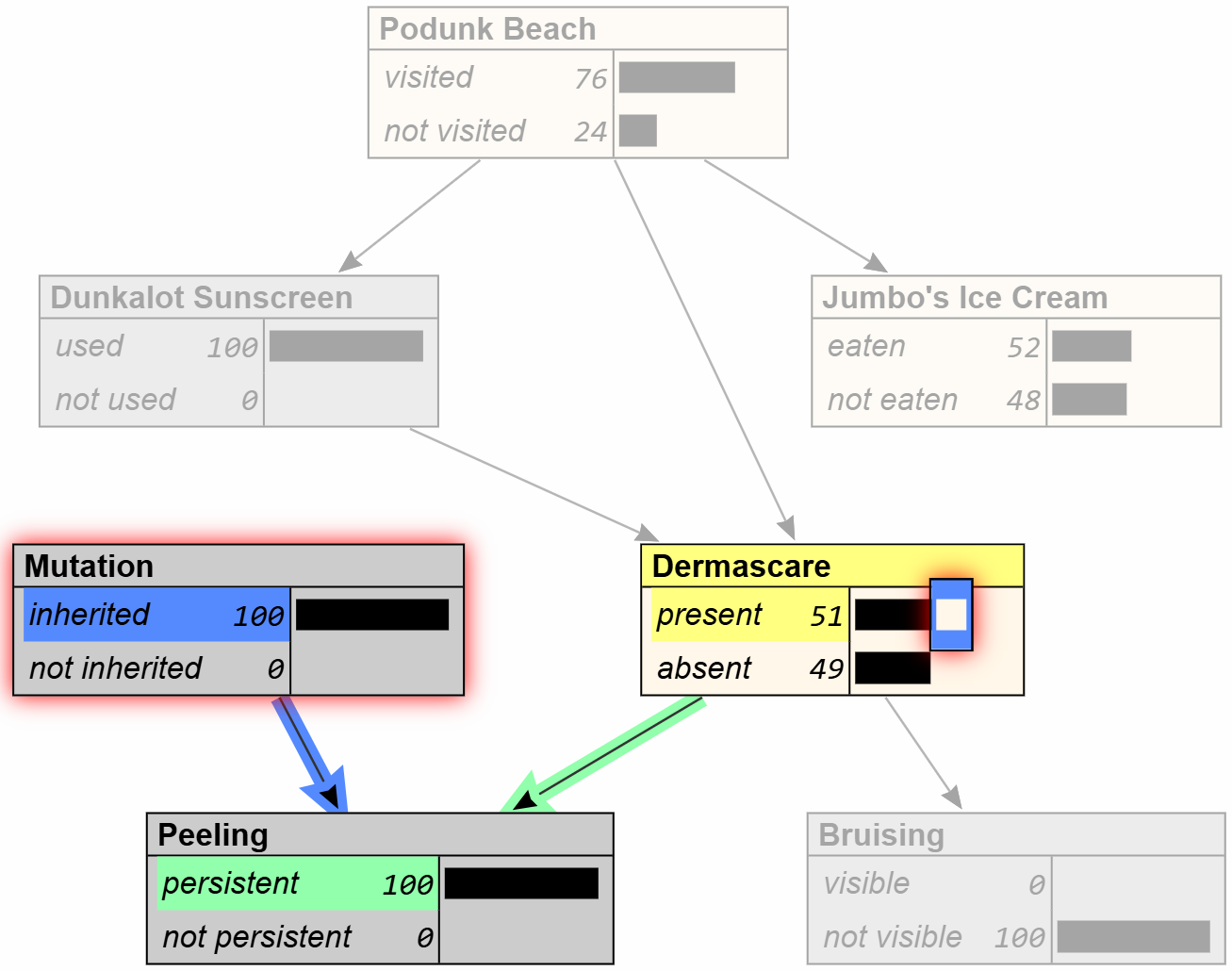}
    \caption{\label{fig:mutation_ani_6}with the probability change due to the effect of the highlighted \varsc{Mutation} finding on the target state's probability bar}
    \end{subfigure}
  \caption{\label{fig:mutation_ani}Snapshots from the animation that shows how finding out that the \varft{Mutation} was inherited contributes. The animation starts at~(a) showing only the interacting finding that \varft{Peeling} is persistent and its associated highlighting, and ends at~(f) after progressively redrawing the \varft{Mutation} finding and the associated highlighting shown in Figure~\ref{fig:podunk_mutation_A}.}
\end{figure}

Figure~\ref{fig:explan_statement} displays a screenshot from the introduction to the experiment. 
Figures \ref{fig:rats_sum}--
\ref{fig:rats_ani} display screenshots from the Rats section,
and Figures \ref{fig:podunk_sum}--
\ref{fig:mutation_ani} display screenshots from the Podunk section.
All BN pages except the Summaries 
also include animations that must be activated to continue (via play buttons per Figures \ref{fig:podunk_sunscreen} and \ref{fig:podunk_peeling}).

\FloatBarrier

\clearpage

\begin{figure}[htbp]
    \centering
    \begin{subfigure}[t]{0.9\textwidth} 
    \centering
    \includegraphics[width=\textwidth]{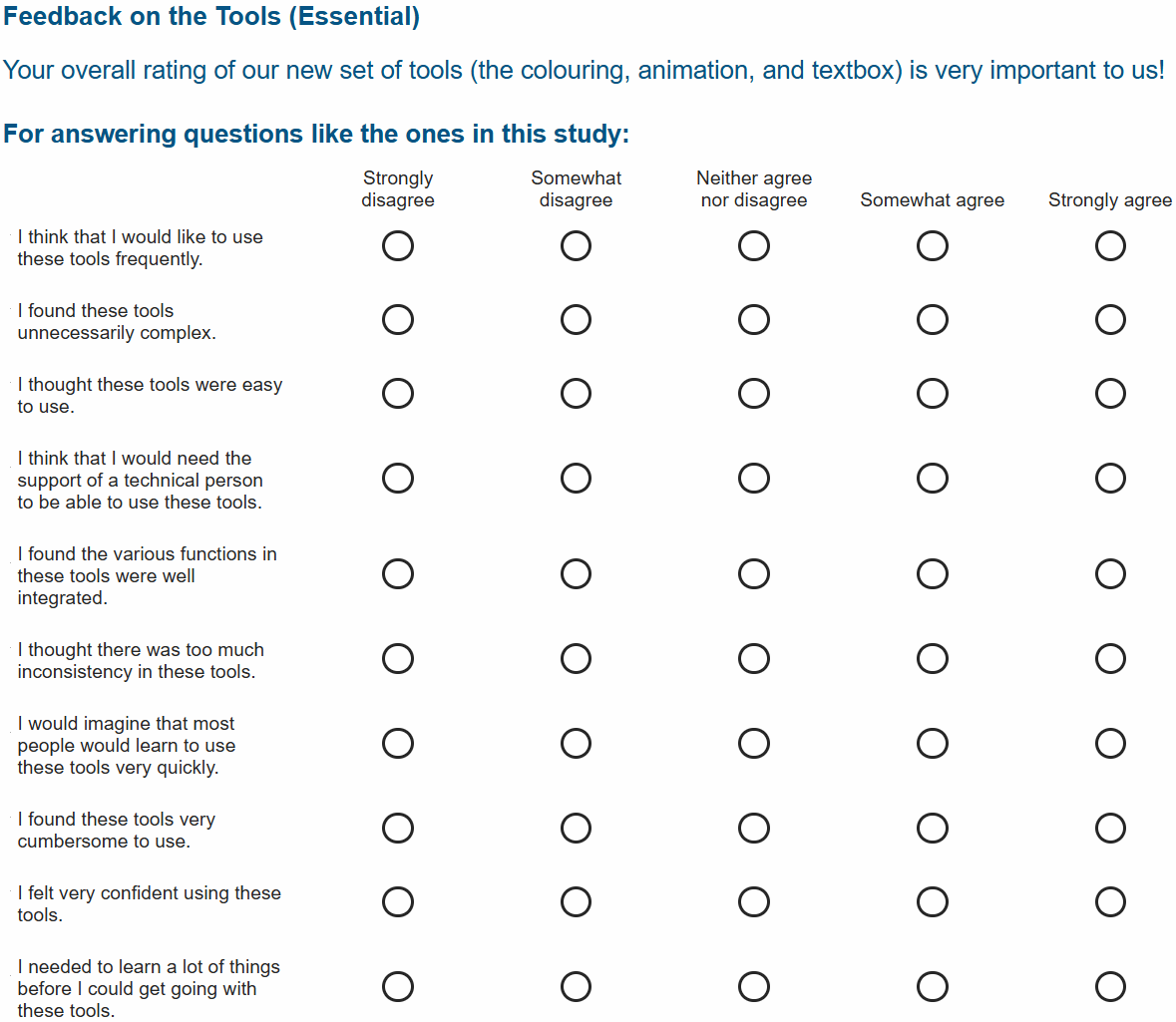}
    \vspace*{0ex}
    \caption{\label{fig:feedback_SUS}Post-study form: participants rate their experience with the explanatory tools in the study (on a 5-point Likert scale), including ease of use, confidence, complexity and integration, using the System Usability Scale (SUS).}
    \end{subfigure}
    
    \vspace*{4ex}
    \begin{subfigure}[t]{0.9\textwidth} 
    \centering
    \includegraphics[width=\textwidth]{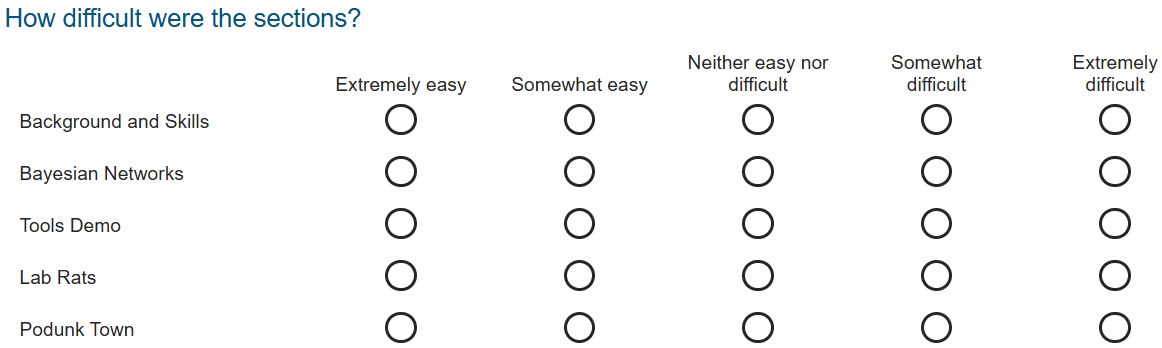}
    \vspace*{0ex}
    \caption{\label{fig:feedback_sections}Optional post-study form: participants rate the difficulty of
    each of the five sections on a 5-point Likert scale.}
    \end{subfigure}
    \caption{\label{fig:feedback}Post-study feedback forms.}
\end{figure}

\clearpage

%% file: AppendixE-Results.tex


\input tab-results-distributionV2

\myparagraph{Relationship between experiment duration and marks per section} 
Table~\ref{tab:marks} shows the median and interquartile ranges for 
overall duration and for participant marks per section. 
The median duration was approximately 54 minutes with an interquartile range of 
43--78 minutes. 

The performance of participants who proceeded to the main study was
highest for the Car Battery BN tutorial (\ttest, statistically significant
with $\pvalue \ll 0.001$ compared to the Rats and Podunk BNs, and $\pvalue
< 0.05$ compared to the Explanatory Tools tutorial). This is expected,
as the Car Battery BN was the simplest, and participants were highly
motivated to succeed in order to remain in the experiment. The next
highest performance was for the Explanatory Tools tutorial (\ttest,
statistically significant with $\pvalue \ll 0.001$ compared to the Rats
and Podunk BNs). The median performance on the Rats questions was higher
than for Podunk (not statistically significant, $\pvalue =0.6$).

\begin{figure}[t]
    \centering
    \includegraphics[width=0.5\linewidth]{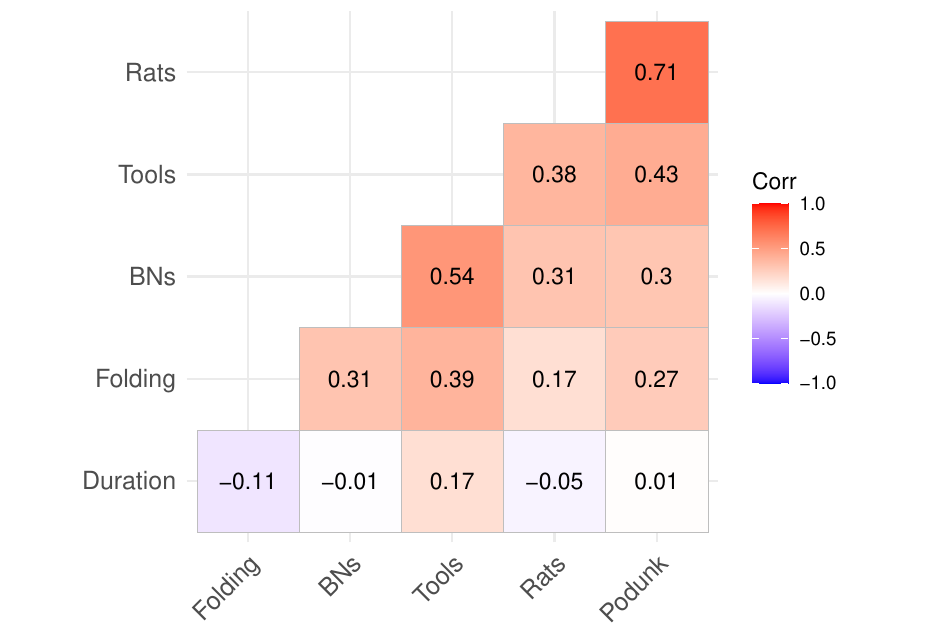}
    \vspace*{-3mm}
    \caption{Pairwise correlations between duration and section scores.}
    \label{fig:cor-section-scores}
\end{figure}

Figure~\ref{fig:cor-section-scores} shows Pearson correlations between
all section scores and overall experiment duration. Duration is very 
weakly correlated with the score obtained for all sections (bottom
row). However, section scores have varying degrees of correlation with
each other, with the strongest correlation between performance in the
Rats and Podunk sections.
Surprisingly, performance in the BN tutorial is only weakly correlated
with performance in the Rats and Podunk BNs; and performance in the
folding task, which is predictive of spatial reasoning, is weakly
correlated with the BN tutorial and the Podunk BN, and very
weakly correlated with the Rats BN.

Performance in the Explanatory Tools tutorial was most predictive of
performance in the experiment (both Rats and Podunk), where participants
used the explanation type on which they were trained. Performance in
the BN tutorial was the next most predictive, and folding performance
the least predictive of the three.

\begin{figure}[t]
    \centering
    \includegraphics[width=0.5\linewidth]{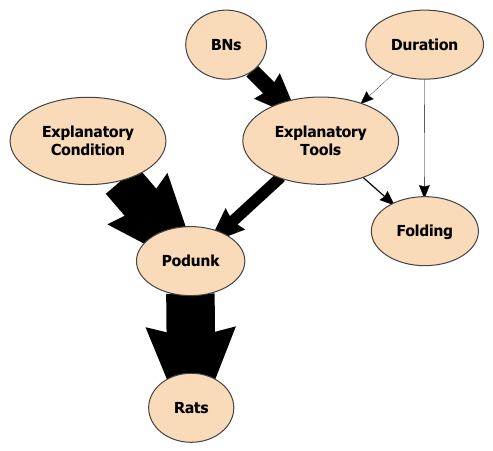}
    \caption{Relationships identified by bnlearn between key variables
    in the study. Demographic factors are omitted, as they were
    not associated with section performance. Arc width indicates the
    statistical confidence ($-\log(\pvalue)$) of bnlearn in an arc
    compared to ablating that arc.\label{fig:predictor-bnlearn-relations}}
    \vspace*{-3mm}
\end{figure}

\myparagraph{Relationship between demographic and skill factors,
experiment duration, explanatory condition and performance}
Figure~\ref{fig:predictor-bnlearn-relations} shows the result of using
the R bnlearn package to learn the conditional dependence relationships
between demographic and skill factors, experiment duration, explanatory
condition and performance in the experiment. The bnlearn model validates
the results in Figure~\ref{fig:cor-section-scores}, and adds the effect
of explanatory condition. It also shows that there are no statistically
significant differences in performance in the BN and Explanatory
Tools tutorials, and Rats and Podunk BNs for any demographic variable or SNS score
($\pvalue>0.1$).

\myparagraph{Relationship between SUS and explanatory condition} There
is an indirect effect of explanatory condition on SUS score via mark,
but little to no direct effect of explanatory condition on SUS score. In
other words, the effect of explanatory condition ($C$) on SUS score
($S$) is almost entirely mediated by mark ($M$). Specifically,
\begin{itemize}
\myitem $C \prec M \prec S$: That is, $C$ occurs before $M$ (by
necessity), and $M$ occurs before $S$ (by order of appearance).

\myitem
$C \rightarrow M$: $C \sim M$ (the association between $C$ and $M$)
is significant ($\pvalue < 0.001$ ANOVA; pairwise differences that are
significant appear in Table~\ref{tab:stat_sig}). In addition, $C$ has
been randomly assigned (and hence has no parent), allowing us to rule
out a common cause confounder, and hence infer $C \rightarrow M$.

\myitem 
$M \rightarrow S$: $M \sim S$ is significant ($\pvalue < 0.001$,
linear regression with no other adjustments). If we adjust for $C$
(for any combination of explanatory conditions), $M \sim S$ remains
significant with the same effect size, which rules out $C$ as a common
cause confounder of $M$ and $S$. In addition, there is no association
$C \sim S$ when adjusting for $M$. This implies that observing $M$ does
not {open} a path between $C$ and $S$, which in turn implies that
there is no {other} common cause confounder of $M$ and $S$. Since $M$
and $S$ have no common cause confounders, $M \sim S$ implies causation,
and the order implies $M \rightarrow S$.

\myitem $C \perp\!\!\!\perp S|M$ -- Finally, as above, adjusting for $M$
leaves no association between $C$ and $S$ ($\pvalue>0.1$), so there is
little to no direct effect $C \rightarrow S$.
\end{itemize}

We assumed above that $M \prec S$ because $M$ is measured before $S$,
but it is possible that system usability drives the mark (even if they
are measured the other way around). We can test this: if adjusting for
$S$ blocks $C \sim M$, then it may be that $C \rightarrow S \rightarrow
M$. However, $S$ does \emph{not} block $C \sim M$, rather they remain strongly
associated ($\pvalue < 0.001$). Hence, $C \rightarrow M \rightarrow S$.





%% file: tab-results-distributionV2.tex
\begin{table}[t]
\setlength{\tabcolsep}{9pt}
\centering
\small
\caption{Summary of median experiment duration and participant marks for each section -- 124 participants.} \label{tab:marks}
\begin{tabular}{lcccccc}
\toprule
& \textbf{Duration} & \multicolumn{5}{c}{\textbf{Section marks}}\\
\cmidrule{3-7}
&   & Folding & BN tutorial & Tools tutorial & Rats BN & Podunk BN \\
\midrule
Median & 53.55 & 0.50 & 0.76 & 0.73 & 0.62 & 0.52 \\
\hspace*{0mm}[IQR] & [42.84, 77.77] & [0.40, 0.80] & [0.71, 0.88] & [0.55, 0.91] & [0.44, 0.69] & [0.37, 0.78]\\ 
\bottomrule
\end{tabular}
\end{table}

%% file: AppendixF-LLMs.tex
\subsection{Similarity metrics}
To assess the similarity between the LLM-generated texts and the 
user-study explanations, and the similarity between LLM-generated
texts, we use three metrics: BLEU~\cite{PapineniEtAl2002Bleu} and 
METEOR~\citep{BanerjeeLavie2005Meteor}, which were developed for
machine translation, and BERTscore~\citep{ZhangEtAl2019Bertscore}.
The first two metrics measure the similarity between machine-produced 
translations and reference translations.
BLEU works by matching n-grams to the reference text, balancing
precision with a brevity penalty. It works well
when wording and structure are very similar or identical to the 
reference text, but penalises paraphrasing. In contrast, METEOR
explicitly handles synonyms and stemming, as well as word-order
changes, and uses F1 score, which combines precision and recall.
METEOR is generally more robust than BLEU to paraphrasing that 
retains the meaning of a text. BERTscore is a primarily semantic
metric; it measures the cosine similarity between vectors in an
embedding. It is very robust to paraphrasing, providing an excellent 
test of semantic similarity, although in some cases it may exhibit 
semantic blindness~\cite{nikiema2025small}.
Together, these metrics provide different perspectives for assessing 
syntactic and semantic similarity between two texts.

\subsection{Metric interpretations} \label{section:metric_interpretations}
Text similarity metrics do not typically have good absolute interpretations,
and only relative comparisons under the same testing conditions are generally
 advised~\cite{babych2014automated,fomicheva2019taking}. There are, however, 
 some rough interpretations that have been suggested for BLEU by 
 Google~\citep{google_bleu_docs}, provided in Table~\ref{tab:bleu_interp}.
 \citet{lavie2011mtevaluation}, one of the creators of METEOR,
suggests that METEOR gives higher scores than BLEU, and 
anything scoring above 0.50 is probably an understandable translation, 
while anything above 0.70 is excellent. \citet{lavie2011mtevaluation} also 
provides colour gradients for the BLEU and METEOR metric values, which we 
used to colour-code the values in Table~\ref{tab:llm-similarities}. To the 
best of our knowledge, there is no absolute interpretation for BERTscore.
Furthermore, since BERTscore is based on vector embeddings, it can be used 
with different base models, which complicates interpretation. Here, we anchor
the colour gradient endpoints of BERTscore to 0 (matching BLEU and METEOR) and
to the highest achieved BERTscore in our data (0.82). This makes the colouring 
visually comparable across metrics.

\begin{table}[H]
\caption{Suggested interpretation of the BLEU metric for machine translation~\cite{google_bleu_docs}}
\label{tab:bleu_interp}
\centering
\small
\setlength{\tabcolsep}{9pt}
\begin{tabular}{cl}
\toprule 
\textbf{BLEU Score}   & \textbf{Interpretation}                                            \\
\midrule 
0.00 -- 0.09         & Almost useless                                            \\ \midrule 
0.10 -- 0.19         & Hard to get the gist                                      \\ \midrule 
0.20 -- 0.29         & The gist is clear, but has significant grammatical errors \\ \midrule 
0.30 -- 0.39         & Understandable to good translations                       \\ \midrule 
0.40 -- 0.49         & High quality translations                                 \\ \midrule 
0.50 -- 0.59         & Very high quality, adequate, and fluent translations      \\ \midrule 
0.60 -- 1.00         & Quality often better than human                           \\ \bottomrule 
\end{tabular}

\end{table}

\newpage
\subsection{Prompts for the tested explanatory tasks} \label{section:LLM_prompts}
For each explanatory task, we show the direct and the \guidedquery.

\subsubsection{Explanatory task: Rats BN - Summary} \label{section:LLM_rats_summary_prompt}

\vspace*{3mm}
\begin{textprompt}[label=\textbf{Direct prompt: Rats BN - Summary}]
We are talking about this causal Bayesian network and set of findings:

BN: [{"name": "E", "states": ["given", "not_given"], "parents": [], "cpt": [[0.5, 0.5]]}, {"name": "N", "states": ["high", "low"], "parents": ["E"], "cpt": [[0.2, 0.8], [0.96, 0.04]]}, {"name": "SA", "states": ["high", "low"], "parents": ["E", "N"], "cpt": [[0.03, 0.97], [0.89, 0.11], [0.12, 0.88], [0.98, 0.02]]}, {"name": "S", "states": ["short", "long"], "parents": ["SA"], "cpt": [[0.65, 0.35], [0.45, 0.55]]}] 
Evidence: {"E": "given", "S": "short"}

Target: "SA" is "high"

Summarize in ordinary language how the findings contribute to the probability that <Target>.

Instructions
------------
Use no more than 48 words.
\end{textprompt}

\begin{textprompt}[label=\textbf{Guided prompt: Rats BN - Summary}]
|\plusguided|

A "contribution" is defined as how much difference learning a finding (or findings) makes to the probability of a <Target> when all the other findings are already known.

Describe probability contributions only according to the following scale:
    greatly increases = 31
    moderately increases = 16
    slightly increases = 1
    doesn't change = 0%
    slightly reduces = -1
    moderately reduces = -16
    greatly reduces = -31

First, describe the contribution all the findings make together.

Second, list with dot points each finding's contribution (in order of decreasing contribution magnitude). For each finding, if not knowing another finding would change its contribution by at least 10

Avoid pronouns if they're ambiguous.

Refer to findings with phrasing like "the finding that variable is state" or "finding out that variable is state".

Do not mention any other variables or any causal connections.
\end{textprompt}

\newpage
\subsubsection{Explanatory task: Podunk BN - Summary}
\label{section:LLM_podunk_summary_prompt}

\vspace*{3mm}
\begin{textprompt}[label=\textbf{Direct prompt: Podunk BN - Summary}]
We are talking about this causal Bayesian network and set of findings:

BN: [{"name": "PB", "states": ["visited", "not_visited"], "parents": [], "cpt": [[0.25, 0.75]]}, {"name": "DS", "states": ["used", "not_used"], "parents": ["PB"], "cpt": [[0.75, 0.25], [0.08, 0.92]]}, {"name": "JIC", "states": ["eaten", "not_eaten"], "parents": ["PB"], "cpt": [[0.664, 0.336], [0.05, 0.95]]}, {"name": "D", "states": ["present", "absent"], "parents": ["PB", "DS"], "cpt": [[0.656364, 0.343636], [0.7575411, 0.2424589], [0.0521211905212119, 0.9478788094787881], [0.08584743656610254, 0.9141525634338975]]}, {"name": "M", "states": ["inherited", "not_inherited"], "parents": [], "cpt": [[0.2, 0.8]]}, {"name": "P", "states": ["persistent", "not_persistent"], "parents": ["M", "D"], "cpt": [[0.77, 0.23], [0.48, 0.52], [0.3, 0.7], [0.01, 0.99]]}, {"name": "B", "states": ["visible", "not_visible"], "parents": ["D"], "cpt": [[0.3825, 0.6175], [0.01, 0.99]]}] 
Evidence: {"DS": "used", "M": "inherited", "P": "persistent", "B": "not_visible"}

Target: "D" is "present"

Summarize in ordinary language how the findings contribute to the probability that <Target>.

Instructions
------------
Use no more than 120 words.
\end{textprompt}

\begin{textprompt}[label=\textbf{Guided prompt: Podunk BN - Summary}]
|\plusguided|

A "contribution" is defined as how much difference learning a finding (or findings) makes to the probability of a <Target> when all the other findings are already known.

Describe probability contributions only according to the following scale:
    greatly increases = 31
    moderately increases = 16
    slightly increases = 1
    doesn't change = 0%
    slightly reduces = -1
    moderately reduces = -16
    greatly reduces = -31

First, describe the contribution all the findings make together.

Second, list with dot points each finding's contribution (in order of decreasing contribution magnitude). For each finding, if not knowing another finding would change its contribution by at least 10

Avoid pronouns if they're ambiguous.

Refer to findings with phrasing like "the finding that variable is state" or "finding out that variable is state".

Do not mention any other variables or any causal connections.
\end{textprompt}

\newpage
\subsubsection{Explanatory task: Rats BN - Path}
\label{section:LLM_rats_path_prompt}

\vspace*{3mm}
\begin{textprompt}[label=\textbf{Direct prompt: Rats BN - Path}]
We are talking about this causal Bayesian network and set of findings:

BN: [{"name": "E", "states": ["given", "not_given"], "parents": [], "cpt": [[0.5, 0.5]]}, {"name": "N", "states": ["high", "low"], "parents": ["E"], "cpt": [[0.2, 0.8], [0.96, 0.04]]}, {"name": "SA", "states": ["high", "low"], "parents": ["E", "N"], "cpt": [[0.03, 0.97], [0.89, 0.11], [0.12, 0.88], [0.98, 0.02]]}, {"name": "S", "states": ["short", "long"], "parents": ["SA"], "cpt": [[0.65, 0.35], [0.45, 0.55]]}]
Evidence: {"E": "given", "S": "short"}

Target: "SA" is "high"
FocusFinding: "E" is "given"

Describe in ordinary language how the finding <FocusFinding> contributes to the probability that <Target>.

Instructions
------------
Use no more than 72 words.
\end{textprompt}

\begin{textprompt}[label=\textbf{Guided prompt: Rats BN - Path}]
|\plusguided|

A "contribution" is defined as how much difference learning a finding (or findings) makes to the probability that <Target> when all the other findings are already known.

Describe probability contributions only according to the following scale:
    greatly increases = 31
    moderately increases = 16
    slightly increases = 1
    doesn't change = 0%
    slightly reduces = -1
    moderately reduces = -16
    greatly reduces = -31

A "direct connection" from variable A to variable B is defined as a single arrow where its head or tail is attached to variable A and its head or tail is attached to variable B.

An "indirect connection" from variable A to variable B is defined as a sequence of variables and arrows where:
Only the first arrow in the sequence has its head or tail attached to A.
Only the last arrow in the sequence has its head or tail attached to B.
For each successive pair of arrows in the sequence, both have their head or tail attached to the same variable (which is defined as an "intermediate variable" in the connection).
In the sequence, there is more than one arrow and each arrow occurs only once.

Measure the difference an indirect connection makes to the contribution of the finding that <finding variable is finding state> by:
Identifying the last variable on the indirect connection that is not included in any other connection.
Identifying the change in the probability distribution that occurs on this last variable when the finding is added.
Assessing how much difference this change makes to the contribution of the finding.

Measure the difference a direct connection makes to the contribution of the finding that <finding variable is finding state> by:
Measuring the difference each indirect connection makes to the contribution of the finding.
Subtracting all these indirect differences from the contribution of the finding.

First, state how many connections there are between Ecstazine and <Target>.

Second, list with dot points each connection according to how much difference (in order of decreasing magnitude) it makes to the contribution of the finding. For each connection:
For each intermediate variable in sequence, describe any increases the finding makes to the probability of any of the intermediate variable's states.
Then describe the difference the finding makes to the probability that <target variable is target state> due to the connection.

Last, describe the contribution the finding makes due to all the connections together.

In your response, you may use words like "causes" and "effects", but not "arrow", "head" or "tail".

Avoid pronouns if they're ambiguous.

Refer to findings with phrasing like "the finding that variable is state" or "finding out that variable is state".

Do not mention any other variables or any causal connections.
\end{textprompt}

\newpage
\subsubsection{Explanatory task: Podunk BN - Common Effect}
\label{section:LLM_podunk_collider_prompt}

\vspace*{3mm}
\begin{textprompt}[label=\textbf{Direct prompt: Podunk BN - Common effect}]
We are talking about this causal Bayesian network and set of findings:

BN: [{"name": "PB", "states": ["visited", "not_visited"], "parents": [], "cpt": [[0.25, 0.75]]}, {"name": "DS", "states": ["used", "not_used"], "parents": ["PB"], "cpt": [[0.75, 0.25], [0.08, 0.92]]}, {"name": "JIC", "states": ["eaten", "not_eaten"], "parents": ["PB"], "cpt": [[0.664, 0.336], [0.05, 0.95]]}, {"name": "D", "states": ["present", "absent"], "parents": ["PB", "DS"], "cpt": [[0.656364, 0.343636], [0.7575411, 0.2424589], [0.0521211905212119, 0.9478788094787881], [0.08584743656610254, 0.9141525634338975]]}, {"name": "M", "states": ["inherited", "not_inherited"], "parents": [], "cpt": [[0.2, 0.8]]}, {"name": "P", "states": ["persistent", "not_persistent"], "parents": ["M", "D"], "cpt": [[0.77, 0.23], [0.48, 0.52], [0.3, 0.7], [0.01, 0.99]]}, {"name": "B", "states": ["visible", "not_visible"], "parents": ["D"], "cpt": [[0.3825, 0.6175], [0.01, 0.99]]}]
Evidence: {"DS": "used", "M": "inherited", "P": "persistent", "B": "not_visible"}

Target: "D" is "present"
FocusFinding: "M" is "inherited"

Describe in ordinary language how the finding <FocusFinding> contributes to the probability that <Target>.

Instructions
------------
Use no more than 144 words.
\end{textprompt}

\begin{textprompt}[label=\textbf{Guided prompt: Podunk BN - Common effect}]
|\plusguided|

A "contribution" is defined as how much difference learning a finding (or findings) makes to the probability that <Target> when all the other findings are already known.

Describe probability contributions only according to the following scale:
    greatly increases = 31
    moderately increases = 16
    slightly increases = 1
    doesn't change = 0%
    slightly reduces = -1
    moderately reduces = -16
    greatly reduces = -31

First, state the causal relations between the parent nodes and the child, specifying whether they increase or decrease the probability of the child.

Second, use dot points to say (1) what the contribution of the focus finding would be if we did not know the state of the child, (2) what the contribution of the child finding would be if we did not know the focus finding, (3) how adding the focus finding would change the probability of the child finding, and what the contribution of the child finding is now given that we know the focus finding.

Third, describe how the focus finding contributes to the probability of the target state by indicating whether it increases or decreases the contribution of the child state without the focus finding.

In your response, you may use words like "causes" and "effects", but not "arrow", "head" or "tail".

Avoid pronouns if they're ambiguous.

Refer to findings with phrasing like "the finding that variable is state" or "finding out that variable is state".

Do not mention any other variables or any causal connections.
\end{textprompt}

\newpage
\subsection{User-study and LLM-generated explanations for the tested explanatory tasks}
\label{section:llm_scenarios}
For each explanatory task, we show the visual and verbal 
explanation from the user study, and the first two texts 
generated for each prompt type by GPT~5.2 (configured for 
high reasoning and tool use), which are representative of
the LLM-generated texts.

\subsubsection{Explanatory task: Rats BN - Summary}
\label{section:LLM_rats_summary}
\begin{figure}[H]
    \centering
    \includegraphics[width=0.17\textwidth]{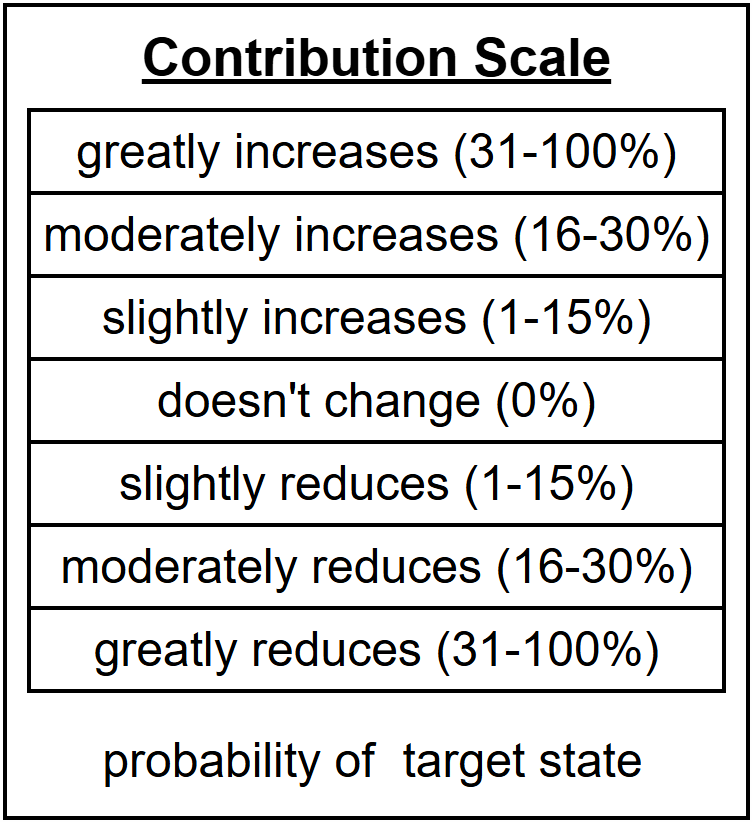}
    \hfill
    \includegraphics[width=0.35\textwidth]{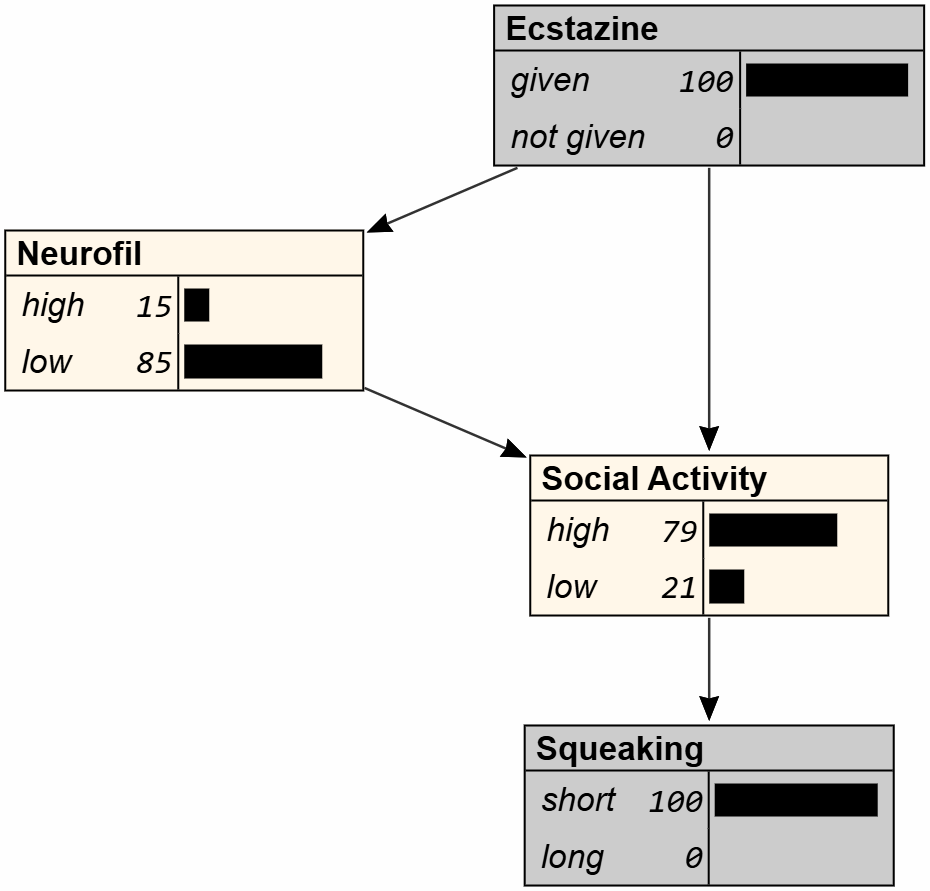}
    \hfill
    \parbox[b]{0.45\textwidth}{
        \includegraphics[width=\linewidth]{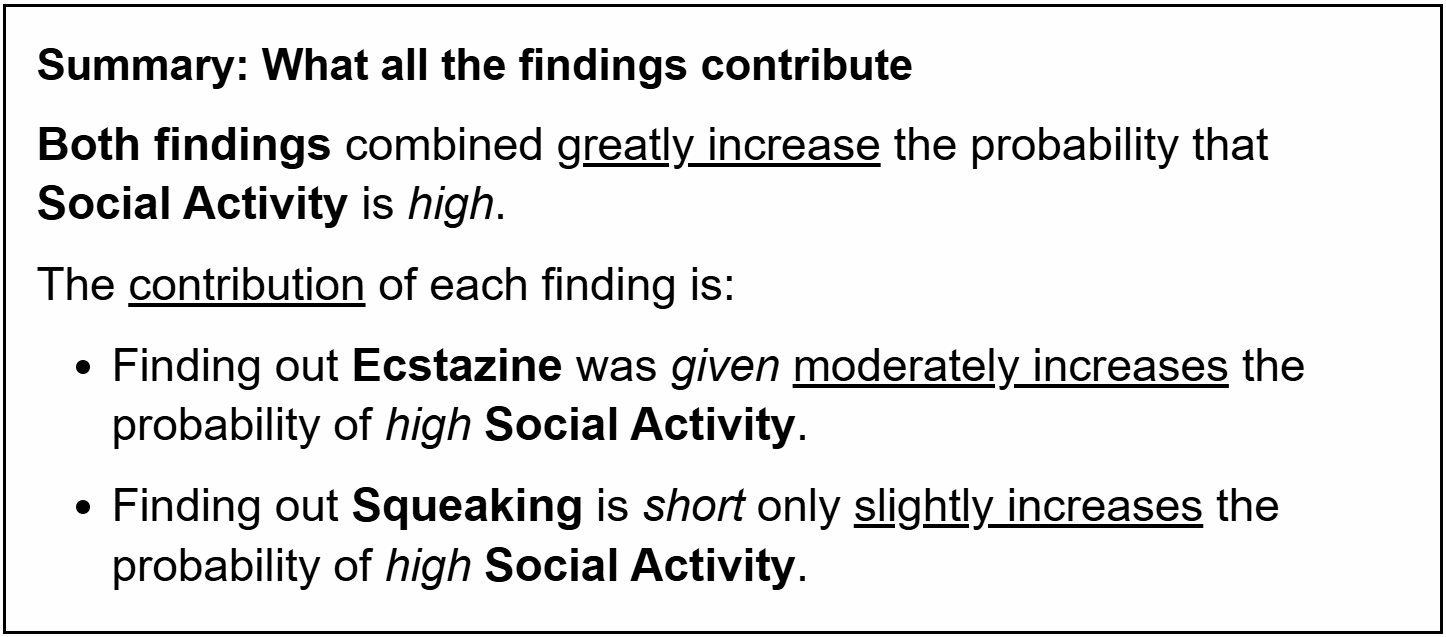}
        \vspace*{0.11\textwidth}
    }
    \caption{Verbal explanation in the user study for the Rats BN - Summary.}
    \label{fig:LLM_rats_summary}
\end{figure}

\begin{textprompt}[label=\textbf{Sample outputs -- \directquery: Rats BN - Summary}]
Because **Ecstazine is given**, **Neurofil** is usually **low**, and when **Ecstazine is given** with **Neurofil low**, **Social Activity** is very likely **high**. Also, observing **Squeaking is short** slightly favors **Social Activity high** because short outcomes are more common when **Social Activity is high** than low.
|\myhrule|
Knowing Ecstazine was given makes Social Activity=high likely, since with Ecstazine given Neurofil is usually low and Social Activity is high in that case. Observing Squeaking=short adds extra support because short outcomes are more common when Social Activity is high than low, raising Social Activity=high to about 79
\end{textprompt}

\begin{textprompt}[label=\textbf{Sample outputs -- \guidedquery: Rats BN - Summary}]
Together, the findings that Ecstazine is given and Squeaking is short greatly increase the probability that Social Activity is high.  
- The finding that Ecstazine is given moderately increases that probability.  
- The finding that Squeaking is short slightly increases that probability.
|\myhrule|
Together, finding out that Ecstazine is given and Squeaking is short greatly increases the probability that Social Activity is high.

- The finding that Ecstazine is given moderately increases the probability that Social Activity is high.  
- Finding out that Squeaking is short slightly increases the probability that Social Activity is high.
\end{textprompt}

\subsubsection{Explanatory task: Podunk BN - Summary}
\label{section:LLM_podunk_summary}

\begin{figure}[H]
    \centering
    \includegraphics[width=0.17\textwidth]{figF23a_scale_uncoloured-400}
    \hfill
    \includegraphics[width=0.45\textwidth]{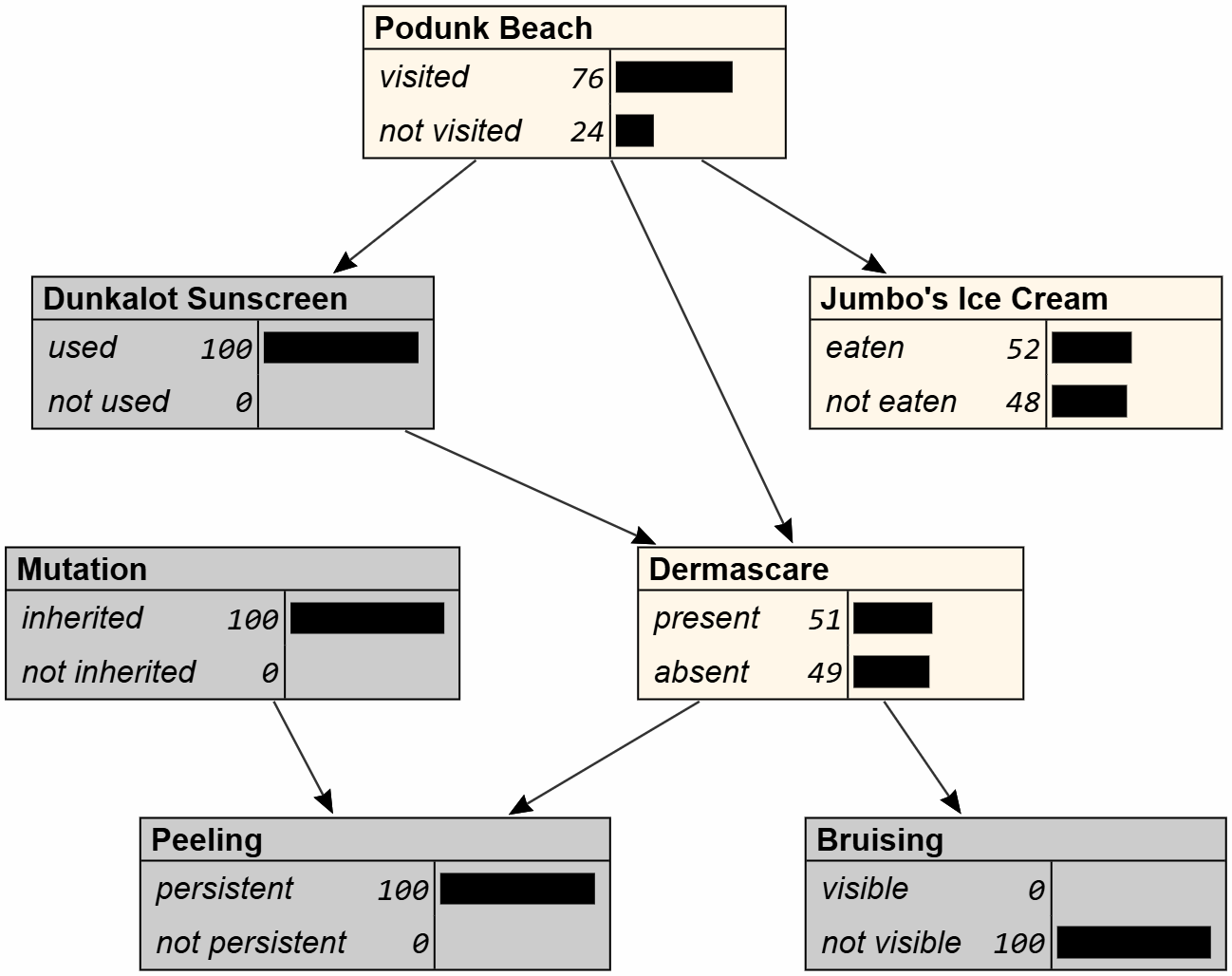}
    \hfill
    \parbox[b]{0.35\textwidth}{
        \includegraphics[width=\linewidth]{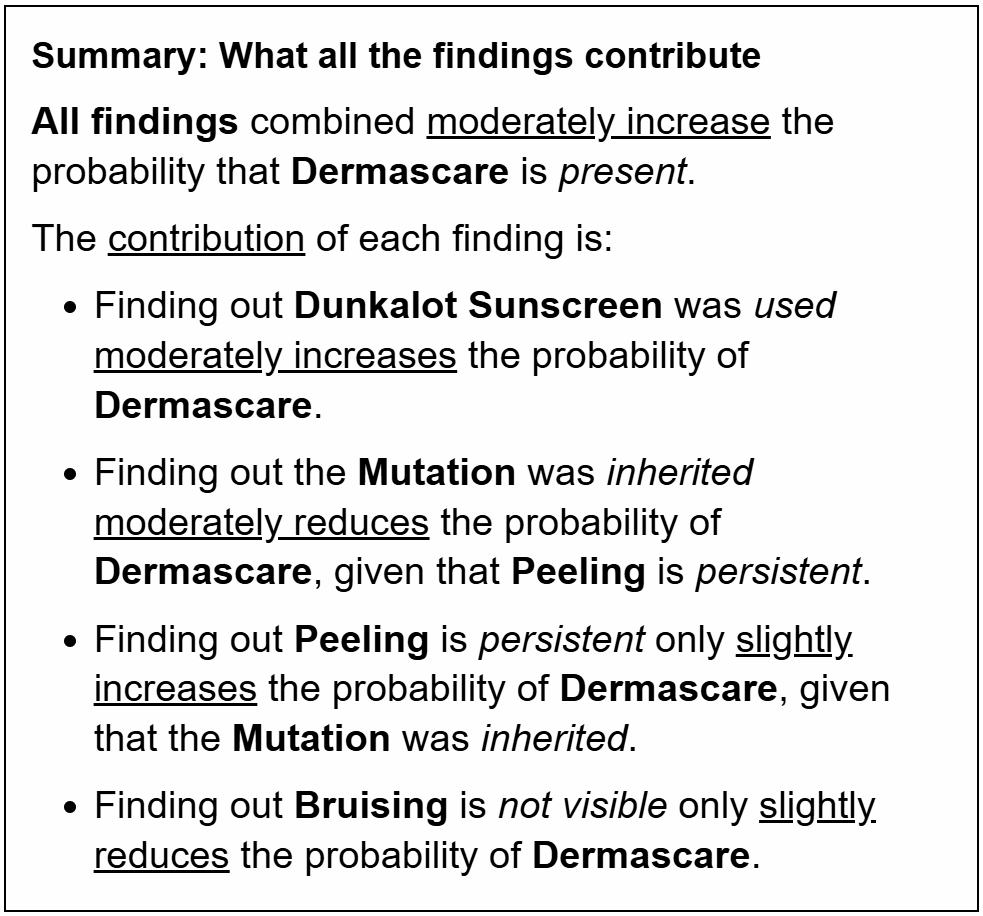}
        \vspace*{0.005\textwidth}
    }
    \caption{Verbal explanation in the user study for the Podunk BN - Summary.}
    \label{fig:LLM_podunk_summary}
\end{figure}


\begin{textprompt}[label=\textbf{Sample outputs -- \directquery: Podunk BN - Summary}]
Using **Dunkalot Sunscreen** makes **Podunk Beach (visited)** much more likely; and **having visited Podunk Beach** makes **Dermascare being present** far more likely than if Podunk Beach was not visited. The fact that **Peeling is persistent** in someone with **inherited Mutation** also points toward **Dermascare**, because persistence is more common when **Dermascare is present** than when it is absent (given inherited Mutation). However, **Bruising being not visible** argues against **Dermascare**, since **Bruising is almost always not visible when Dermascare is absent**, but is noticeably more often visible when **Dermascare is present**. Overall, Dunkalot Sunscreen and persistent Peeling push Dermascare up, while not-visible Bruising pulls it down.
|\myhrule|
“Dunkalot Sunscreen used” makes it more likely the person visited Podunk Beach (Dunkalot Sunscreen is usually used after a Podunk Beach visit), and a Podunk Beach visit together with Dunkalot Sunscreen use is associated with a much higher chance that Dermascare is present than Dunkalot Sunscreen use without a visit.  

“Mutation inherited” combined with “Peeling persistent” pushes strongly toward Dermascare being present, because persistent Peeling is far more common when Dermascare is present (especially given inherited Mutation).  

“Bruising not visible” pulls the other way, since Dermascare tends to make Bruising visible more often; the absence of visible Bruising is more consistent with Dermascare being absent.  

Net effect: Dermascare becomes much more likely than at baseline, roughly around even odds.
\end{textprompt}

\begin{textprompt}[label=\textbf{Sample outputs -- \guidedquery: Podunk BN - Summary}]
All findings together **moderately increase** the probability that **Dermascare is present**.

- The finding that **Dunkalot Sunscreen is used** **moderately increases** the probability that **Dermascare is present**.  
- The finding that **Mutation is inherited** **moderately reduces** the probability that **Dermascare is present**; without knowing the finding that **Peeling is persistent**, this reduction would be much smaller.  
- Finding out that **Peeling is persistent** **slightly increases** the probability that **Dermascare is present**; without knowing the finding that **Mutation is inherited**, this increase would be much larger.  
- The finding that **Bruising is not_visible** **slightly reduces** the probability that **Dermascare is present**.
|\myhrule|
Taken together, the findings **moderately increase** the probability that **Dermascare is present**.

- The finding that **Dunkalot Sunscreen is used** **moderately increases** the probability that **Dermascare is present**.  
- The finding that **Mutation is inherited** **moderately reduces** the probability that **Dermascare is present**; without knowing the finding that **Peeling is persistent**, the finding that **Mutation is inherited** **doesn't change** the probability.  
- The finding that **Peeling is persistent** **slightly increases** the probability that **Dermascare is present**; without knowing the finding that **Mutation is inherited**, the finding that **Peeling is persistent** would **greatly increase** the probability.  
- The finding that **Bruising is not_visible** **slightly reduces** the probability that **Dermascare is present**.
\end{textprompt}

\subsubsection{Explanatory task: Rats BN - Path}
\label{section:LLM_rats_path}

\begin{figure}[H]
    \centering
    \includegraphics[width=0.17\textwidth]{figF23a_scale_uncoloured-400}
    \hfill
    \includegraphics[width=0.35\textwidth]{figF23b_LLM_rats_all_findings}
    \hfill
    \parbox[b]{0.45\textwidth}{
        \includegraphics[width=\linewidth]{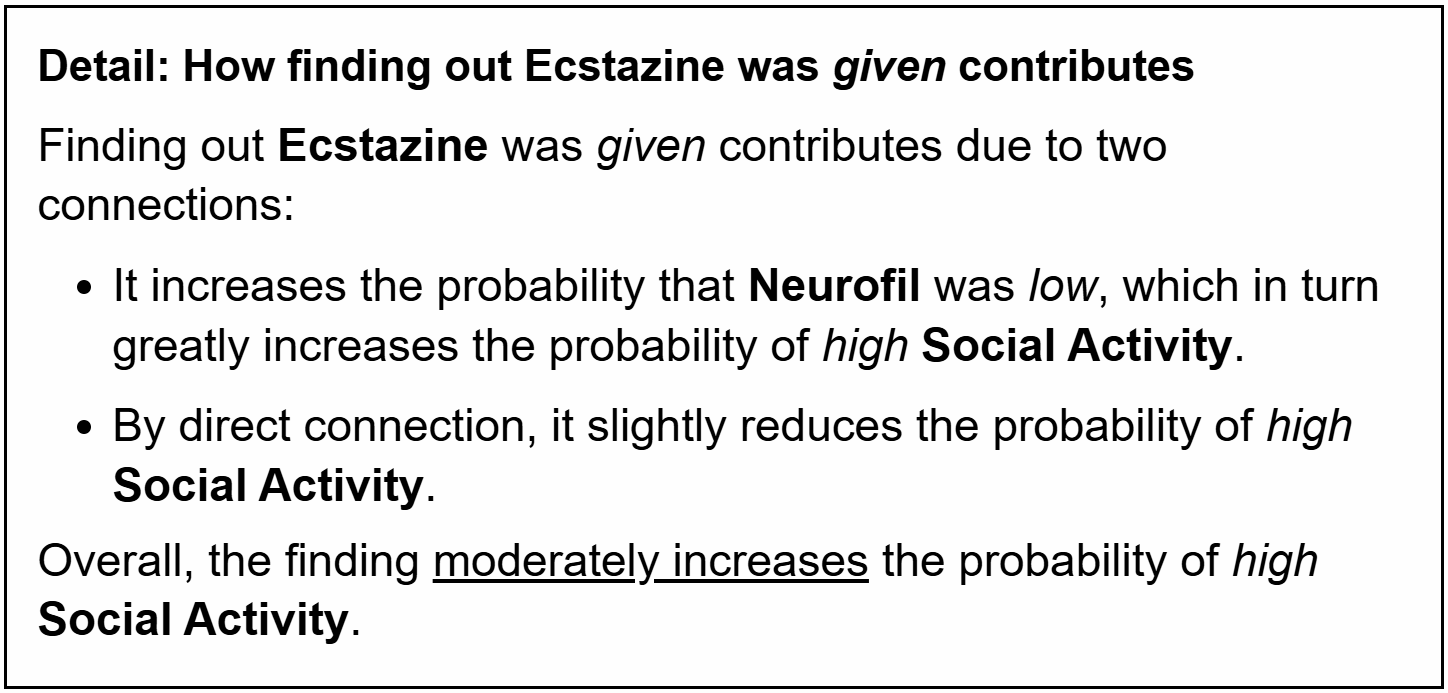}
        \vspace*{0.094\textwidth}
    }
    \caption{Verbal explanation in the user study for the Rats BN - Path.}
    \label{fig:LLM_rats_path}
\end{figure}

\begin{textprompt}[label=\textbf{Sample outputs -- \directquery: Rats BN - Path}]
Knowing **Ecstazine is given** changes expectations about **Neurofil**: it makes **low Neurofil** far more likely. **Low Neurofil** strongly favors **Social Activity** being **high** (compared with high Neurofil). So the evidence **Ecstazine=given** pushes probability toward **Social Activity=high**, and the observed **short Squeaking**—more common when Social Activity is high—amplifies this. Overall, Ecstazine=given provides strong support for Social Activity=high via the **Ecstazine→Neurofil→Social Activity** pathway.
|\myhrule|
Knowing **Ecstazine is given** shifts the network toward **Social Activity being high** in two ways. First, Ecstazine directly affects Social Activity: when Ecstazine is given and **Neurofil is low**, Social Activity is usually high. Second, Ecstazine also makes **Neurofil more likely low** (rather than high), and low Neurofil strongly favors high Social Activity. So Ecstazine=given boosts the probability that Social Activity is high (consistent with Squeaking being short).
\end{textprompt}

\begin{textprompt}[label=\textbf{Sample outputs -- \guidedquery: Rats BN - Path}]
2 connections.

- Ecstazine–Neurofil–Social Activity: the finding that Ecstazine is given greatly increases the chance that Neurofil is low. This greatly increases the probability that Social Activity is high.

- Ecstazine–Social Activity: after accounting for the change in Neurofil, the finding that Ecstazine is given slightly reduces the probability that Social Activity is high.

Together, the finding that Ecstazine is given moderately increases the probability that Social Activity is high.
|\myhrule|
There are 2 connections between Ecstazine and Social Activity.

- Indirect through Neurofil: finding out that Ecstazine is given makes Neurofil low much more likely (≈49
- Direct: aside from the Neurofil-mediated effect, finding out that Ecstazine is given slightly reduces Social Activity high (≈−5

Together, finding out that Ecstazine is given moderately increases Social Activity high (≈+26
\end{textprompt}

\newpage
\subsubsection{Explanatory task: Podunk BN - Common Effect}
\label{section:LLM_podunk_collider}

\begin{figure}[H]
    \centering
    \includegraphics[width=0.17\textwidth]{figF23a_scale_uncoloured-400}
    \hfill
    \includegraphics[width=0.45\textwidth]{figF24b_LLM_podunk_all_findings-250}
    \hfill
    \parbox[b]{0.35\textwidth}{
        \includegraphics[width=\linewidth]{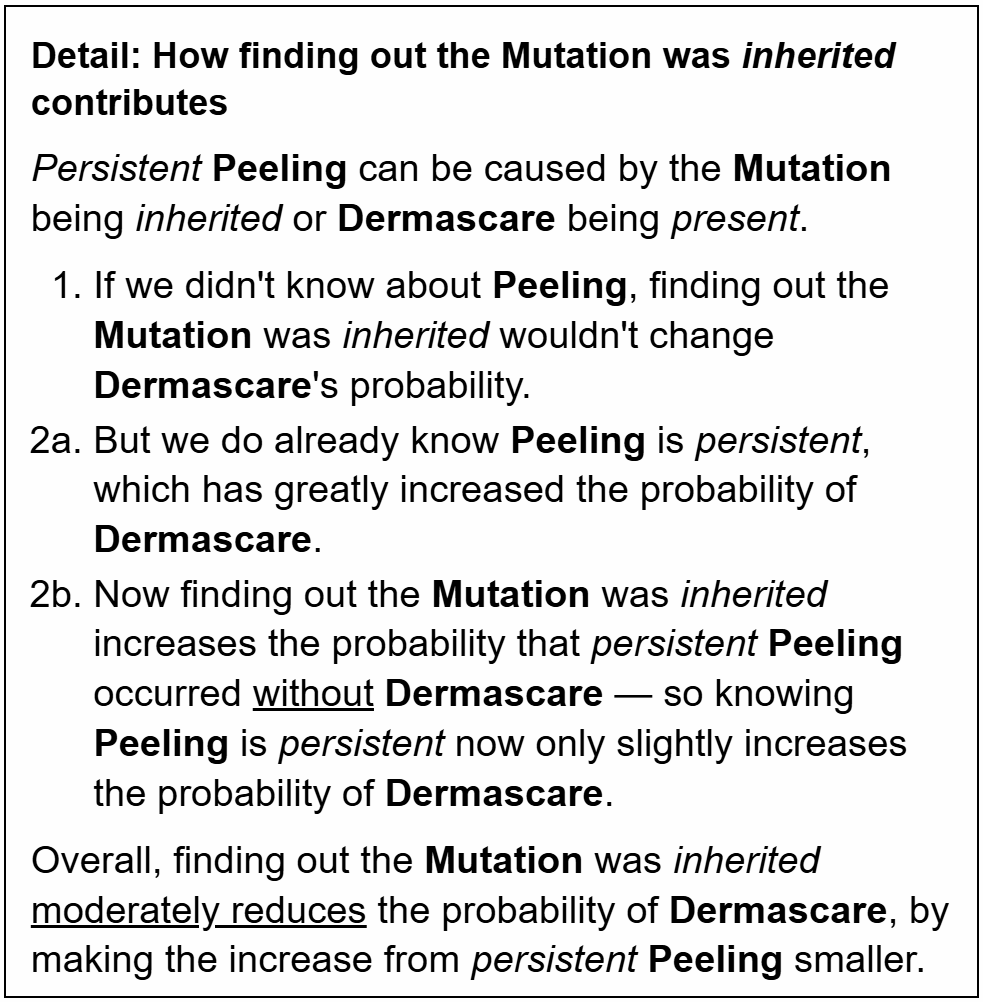}
    }
    \caption{Verbal explanation in the user study for the Podunk BN - Common effect.}
    \label{fig:LLM_podunk_collider}
\end{figure}

\begin{textprompt}[label=\textbf{Sample outputs -- \directquery: Podunk BN - Common effect}]
Knowing **Mutation is inherited** changes how we interpret the fact that **Peeling is persistent**. Because Mutation directly makes persistence more likely, the persistence finding can be explained by inheritance even if **Dermascare is absent**. So, once we learn Mutation is inherited, the persistence evidence becomes **less** diagnostic for Dermascare, and it contributes by **pulling down** the probability that **Dermascare is present** compared with a situation where Mutation was not inherited (in which case persistence would point much more strongly to Dermascare).  

Given the other evidence, **Dunkalot Sunscreen = used** still tends to push Dermascare upward, and **Bruising = not_visible** pushes Dermascare downward; Mutation = inherited mainly weakens the “persistence implies Dermascare” argument.
|\myhrule|
The finding that **Mutation is inherited** mainly matters because **Mutation is a direct cause of Peeling (persistent)**. Since we already observe **Peeling is persistent**, knowing that the condition was inherited provides a strong alternative explanation for why persistence happened, even if **Dermascare were absent**.  

That means persistence no longer points as strongly to **Dermascare being present**: the “need” for Dermascare to explain Peeling is reduced. In Bayesian-network terms, **Mutation inherited “explains away” some of the support for Dermascare** that would otherwise come from observing persistent symptoms.  

So, given the other evidence, **Mutation = inherited tends to *decrease* the probability that Dermascare is present**, compared with a case where Mutation were unknown or not inherited.
\end{textprompt}

\vspace*{-9mm}
\begin{textprompt}[label=\textbf{Sample outputs -- \guidedquery: Podunk BN - Common effect}]
Dermascare being present increases the probability that Peeling is persistent. The finding that Mutation is inherited also increases the probability that Peeling is persistent.

- If the state of Peeling were unknown, the finding that Mutation is inherited doesn’t change the probability that Dermascare is present (0
- If the state of Mutation were unknown, the finding that Peeling is persistent greatly increases the probability that Dermascare is present (about +32
- Adding the finding that Mutation is inherited greatly increases the probability that Peeling is persistent (about +38

Overall, the finding that Mutation is inherited decreases the contribution that the finding that Peeling is persistent would otherwise make, so it moderately reduces the probability that Dermascare is present (about −20
|\myhrule|
Both the finding that Mutation is inherited and the state that Dermascare is present increase the probability that Peeling is persistent; Mutation not inherited or Dermascare absent decrease it.

- If the state of Peeling were unknown (given the other findings), finding out that Mutation is inherited would **greatly increase** the probability that Peeling is persistent.  
- If the state of Mutation were unknown, finding out that Peeling is persistent would **greatly increase** the probability that Dermascare is present.  
- Adding the finding that Mutation is inherited would **greatly increase** the probability that Peeling is persistent; given the finding that Mutation is inherited, finding out that Peeling is persistent only **slightly increases** the probability that Dermascare is present.

Therefore, the finding that Mutation is inherited **moderately reduces** the probability that Dermascare is present by decreasing the contribution of the finding that Peeling is persistent.
\end{textprompt}